\documentclass[twocolumn,aps,prb,superscriptaddressx,amsfonts,longbibliography]{revtex4-2}
\usepackage[colorlinks,linkcolor=blue,urlcolor=red,citecolor=red]{hyperref}

\usepackage{graphicx,amsmath,amsfonts,amssymb,capt-of}
\usepackage[tight]{subfigure}
\usepackage{tikz}
\usetikzlibrary{quantikz}
\usepackage{yquant}
\usepackage{listofitems}
\usepackage{siunitx}
\usepackage{bigints}
\usepackage{tabularx}
\usepackage{mathtools}
\usepackage{float}
\usepackage{multirow}
\usepackage{newfloat}
\usepackage{array}
\usepackage{tikz}
\usetikzlibrary{decorations.pathreplacing}
\usetikzlibrary{colorbrewer}
\usetikzlibrary{arrows}
\usetikzlibrary{arrows.meta}
\usetikzlibrary{calc}
\usetikzlibrary{positioning}
\usepackage{makecell}
\usepackage{comment}
\usepackage{physics}
\usepackage[utf8]{inputenc}
\usepackage{amsmath}

\usepackage{geometry}
\usepackage{xcolor}
\usepackage{bbm}
\newcolumntype{P}[1]{>{\centering\arraybackslash}p{#1}}

\usepackage{enumitem}
\usepackage{chemfig}
\usepackage{circuitikz}
\usepackage[normalem]{ulem}
\newcolumntype{x}[1]{>{\centering\let\newline\\\arraybackslash\hspace{0pt}}p{#1}}
\DeclareFloatingEnvironment[
    fileext=loa,
    listname=List of Algorithms,
    name=ALGORITHM,
    placement=tbhp,
]{algorithm}

\makeatletter
\newcommand{\vast}{\bBigg@{4}}
\newcommand{\Vast}{\bBigg@{5}}
\newcommand{\VAst}{\bBigg@{6}}
\makeatother

 \newenvironment{packed_enum}{
 \begin{enumerate}
   \setlength{\leftmargin}{2em}
   \setlength{\topsep}{1pt}
   \setlength{\partopsep}{1pt}
   \setlength{\itemsep}{1pt}
   \setlength{\parskip}{0pt}
   \setlength{\parsep}{0pt}
 }{\end{enumerate}}

\begin{document}

\normalem
\title{A large language model-type architecture for high-dimensional molecular potential energy surfaces}
\author{Xiao Zhu, Srinivasan S. Iyengar\email{Corresponding Author: iyengar@iu.edu},}
\affiliation{Department of Chemistry, Department of Physics, and the Indiana University Quantum Science and Engineering Center (IU-QSEC), Indiana University, 800 E. Kirkwood Ave, Bloomington, IN-47405}
\date{\today}
\begin{abstract}
    Computing high-dimensional potential energy surfaces for molecular systems and materials is considered to be a great challenge in computational chemistry with potential impact in a range of areas including the fundamental prediction of reaction rates. In this paper, we design and discuss an algorithm that has similarities to large language models in generative AI and natural language processing. Specifically, we represent a molecular system as a graph which contains a set of nodes, edges, faces, etc. Interactions between these sets, which represent molecular subsystems in our case, are used to construct the potential energy surface for a reasonably sized chemical system with 51 nuclear dimensions. For this purpose, a family of neural networks that pertain to the graph-theoretically obtained subsystems get the job done for this 51 nuclear dimensional system. We then ask if this same family of lower-dimensional graph-based neural networks can be transformed to provide accurate predictions for a 186-dimensional potential energy surface. We find that our algorithm does provide accurate results for this larger-dimensional problem with sub-kcal/mol accuracy for the higher-dimensional potential energy surface problem. Indeed, as a result of these developments, here we produce the first efforts towards a full-dimensional potential energy surface for the protonated 21-water cluster (186 nuclear dimensions) at CCSD level accuracy. 
\end{abstract}

\maketitle

\section{Introduction}
\label{sec_intro}
The problem of computing accurate post-Hartree-Fock potential energy surfaces for large chemical systems is considered an exponentially complex task in quantum chemistry. However, computing such accurate potentials is essential for a range of problems that include not only the fundamental description of chemical reactions and molecular vibrations beyond the harmonic approximation\cite{varandas-PES,Bowman-IRPC-PES,Bowman-polybasis-fitting,alics2001efficient,jackle1996product,DieterPOTFIT,TuckerVibSpec2,full-dimPES-tunneling-Bowman2021}, but also topics such as the design of new molecular assemblies with desired properties
\cite{grossmann2023positionpapermaterialsdesign,Mat-Design-ML-JAP,Mat-Design-ML-AMR,Ceder-PhysRevB.82.125416,Aspuru-design}.
The complexity of describing these phenomena is governed by the following fundamental limitations. As the system size grows, the number of nuclear dimensions increases. For ${\cal D}$ nuclear dimensions with each dimension discretized and represented using $M$ multi-dimensional basis points where such basis-points may represent nuclear geometries, the number of such geometries grows exponentially with the number of nuclear dimensions, and the problem may scale as $O(M^{\cal D})$.\cite{Feynman1982,Feynman-Comp,hibbs,MCTDH-Meyer1,Nielsen-Chuang-QuantComp,varandas-PES,Rabitz-HDMR,manzhos2008using,Bowman-IRPC-PES,Bowman-polybasis-fitting,Otto_POTFIT,DieterPOTFIT,qwaimd-wavelet,nicole-Shannon}
Additionally, if post-Hartree Fock accuracy (such as coupled cluster theory) is desired, computing each potential energy may scale as $O(N^{7-10})$, for $N$ electrons\cite{schlegel1991computational}. Thus, the problem of computing accurate potential energy surfaces 
grows as $O(M^{\cal D}*N^{7-10})$, which becomes intractable even for moderate sized systems. 

Recently, many groups have used machine learning methods to compute accurate potential energy surfaces. 
\cite{waters_transfer_nnp,ml_compare_paesani, waters_nnp,H+CH3OH_NNP,H2+SH_NNP,critical_compare_nnp,dft-ccsd-transfer_nnp,H2O++H2_NNP,C2H5nnp_miller,Aspuru-Guzik-ML-QC}
However, for obtaining multidimensional potentials, the amount of data needed to train these models could also grow steeply, and additionally, when accurate electron correlation information is needed, as noted above, the training cost may be prohibitive.
\begin{figure}
\includegraphics[width=0.15\columnwidth]{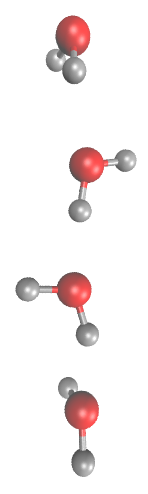}
\hspace{-4.5em}
\resizebox{0.95\columnwidth}{!}{\definecolor{purple1}{rgb}{0.6, 0.4, 0.8}
\definecolor{purple2}{rgb}{0, 0.4, 0}
\definecolor{purple3}{rgb}{0, 0, 0.8}
\definecolor{purple4}{rgb}{0, 0, 0}
\definecolor{orange}{rgb}{0.83, 0.4, 0.32}
\definecolor{ggray}{rgb}{0.55, 0.55, 0.55}

\tikzstyle{mynodeV1}=[thick,draw=purple1,fill=purple1!10,circle,minimum size=0.3cm,inner sep=0pt]
\tikzstyle{mynodeV2}=[thick,draw=purple2,fill=purple2!10,circle,minimum size=0.3cm,inner sep=0pt]
\tikzstyle{mynodeV3}=[thick,draw=purple3,fill=purple3!10,circle,minimum size=0.3cm,inner sep=0pt]
\tikzstyle{mynodeV4}=[thick,draw=purple4,fill=purple4!10,circle,minimum size=0.3cm,inner sep=0pt]
\tikzstyle{mynodeH}=[thick,draw=orange,fill=orange!10,circle,minimum size=0.3cm,inner sep=0pt]

\tikzstyle{virtual}=[thick,draw=white,fill=white!30,circle,minimum size=0.3cm,inner sep=0pt]

\tikzstyle{mynodeN1}=[thick,draw=purple1,fill=purple1!30,circle,minimum size=0.6cm,inner sep=0pt]
\tikzstyle{mynodeN2}=[thick,draw=purple2,fill=purple2!30,circle,minimum size=0.6cm,inner sep=0pt]
\tikzstyle{mynodeN3}=[thick,draw=purple3,fill=purple3!30,circle,minimum size=0.6cm,inner sep=0pt]
\tikzstyle{mynodeN4}=[thick,draw=purple4,fill=purple4!30,circle,minimum size=0.6cm,inner sep=0pt]

\begin{tikzpicture}[x=1cm,y=0.75cm,scale=3]

\node[mynodeN1] (N0-1) at (0,0.9) {};
\node[mynodeN2] (N0-2) at (0.15,0.3) {};
\node[mynodeN3] (N0-3) at (-0.02,-0.3) {};
\node[mynodeN4] (N0-4) at (0.1,-0.9) {};
\draw[line width=1mm, ggray] (N0-1) -- (N0-2);
\draw[line width=1mm, ggray] (N0-2) -- (N0-3);
\draw[line width=1mm, ggray] (N0-3) -- (N0-4);

    \readlist\Nnod{12,12,12,12,12,1} 
    \foreachitem \N \in \Nnod{ 
    \foreach \i [evaluate={\x=\Ncnt/2.5; \y=\N/10-0.2*\i+0.1; \prev=int(\Ncnt-1);}] in {1,...,\N}{ 
      \node[mynodeH] (N\Ncnt-\i) at (\x,\y) {};
      \ifnum\Ncnt>1 
        \ifnum\Ncnt<\Nnodlen
            \node[] () at (\x,\N/10+0.1) {} 
        \fi
      \fi;
      \ifnum\Ncnt=1 \node[virtual] () at (\x,\N/10+0.1) {$\bar{x}$} \fi;
      \ifnum\Ncnt=\Nnodlen \node[virtual] () at (\x+0.1,\N/10+0.1) {$E(\bar{x})$} \fi;
    }
    }

  \readlist\Nnod{12,12,12,12,12,1} 
  \foreachitem \N \in \Nnod{ 
    \foreach \i [evaluate={\x=\Ncnt/2.5; \y=\N/10-0.2*\i+0.1; \prev=int(\Ncnt-1);}] in {1,...,\N}{ 
      \node[mynodeH] (N\Ncnt-\i) at (\x,\y) {};
      \ifnum\Ncnt>1 
        \foreach \j in {1,...,\Nnod[\prev]}{ 
          \draw[line width=0mm, ggray] (N\prev-\j) -- (N\Ncnt-\i); 
        }
      \fi 
    }
  }


  \readlist\Nnod{12} 
  \foreachitem \N \in \Nnod{ 
    \foreach \i [evaluate={\x=\Ncnt/2.5; \y=\N/10-0.2*\i+0.1; \prev=int(\Ncnt-1);}] in {1,...,3}{ 
      \node[mynodeV1] (N\Ncnt-\i) at (\x,\y) {};
      \ifnum\Ncnt>1 
        \foreach \j in {1,...,\Nnod[\prev]}{ 
          \draw[very thin] (N\prev-\j) -- (N\Ncnt-\i); 
        }
      \fi 
    }
  }

      \readlist\Nnod{12} 
    \foreachitem \N \in \Nnod{ 
    \foreach \i [evaluate={\x=\Ncnt/2.5; \y=\N/10-0.2*\i+0.1; \prev=int(\Ncnt-1);}] in {4,...,6}{ 
      \node[mynodeV2] (N\Ncnt-\i) at (\x,\y) {};
      \ifnum\Ncnt>1 
        \ifnum\Ncnt<\Nnodlen
            \node[] () at (\x,\N/10+0.1) {} 
        \fi
      \fi;
      \ifnum\Ncnt=1 \node[] () at (\x,\N/10+0.1) {} \fi;
      \ifnum\Ncnt=\Nnodlen \node[] () at (\x,\N/10+0.1) {} \fi;
    }
    }

          \readlist\Nnod{12} 
    \foreachitem \N \in \Nnod{ 
    \foreach \i [evaluate={\x=\Ncnt/2.5; \y=\N/10-0.2*\i+0.1; \prev=int(\Ncnt-1);}] in {7,...,9}{ 
      \node[mynodeV3] (N\Ncnt-\i) at (\x,\y) {};
      \ifnum\Ncnt>1 
        \ifnum\Ncnt<\Nnodlen
            \node[] () at (\x,\N/10+0.1) {} 
        \fi
      \fi;
      \ifnum\Ncnt=1 \node[] () at (\x,\N/10+0.1) {} \fi;
      \ifnum\Ncnt=\Nnodlen \node[] () at (\x,\N/10+0.1) {} \fi;
    }
    }

          \readlist\Nnod{12} 
    \foreachitem \N \in \Nnod{ 
    \foreach \i [evaluate={\x=\Ncnt/2.5; \y=\N/10-0.2*\i+0.1; \prev=int(\Ncnt-1);}] in {10,...,12}{ 
      \node[mynodeV4] (N\Ncnt-\i) at (\x,\y) {};
      \ifnum\Ncnt>1 
        \ifnum\Ncnt<\Nnodlen
            \node[] () at (\x,\N/10+0.1) {} 
        \fi
      \fi;
      \ifnum\Ncnt=1 \node[] () at (\x,\N/10+0.1) {} \fi;
      \ifnum\Ncnt=\Nnodlen \node[] () at (\x,\N/10+0.1) {} \fi;
    }
    }

\end{tikzpicture}}
\caption{\label{NN-full} 
The density of gray edges represents the number of network weights that need to be computed to obtain a representation of the potential energy surface. 
}
\end{figure}
 The complexity of obtaining accurate neural network(NN) models may be understood from the density of the edges on the right side of 
 Figure \ref{NN-full}. Here, the multi-dimensional electronic potential energy surface for a simple water wire system, depicted as a graphical chain with individual water molecules shown as nodes, is extrapolated using the neural network on the right. The gray edges represent the neural network weights that need to be computed through training, and the density of such edges, represents the complexity of the training problem. 
  Figure 
 \ref{NN-full} does not account for the steep algebraic scaling needed to compute the training data (which is generally represented by the span of the nuclear coordinates variable, $\bar{x}$ in Figure \ref{NN-full}) within the training set. 
Due to such reasons, 
the largest system studied
to date using
Coupled-Cluster data is $H_9O_4^+$\cite{waters_nnp} (with absolute error of 0.09 kcal/mol). Furthermore, the resultant neural network model shows limited transferability and the mean absolute error for $H_{13}O_6^+$, using this model is found to be 1.32 kcal/mol\cite{waters_transfer_nnp,BPNN,nnp_review1}. It is precisely this transferability problem that we address in this paper. In contrast to the above listed studies, the machine learning methods developed here result in 0.36 kcal/mol accuracy for the Coupled Cluster level potential energy surface for the complex protonated 21-water cluster (H$_{64}$O$_{21}^+$) system.

\begin{figure} 
\input{Fig2}
\caption{\label{NN-node} Visual illustration of neural networks used to compute $\left\{ \Delta E_{\alpha,r}^{ML}({\bf {\bar x}}) \right\}$ for $r=0$. See Eq. (\ref{eq_graph-ML-main}).}
\end{figure}

\begin{figure}
\input{Fig3}
\caption{\label{NN-edge} Visual illustration of neural networks used to compute $\left\{ \Delta E_{\alpha,r}^{ML}({\bf {\bar x}}) \right\}$ for $r=1$. See Eq. (\ref{eq_graph-ML-main}).}
\end{figure}
Towards this effort, in this paper, we introduce a machine-learning based 
approach to compute multidimensional potential energy surfaces that we show is applicable for much larger systems than previously described. The goals that we address in this paper are as follows: (i) We first ask if it is possible to construct a general formalism for accurate multidimensional potential energy surfaces for moderately sized systems using neural networks that are far less dense and require far less training than implied by Figure \ref{NN-full}. We propose an approach for this purpose using graph-theoretic protocols. (ii) Once problem (i) is addressed reasonably well, we then wonder if it is possible to use the knowledge from these neural networks to then develop a solution for a potential energy surface size that is beyond what can currently be done. (iii) In solving the problem in (ii) we devise a method that we find to be related, in a very fundamental way, to the concept of ``attention'' in large language model methods\cite{trans-attention} and generative AI. 

More specifically, the first key question we attempt to address here, as part of item (i) above, is whether neural networks of lower complexity, that is, neural networks with fewer degrees of freedom corresponding to systems requiring fewer learning cycles and training effort, can be rigorously harnessed together to create a larger network that can be used to solve a larger problem. Towards this, we use our graph-theoretic procedure for molecular fragmentation\cite{fragAIMD,fragAIMD-elbo,fragAIMD-CC,CGAIMD,frag-BSSE-AIMD,frag-AIMD-multitop,fragPBC,fragIJQC-review,frag-PFOA,frag-AIMD-multitop-2,Harry-weighted-graphs,frag-ML-Xiao,frag-QC-Harry,frag-TN-Anup,SSI-Review1-QC-ES-QN,frag-QC-2,frag-ONIOM-frag-reformulation}. 
This approach may be thought of as a generalized convolution neural network\cite{Zhang_cnn,duvenaud2015convolutional} where the convolution is not uniform and is dictated by the spatial proximity of groups within a molecular system and is defined through a graphical network. An example of how such a scheme works is shown in Figures \ref{NN-node} and \ref{NN-edge}, where the neural network in Figure \ref{NN-full} is visually simplified into a family of networks defined by a graphical representation of the molecular system. Note that Figure \ref{NN-node} results in four separate neural networks, and Figure \ref{NN-edge} results in three separate neural networks, each of which is much lower in complexity as compared to Figure \ref{NN-full}. These separate neural networks are then combined using the topology of the graph description. 
In Refs. \onlinecite{frag-ML-Xiao,frag-PFOA}, we have shown that this approach can be used as a graph-theory-based transfer learning protocol where fragment neural networks can lead to accurate potential energy surfaces in higher dimensions\cite{frag-ML-Xiao,frag-PFOA}. 

Then, as part of item (ii) above, we show that the individual models of these graph-based molecular fragments can be incrementally further improved to provide accurate results for systems even larger than those for which they were originally parameterized. Specifically in Ref. \onlinecite{frag-ML-Xiao}, we showed how neural networks written for small water clusters such as protonated/neutral dimers and trimers can be used to compute accurate potentials with errors in the sub-kcal/mol range for a larger solvated Zundel system. Here, as part of item (ii) above, we show how these same sets of neural networks can be further refined
to obtain sub-kcal/mol accuracy for potential energy surfaces for even larger systems such as the protonated 21-water cluster system. Thus, in this paper, for the first time, we provide a systematic strategy to obtain accurate, coupled-cluster-level potentials for large-dimensional systems. Here we illustrate our procedure for a 186-dimensional problem,  the protonated 21-water cluster system. Finally, as part of item (iii) above, we discover that our approach here is deeply related to the concept of ``attention'' in generative AI and large-language models\cite{trans-attention}.

This paper is organized as follows: In Section \ref{graphs}, we introduce our graph-theory approach for accurate post-Hartree-Fock electronic structure potential energies; then in Section \ref{sec_graph}, we show how this approach yields a method for computing electronic energies by introducing a family of neural networks with far reduced computational complexity. Training of the resultant neural networks is carried out using a geometric tessellation of high-dimensional potential energy surface spaces, and this method is described in Appendix \ref{sec_kmeans_nn}. This geometric tessellation approach allows us to train neural networks with only 10\% training data, and the remaining 90\%  data are used for testing and validation. This is a critical aspect of our approach, because, if coupled cluster accuracy is needed (scaling: ${\cal O}(N^{6\cdots7})$), the training data must be as compressed as possible. 
In Section \ref{sec_compare}, we probe the connections between our graph-theoretic methodology and the concept of  ``attention'' in ``transformers'' used in large language models (LLMs) and generative AI. As discussed later, our method has key differences from the well-known Graph Attention Network (GAT) architecture\cite{veličković2018graphattentionnetworks}. While GATs apply learnable attention mechanisms to graph components via neural network–based embeddings, the graph weights in our formalism that lend naturally to an interpretation as attention in our approach, are defined deterministically, based on the topology and connectivity of the graph. Furthermore, the graph weights used here include all simplex ranks, that is, includes, nodes, edges, faces and higher order molecular interactions tempered by their respective geometric weights. As we discuss in Section \ref{graphs}, the weights for each graph component are computed directly from geometric and topological considerations, and no learnable parameters are introduced for the embedding construction, and no optimization is needed to obtain or refine these attention scores. 

Our numerical tests are conducted on water clusters. Water clusters have received wide attention in the literature due to their broad and fundamental significance\cite{Johnson-Jordan-H+-water-cluster-sizes,johnson-jordan-21mer,admp-21mer,admp-21mer-2,Johnson-Jordan-Zundel-JCP,Johnson-Jordan-Zundel-Science,Zundel-scott,Asmis-Zundel,HDMeyer-Zundel-1,Johnson-Zundel-OH-H2O-quantum,JohnsonOH-Science} and also due to challenges presented to both experiment and theory\cite{johnson-jordan-21mer,admp-21mer,admp-21mer-2,Johnson-Jordan-Zundel-JCP,Johnson-Jordan-Zundel-Science,HDMeyer-Zundel-1,Zundel-scott}. The challenges arise from the need for multidimensional quantum-mechanical treatment of the light hydrogen nuclear degrees of freedom (that is span and complexity of the potential energy surface), along with the need for potential energy surfaces with post-Hartree-Fock (electron-correlation) accuracy. Thus, using the approaches developed here, we show that we are able to generate accurate potential energy surfaces for the solvated Zundel ($H_{13}O_{6}^+$) system (51 nuclear dimensions) and these results are discussed in Section \ref{solZ-results}. Additional technical aspects are discussed in Appendix \ref{sec_kmeans_nn} and \ref{sec_descriptor}. The next question addressed in this paper is whether the family of neural networks which was proved to be sufficient for the solvated Zundel system in Section \ref{solZ-results}, remains sufficient for the much larger and more complex protonated 21-water cluster system. The protonated 21-water cluster system, as we discuss has presented a great challenge for experiment and theory due to the large number of quantum nuclear degrees of freedom and associated complexity of potential energy surface evaluations. We find, in Section \ref{solZ-results}, that the potential energy surface for the protonated 21-water cluster is such that the neural networks computed from the solvated Zundel system are not sufficient to capture the surface accurately. We, thus present an incremental tesselation scheme to improve the neural network models that were used for solvated Zundel system which helps achieve sub-kcal/mol accuracy. Conclusions are given in Section \ref{sec_conclusion}. 

\section{Graph theoretic methods to reduce neural network complexity}
\label{graphs}
In order to reduce the complexity of the network in Figure \ref{NN-full} to that in Figures \ref{NN-node} and \ref{NN-edge}, we first discuss a graph-based molecular fragmentation of our chemical system. This graph then becomes the basis for creating our convolution diagrams that leads to a family of decoupled neural networks. A graph, ${\cal G}$, composed of molecular fragments treated as nodes (or vertices), is defined as ${\cal G} \equiv \left\{ {\bf V_0}; {\bf V_1} \right\}$. Here, ${\bf V_0}$ is the set of vertices (fragments within a molecule or a cluster), and ${\bf V_1}$ is the set of edges that capture interactions between these vertices and hence are larger fragments. Once such a graph is defined, higher rank objects, and hence larger fragments and molecular aggregates that represent high-order interactions up to a maximum rank ${\cal R}$, are also specified by the graph as  
\begin{align}
\left\{ {\bf V_r}
\left\vert r=0, \cdots,  {\cal R} \right. \right\} \equiv \left\{ {\bf V_0}, {\bf V_1}, {\bf V_2}, \cdots, {\bf V_r}, \cdots, {\bf V_{\cal R}} \right\}.
\label{G-powerset-R}
\end{align}
The quantity ${\bf V_{r}}$ represents the set of all rank-${\bf r}$ entities in the graph that may be used to capture the interactions between $({\bf r}+1)$-rank molecular fragments.

The post-Hartree-Fock, ``{\em target}'', electronic potential energy surface may  then be written using such a graph-theoretic representation as
\begin{align}
E^{target}({\bf {\bar x}}) =& E^{Ref}({\bf {\bar x}}) +
\sum_{r=0}^{\cal R}  { \sum_{\alpha_r  \in {\bf V}_r}{\cal M}_{\alpha_r, r}^{\cal R} \;  \Delta E_{\alpha_r,r}({\bf {\bar x}})
} 
\label{eq_graph-ML-main}
\end{align}
where
\begin{align}
     \Delta E_{\alpha_r, r}({\bf {\bar x}}) =  E_{\alpha_r, r}^{target}({\bf {\bar x}}) - E_{\alpha_r, r}^{Ref.}({\bf {\bar x}})
     \label{eq_deltaE}
\end{align}
Note here that the variable ${\bf {\bar x}}$ on the left side of Eq. (\ref{eq_graph-ML-main}) represents the full multidimensional molecular coordinate system corresponding to all nuclear degrees of freedom in the problem. On the right side of Eq. (\ref{eq_graph-ML-main}), however, only that portion of the system that is contained within the fragment labeled by the indices $\left({\alpha_r, r}\right)$ needs to be included within the functional dependence of $\Delta E_{\alpha_r,r}({\bf {\bar x}})$. That is $\Delta E_{\alpha_r,r}({\bf {\bar x}}) \equiv \Delta E_{\alpha_r,r}({\bf {\bar x_{\alpha_r,r}}})$. This drastic reduction in dimensionality is similar to that seen in methods such as high-dimensional model representations (HDMR)\cite{alics2001efficient,manzhos2008using,Kolmogorov-HDMR,Kolmogorov-HDMR-2,SOBOL196786,Rabitz-HDMR}. Thus, 
\begin{align}
\Delta E^{target\leftarrow Ref.}({\bf {\bar x}}) &=E^{target}({\bf {\bar x}}) - E^{Ref}({\bf {\bar x}}) \nonumber \\ &=
\sum_{r=0}^{\cal R}  { \sum_{\alpha_r  \in {\bf V}_r}{\cal M}_{\alpha_r, r}^{\cal R} \;  \Delta E_{\alpha_r,r}({\bf {\bar x}_{\alpha_r,r}})
} 
\label{eq_graph-ML-main-delta}
\end{align}
The designation ``$Ref$'', in the equations above  is for some reference electronic structure level such as DFT at nuclear configuration ${\bf {\bar x}}$.
In Eq. (\ref{eq_graph-ML-main}), a high-level energy correction $\Delta E_{\alpha_r, r}({\bf {\bar x}})$ for each fragment or simplex, $\alpha_r \in {\bf V_{r}}$ is added to a reference energy, such as DFT. 
The quantity,  
\begin{align}
{\cal M}_{\alpha_r, r} =  \sum_{m\geq r} {(-1)}^{m+r} p_{\alpha_r}^{r,m}.
\label{eq_m}
\end{align}
represents the total number of times the simplex $\alpha_r \in {\bf V_{r}}$ appears in all simplexes with rank greater than ${\bf r}$. Thus, ${\cal M}_{\alpha_r, r}$ is an over-counting correction and $p_{\alpha_r}^{r,m}$ is the number of times the $\alpha_r^{th}$ rank-$r$ simplex is included within all rank-$m$ simplexes, for ${m\geq r}$.

Equation (\ref{eq_graph-ML-main}) has been widely benchmarked over a series of publications. 
Specifically, we have shown that Eq. (\ref{eq_graph-ML-main}) may be used to construct on-the-fly AIMD trajectories\cite{fragAIMD,fragAIMD-elbo,fragAIMD-CC,CGAIMD,frag-BSSE-AIMD} and potential energy surface calculations\cite{frag-AIMD-multitop,frag-AIMD-multitop-2,Harry-weighted-graphs,frag-TN-Anup} for gas-phase as well as condensed-phase systems\cite{fragPBC,frag-PFOA}.
Both extended Lagrangian as well as Born-Oppenheimer based \textit{ab initio} molecular dynamics simulations can be performed at accuracy comparable to CCSD and MP2 levels of theory with DFT-computational cost.\cite{fragAIMD,fragAIMD-elbo,fragAIMD-CC,CGAIMD}
Hence for the first time, in Refs. \onlinecite{fragAIMD-elbo} and \onlinecite{fragAIMD-CC} we presented Car-Parrinello-style dynamics, but with CCSD accuracy. 
Similarly, we have shown how multiple graphical representations of molecular systems can be used simultaneously to construct accurate potential energy surfaces in agreement with the MP2 and CCSD levels of theory, again at DFT cost.\cite{frag-AIMD-multitop,frag-AIMD-multitop-2,fragIJQC-review,Harry-weighted-graphs,frag-TN-Anup}. 
In Ref. \onlinecite{frag-BSSE-AIMD}, we have also shown that weak interactions (specifically hydrogen bonds) can be accurately captured and efficient approximations to large-basis AIMD trajectories, such as 6-311++G(2df,2pd), can be constructed through computational effort commensurate with much smaller basis set sizes, sets such as 6-31+G(d).
Furthermore, we have also shown in Ref. \onlinecite{fragPBC,frag-PFOA} how condensed-phase simulations on interfaces and liquids may be constructed with hybrid DFT accuracy at gradient-corrected DFT accuracy. We also provide novel approaches to construct reduced circuit depth quantum computing algorithms in Ref. \onlinecite{frag-QC-Harry,frag-QC-2}. Finally, the approach has been demonstrated for hydrogen-bonded systems (such as water clusters and condensed phase systems) and also for covalently bonded biological systems and hydrocarbon where link-atoms\cite{CGAIMD,frag-BSSE-AIMD,frag-PFOA} are included within Eq. (\ref{eq_graph-ML-main}). 
This work is reviewed in Refs. \onlinecite{fragIJQC-review,SSI-Review1-QC-ES-QN,frag-ONIOM-frag-reformulation}.

Within the context of smooth molecular potential energy surfaces\cite{frag-AIMD-multitop,frag-AIMD-multitop-2,fragIJQC-review}, we have shown how multiple graphical topologies can be combined into a single weighted graph formalism. Since the graphs are defined based on the instantaneous molecular geometry, the network encoded within the graphs (that help to capture local and non-local correlation and basis-set effects) may  change when atoms move. This can happen when chemical bonds break or new ones are created. In Refs. \onlinecite{frag-AIMD-multitop,frag-AIMD-multitop-2,fragIJQC-review}, we show that this also happens when protons move in a water wire, such as that appearing in the Gramycidin ion-channel\cite{roux,allen2003gramicidin,pullman1983gramicidin}. This may create singular hops in potential energy surfaces as atoms move. In Refs. \onlinecite{frag-AIMD-multitop,frag-AIMD-multitop-2} we have introduced a weighted graph approach to overcome these singularities. 
Given multiple graphs, $\{{\cal G}_{\beta}\}$, for a  molecular configuration, $\bar{x}$, with each graph yielding a single energy correction function: 
$\left( {\bar{x}}, {\cal G}_\beta
\right) \mapsto \Delta E^{target\leftarrow Ref.}_{\beta,{\cal R}_\beta} \left( {\bar{x}}, {\cal G}_\beta \right)$, based in Eq. (\ref{eq_graph-ML-main-delta}),
the overall energy of the system becomes a probabilistic sum over the set of energies obtained from all the associated set of graphs:
\begin{align}
&\Delta E^{target\leftarrow Ref.}_{\left\{ {\cal G}_\beta \right\}} \left( {\bar{x}} \right) 
\nonumber \\ &= \sum_{\beta}  \sum_{r=0}^{{\cal R}_\beta}  { \sum_{\alpha_r^\beta  \in {\bf V}_r^\beta}\rho_{\beta}\left( {\bar{x}} \right) 
\; {\cal M}_{\alpha_r^\beta, r}^{{\cal R}_\beta} \;  \Delta E_{\alpha_r^\beta,r}({\bf {\bar x}})
} \nonumber \\ &= \sum_{\beta}  \sum_{r=0}^{{\cal R}_\beta}   \sum_{\alpha_r^\beta  \in {\bf V}_r^\beta}\tilde{\rho}_{\beta}\left( {\bar{x}} \right) 
\; \Delta E_{\alpha_r^\beta,r}({\bf {\bar x}})
\label{aveE}
\end{align}
The graph weights $\tilde{\rho}_{\beta}\left( {\bar{x}} \right)$ are obtained variationally in Ref. \onlinecite{frag-AIMD-multitop,frag-AIMD-multitop-2} and through numerical considerations in Ref. \onlinecite{Harry-weighted-graphs}.

In this publication we only consider a single graph topology and we show that a family of neural networks used to train the quantities, $\left\{ \Delta E_{\alpha_r, r}({\bf {\bar x}_{\alpha_r,r}}) \rightarrow \Delta E_{\alpha_r,r}^{ML, 1}({\bf {\bar x}_{\alpha_r,r}}) \right\}$ and obtain an accurate potential energy surface for one system, that is $\left\{ \Delta E_{\alpha_r,r}^{ML, 1}({\bf {\bar x}_{\alpha_r,r}}) \right\} \rightarrow E_{{\text{system,1}}}^{target}({\bf {\bar x}})$, 
\begin{align}
\Delta E_{{\text{system,1}}}^{target \leftarrow Ref.}({\bf {\bar x}}) =& E_{{\text{system,1}}}^{target}({\bf {\bar x}}) - E_{{\text{system,1}}}^{Ref.}({\bf {\bar x}}) \nonumber \\ =&
\sum_{r=0}^{\cal R}  { \sum_{\alpha_r  \in {\bf V}_r}{\cal M}_{\alpha_r, r}^{\cal R} \;  \Delta E^{ML,1 }_{\alpha_r,r}({\bf {\bar x}_{\alpha_r,r}})
} 
\label{eq_graph-ML-main-1}
\end{align}
can be refined in a well-defined and consistent numerical manner to yield an accurate potential for a much larger system, that is, 
\begin{align}
\Delta E_{\alpha_r,r}^{ML, 1}({\bf {\bar x}_{\alpha_r,r}}) \xrightarrow{\text{refine}}  \Delta E_{\alpha_r,r}^{ML, 2}({\bf {\bar x}_{\alpha_r,r}})
\label{refine}
\end{align}
and $\left\{ \Delta E_{\alpha_r,r}^{ML, 2}({\bf {\bar x}_{\alpha_r,r}}) \right\} \rightarrow E_{{\text{system,2}}}^{target}({\bf {\bar x}})$, yielding 
\begin{align}
\Delta E_{{\text{system,2}}}^{target \leftarrow Ref.}({\bf {\bar x}}) =& E_{{\text{system,2}}}^{target}({\bf {\bar x}}) - E_{{\text{system,2}}}^{Ref.}({\bf {\bar x}}) \nonumber \\ =&
\sum_{r=0}^{\cal R}  { \sum_{\alpha_r  \in {\bf V}_r}{\cal M}_{\alpha_r, r}^{\cal R} \;  \Delta E^{ML, 2}_{\alpha_r,r}({\bf {\bar x}_{\alpha_r,r}})
} 
\label{eq_graph-ML-main-2}
\end{align}
Thus, the key idea in this paper is captured in Eq. (\ref{refine}). In this publication, specifically, we refine the graph-theory-based fragment neural networks used to construct the potential energy surface for solvated Zundel to then obtain the surface for the protonated 21-water cluster. 
In this way, we provide an algorithm for growth in learning and our approach looks very similar to the ``transformer'' architecture used in large language models\cite{trans-attention}. This connection is discussed in Section \ref{sec_compare}. To explicitly outline how we build our family of neural network models with reduced network depth as well as breadth, we outline a network projection technique in the section below to compute ML approximations to Eq. (\ref{eq_graph-ML-main}).

\subsection{Neural network projections from the graph-theoretic scheme}
\label{sec_graph}
\begin{figure}
\includegraphics[width=0.85\columnwidth]{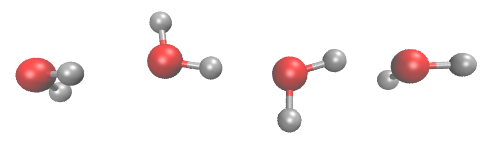}
\usetikzlibrary{shapes.geometric}

\definecolor{purple1}{rgb}{0.6, 0.4, 0.8}
\definecolor{purple2}{rgb}{0, 0.4, 0}
\definecolor{purple3}{rgb}{0, 0, 0.8}
\definecolor{purple4}{rgb}{0, 0, 0}
\definecolor{orange}{rgb}{0.83, 0.4, 0.32}
\definecolor{ggray}{rgb}{0.55, 0.55, 0.55}

\tikzstyle{mynodeV1}=[thick,draw=purple1,fill=purple1!10,circle,minimum size=0.3cm,inner sep=0pt]
\tikzstyle{mynodeV2}=[thick,draw=purple2,fill=purple2!10,circle,minimum size=0.3cm,inner sep=0pt]
\tikzstyle{mynodeV3}=[thick,draw=purple3,fill=purple3!10,circle,minimum size=0.3cm,inner sep=0pt]
\tikzstyle{mynodeV4}=[thick,draw=purple4,fill=purple4!10,circle,minimum size=0.3cm,inner sep=0pt]
\tikzstyle{mynodeH}=[thick,draw=orange,fill=orange!10,circle,minimum size=0.3cm,inner sep=0pt]

\tikzstyle{virtual}=[thick,draw=white,fill=white!30,circle,minimum size=0.3cm,inner sep=0pt]

\tikzstyle{mynodeN1}=[thick,draw=purple1,fill=purple1!30,circle,minimum size=0.6cm,inner sep=0pt]
\tikzstyle{mynodeN2}=[thick,draw=purple2,fill=purple2!30,circle,minimum size=0.6cm,inner sep=0pt]
\tikzstyle{mynodeN3}=[thick,draw=purple3,fill=purple3!30,circle,minimum size=0.6cm,inner sep=0pt]
\tikzstyle{mynodeN4}=[thick,draw=purple4,fill=purple4!30,circle,minimum size=0.6cm,inner sep=0pt]

\begin{tikzpicture}[x=1cm,y=0.75cm,scale=3]

\node[mynodeN1] (N0-1) at (-1.,0) {};
\node[mynodeN2] (N0-2) at (-0.4,0.15) {};
\node[mynodeN3] (N0-3) at (0.2,-0.02) {};
\node[mynodeN4] (N0-4) at (0.8,0.1) {};
\draw[line width=1mm, ggray] (N0-1) -- (N0-2);
\draw[line width=1mm, ggray] (N0-2) -- (N0-3);
\draw[line width=1mm, ggray] (N0-3) -- (N0-4);

\draw[rotate =7] (-0.7,0.2) ellipse (0.44cm and 0.22cm);
\draw[rotate =-10] (-0.12,0.05) ellipse (0.44cm and 0.22cm);
\draw[rotate =5] (0.5,-0.01) ellipse (0.44cm and 0.22cm);

\node[virtual] () at (-1.15,-0.2) {A};
\node[virtual] () at (-0.1,-0.1) {B};
\node[virtual] () at (1.,0) {C    };

\end{tikzpicture}
\caption{\label{Sets-illustration} Illustration of sets and graphs described in Section \ref{sec_graph}. The water wire on top is represented as a graph, with nodes being individual water molecules, and the graph also defines the sets $A$, $B$ and $C$. See Figure \ref{fig_graph_complex} for a graphical representation of systems studied in this paper.}
\end{figure}

\begin{figure}
    \centering
    \subfigure[]{\includegraphics[width=0.48\linewidth]{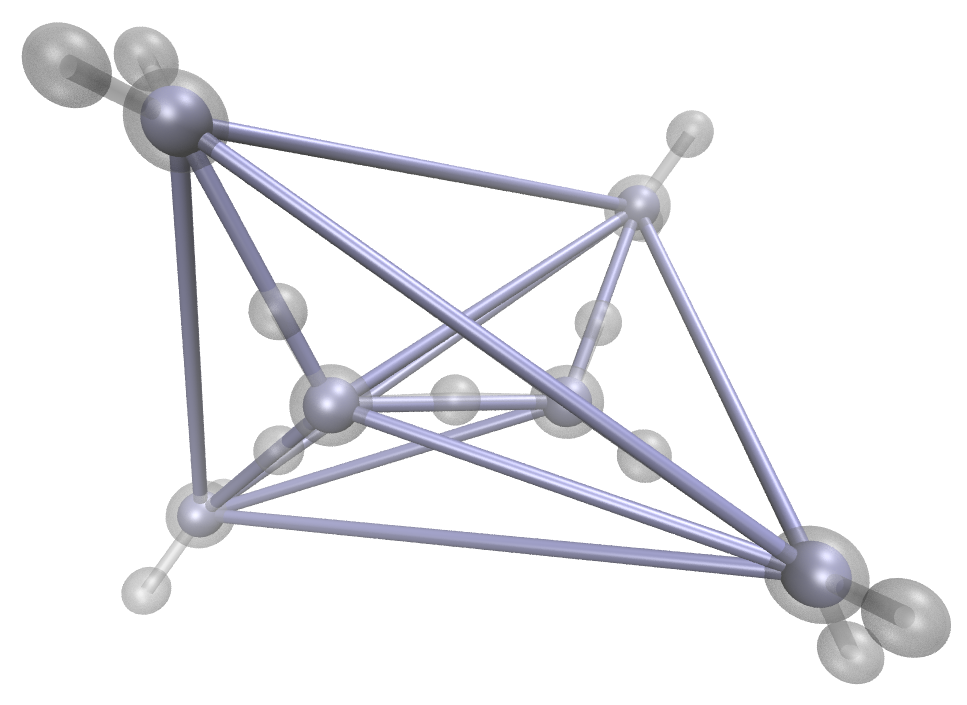}}
    \subfigure[]{\includegraphics[width=0.48\linewidth]{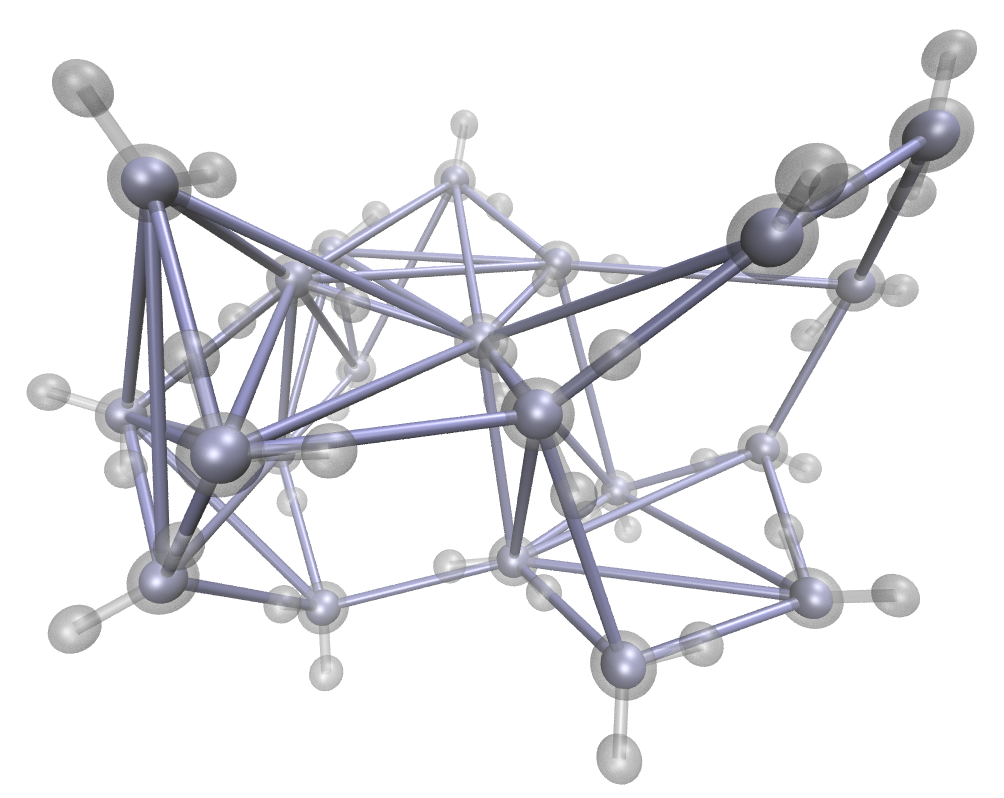}}
    \caption{The graphical complexity for the two systems.}
    \label{fig_graph_complex}
\end{figure}

To arrive at Eq. (\ref{eq_graph-ML-main}), 
we begin with a general graph-based Hilbert space decomposition scheme and illustrate the procedure using Figure \ref{Sets-illustration}.
To illustrate the idea, we begin with a general partition of $R^n$ (the $n$-dimensional real coordinate space). (We may also consider $C^n$, but for simplicity here we partition $R^n$.)
The 
electronic Hilbert space that defines the full molecular system may, for simplicity, be represented using the coordinate representation, $\left\{ \ket{x} \right\}$, and these elements are members of $R^n$. 
Hence, we may then divide the molecular space into regions such as $A$, $B$, etc., as shown in Figure \ref{Sets-illustration}. At this stage, the boundaries of these regions are not considered to be strict, and we may choose these regions based on some prior chemical insight. For example the electrons are confined by attraction to the nuclear framework and one may use this idea to divide the molecular space into chemically meaningful regions. Examples of how such a partitioning can be carried out can be found in Appendix \ref{partitioning}, for both bonded as well as non-bonded systems. 
We may then resolve the basis set resolution of identity, 
using individual projectors, $\left\{ {\cal P}_A, {\cal P}_B, \cdots \right\}$, along with the set-theoretic inclusion-exclusion principle\cite{PIE} as
\begin{align}
    {\bf I} =& {\cal P}_A + {\cal P}_B + {\cal P}_C +... \nonumber
    \\&-{\cal P}_{A \cap B} - {\cal P}_{A \cap C} - {\cal P}_{B \cap C} - ... \nonumber
    \\&+ {\cal P}_{A \cap B \cap C} + ... \nonumber
    \\&...
    \label{IEP-projections}
\end{align}
where
\begin{align}
    {\cal P}_A= \int_A dx|x\rangle \langle x|.
\end{align}
Thus, the regions $A$, $B$, etc. are partitions of a molecular Hilbert space and for the coordinate representation, the spatial decomposition may be seen from Figure \ref{Sets-illustration}. 
In Refs. \onlinecite{frag-ML-Xiao,frag-QC-Harry,frag-QC-2,frag-ONIOM-frag-reformulation}, we have shown that there is another, more general and robust approach to construct such a resolution using a graph theoretic representation of molecular fragmentation, which divides the molecular electronic Hilbert space using nodes, edges, faces, and higher rank simplexes\cite{dey1997267,adams2008introduction,berger1984affine} that constitute a graph. 
This yields an alternative way to represent the molecule's electronic Hilbert space through
a family of projection operators ${\cal P}_{\alpha_r,r}$ to obtain
\cite{frag-ML-Xiao,frag-QC-Harry,frag-QC-2,frag-ONIOM-frag-reformulation}
\begin{align}
{\bf I} = \sum_{r} \sum_{\alpha_r \in {\bf V_r}}  {\cal M}_{\alpha_r,r} {\cal P}_{\alpha_r, r}.
\label{eq_I_graph}
\end{align}
where
\begin{align}
    {\cal P}_{\alpha_r, r} = \int_{\forall x \in \alpha_r} dx \ket{x}\bra{x}
\end{align}
The notation, $\alpha_r$, $r$, and ${\bf V}_r$, are as described at the top of Section \ref{sec_graph}. When the operator in Eq. (\ref{eq_I_graph}) is applied to a molecular system it will decompose the system  
into fragments represented using nodes (local, connected or interacting groups of atoms), edges (node-node interactions and this may include connected, bonded\cite{fragAIMD-elbo,CGAIMD,frag-BSSE-AIMD,fragIJQC-review,frag-PFOA}, or disconnected, and hence through space, non-bonded interactions\cite{fragAIMD,fragAIMD-elbo,fragAIMD-CC,frag-AIMD-multitop,fragPBC,fragIJQC-review,frag-AIMD-multitop-2,Harry-weighted-graphs,frag-QC-Harry,frag-ML-Xiao,frag-TN-Anup}), faces (three-node interactions), tetrahedrons (four-node multi-body interactions), etc, and produce an energy expression\cite{fragAIMD,fragAIMD-elbo,fragAIMD-CC,CGAIMD,frag-BSSE-AIMD,frag-AIMD-multitop,fragPBC,fragIJQC-review,frag-AIMD-multitop-2,Harry-weighted-graphs,frag-QC-Harry,frag-ML-Xiao,frag-TN-Anup,frag-PFOA},
\begin{align}
E^{target}_{graph}({\bf {\bar x}}) =& 
\sum_{r=0}^{\cal R}  { \sum_{\alpha_r  \in {\bf V}_r} {\cal M}_{\alpha_r, r}^{\cal R} \; E_{\alpha_r,r}^{target}({\bf {\bar x}_{\alpha_r,r}})
}
\label{eq_graph-MBE}
\end{align}
and ``$target$'' here is some target level of electronic structure theory, such as CCSD, as before. 
The application of Eq. (\ref{eq_I_graph}) onto the full system hence yields a set of fragment energies $\left\{ E_{\alpha_r,r}^{target}({\bf {\bar x}_{\alpha_r,r}}) \right\}$ that then provide an approximation to the total energy as per Eq. (\ref{eq_graph-MBE}). 
We also note that Eq. (\ref{eq_I_graph}) is simply an inclusion-exclusion principle\cite{PIE} interpretation of the resolution of identity within the graph structure and
\begin{align}
{\bf I} &= \sum_{r} \sum_{\alpha_r \in {\bf V_r}}  {\cal M}_{\alpha_r,r} \int_{\forall {\bf {\bar x}_{\alpha_r,r}} \in \alpha_r} d{\bf {\bar x}_{\alpha_r,r}} \ket{{\bf {\bar x}_{\alpha_r,r}}}\bra{{\bf {\bar x}_{\alpha_r,r}}} \nonumber \\ 
&= \sum_{r} \sum_{\alpha_r \in {\bf V_r}}  {\cal M}_{\alpha_r,r} {\cal P}_{\alpha_r, r}
\end{align}
Connections between Eq. (\ref{eq_graph-MBE}) and the commonly used many-body expansions have been discussed in Refs. \onlinecite{CGAIMD,frag-BSSE-AIMD,Harry-weighted-graphs}. For this to be the case it is required the the graph above be a simplicial complex\cite{dey1997267,adams2008introduction,berger1984affine,Bowyer_Dirichlet,Watson_Voronoi,Voronoi-1,Voronoi-2,cgal,Dirichlet-tessellations-Farin}. 
However, it has been shown in Refs. \onlinecite{Harry-weighted-graphs,fragPBC} that such an approximation converges slower with the maximum truncation order ${\cal R}$ above, as compared to Eqs. (\ref{eq_graph-ML-main}) and (\ref{eq_graph-ML-main-delta}) where this idea is included within an ONIOM-like\cite{oniom} formalism and energy differences are used.

However, as the complexity and size of the system grow, we find that this approach also becomes expensive by needing post-Hartree-Fock energies and gradients for larger and larger clusters obtained from higher rank objects discussed above. That is $\left\{ E_{\alpha_r,r}^{target}({\bf {\bar x}_{\alpha_r,r}}) \right\}$ may be needed for larger values of $r$. 
Furthermore, the number of fragments produced would also grow rapidly (in a combinatorial fashion) as the system size grows since higher rank objects can be formed by all possible combinations of nodes within a distance cutoff. 
In Refs. \onlinecite{frag-ML-Xiao, frag-PFOA},  we provided machine learning approximations for the $\Delta E_{\alpha_r, r}$ values in larger clusters,  to scale down the cost of obtaining potential energy surfaces for large gas-phase and condensed-phase problems. In other words, we use neural network predictions for $\Delta E^{ML}_{\alpha_r,r}({\bf {\bar x}_{\alpha_r,r}})$ to replace the larger rank $\Delta E_{\alpha_r, r}({\bf {\bar x}_{\alpha_r,r}})$ values in Eq. (\ref{eq_deltaE}), as shown in Eqs. (\ref{eq_graph-ML-main-1}) and (\ref{eq_graph-ML-main-2}).  Thus, the approach here and in Refs. \cite{frag-ML-Xiao,frag-PFOA} may also be seen as being related to the $\Delta-ML$ ideas in the literature\cite{bowman-2021-deltaML,delta_ml_intr,delta_ml_2,delta_ml_3,deltaml_2024}, but is in fact couched within the ONIOM formalism as discussed in Ref. \onlinecite{fragAIMD}. As seen from Eqs. (\ref{eq_graph-ML-main-1}) and (\ref{eq_graph-ML-main-2}), the ideas here exploit the graph-based fragmentation technique that helps partition the learning problem.
This is achieved by first assuming the existence of some neural network learning model, ${\cal Q}$, which provides the energy of the system at the ``$target$'' post-Hartree Fock level,
\begin{align}
{\bf {\bar x} }\xrightarrow[]{{\cal Q}} \left[ E^{target}({\bf {\bar x}}) - E^{Ref}({\bf {\bar x}}) \right] \equiv \Delta E_{{\text{system}}}^{target \leftarrow Ref.}({\bf {\bar x}})
\end{align}
Such a neural network, ${\cal Q}$ may be visually represented, for example, by the network in Figure \ref{NN-full}.
We may then simplify ${\cal Q}$ using Eq. (\ref{eq_I_graph}) to obtain
\begin{align}
    {\bf I} \; {\cal Q} &\approx \left[ \sum_{r=0}^{\cal R} \sum_{\alpha_r \in {\bf V_r}}  {\cal M}_{\alpha_r,r} {\cal P}_{\alpha_r, r} \right] \; {\cal Q} \nonumber \\  &= \sum_{r=0}^{\cal R} \sum_{\alpha_r \in {\bf V_r}}  {\cal M}_{\alpha_r,r} {\cal Q}_{\alpha_r, r}.
    \label{eq_proj-Q}
\end{align}
The full system neural network mapping ${\cal Q}$ now has been decomposed into a family of much smaller neural networks $\left\{ {\cal Q}_{\alpha_r,r} \right\}$ that depends only on the corresponding fragment geometry. Consequently, the cost of preparing training samples from fragments becomes much easier, and the complexity of neural networks is also largely reduced.\cite{frag-ML-Xiao, frag-PFOA,frag-ONIOM-frag-reformulation}
The corresponding energy expression, in Eq. (\ref{eq_graph-ML-main}) then becomes a delta machine learning formalism\cite{bowman-2021-deltaML,delta_ml_intr,delta_ml_2,delta_ml_3,deltaml_2024} as the learning algorithm in Eq. (\ref{eq_proj-Q}) is modified to
\begin{align}
E^{target}_{ML}({\bf {\bar x}}) - E^{Ref}({\bf {\bar x}}) =
\sum_{r=0}^{\cal R}  { \sum_{\alpha_r  \in {\bf V}_r}{\cal M}_{\alpha_r, r}^{\cal R} \; \Delta E_{\alpha_r,r}^{ML}({\bf {\bar x}_{\alpha_r,r}}).
} 
\label{eq_graph-ML}
\end{align}
and the ML error is
\begin{align}
E&^{target}_{ML}({\bf {\bar x}}) - E^{target}({\bf {\bar x}}) \nonumber  \\ & = 
\sum_{r=0}^{\cal R}  \sum_{\alpha_r  \in {\bf V}_r}{\cal M}_{\alpha_r, r}^{\cal R} \left( \Delta E_{\alpha_r,r}^{ML}({\bf {\bar x}_{\alpha_r,r}})
 - \Delta E_{\alpha_r,r}({\bf {\bar x}_{\alpha_r,r}}) \right)
 \nonumber \\ &\le 
\sum_{r=0}^{\cal R}  \sum_{\alpha_r  \in {\bf V}_r} \left\vert {\cal M}_{\alpha_r, r}^{\cal R} \right\vert \left\vert \Delta E_{\alpha_r,r}^{ML}({\bf {\bar x}_{\alpha_r,r}})
 - \Delta E_{\alpha_r,r}({\bf {\bar x}_{\alpha_r,r}}) \right\vert
 .
\label{eq_graph-ML-error}
\end{align}
where we have used the Cauchy inequality\cite{Riesznagy} to provide an upper bound for the error expression. 

The simplification afforded by this approximation may be seen from the visuals provided in Figures \ref{NN-full} \ref{NN-node} \ref{NN-edge}, where we have shown how the neural networks for $\left\{ \Delta E_{\alpha_r,r}^{ML}({\bf {\bar x}_{\alpha_r,r}}) \right\}$ can be used to construct approximations to 
$E^{target}_{ML}({\bf {\bar x}})$. In Figure \ref{NN-node}, we show the independent neural networks for each node that contribute to  $\left\{ \Delta E_{\alpha_r,r=0}^{ML}({\bf {\bar x}_{\alpha_r,r=0}}) \right\}$ terms in Eq. (\ref{eq_graph-ML}).
Similarly, in Figure \ref{NN-edge} we show the independent neural networks for each edge and these two figures visually convey the reduced complexity of obtaining these fragment neural networks as can be seen by comparison with Figure \ref{NN-full}. In previous work\cite{frag-ML-Xiao}, we have successfully constructed the neural network potential energy surface with kJ/mol mean absolute error for the solvated Zundel following Eq. (\ref{eq_graph-ML}) by using only 10\% of fragments. In this paper, as seen in Eqs. (\ref{eq_graph-ML-main-2}) we ask if the graph-theoretically decomposed neural networks used to obtain the potential energy surface for $system, 1$, can be refined to obtain the same for  $system, 2$.
Additionally, in the next section, we discuss how the projection operators discussed above connect to the idea of ``attention'' in natural language processing in modern large language models.

\begin{figure} 
{\includegraphics[width=0.98\columnwidth]{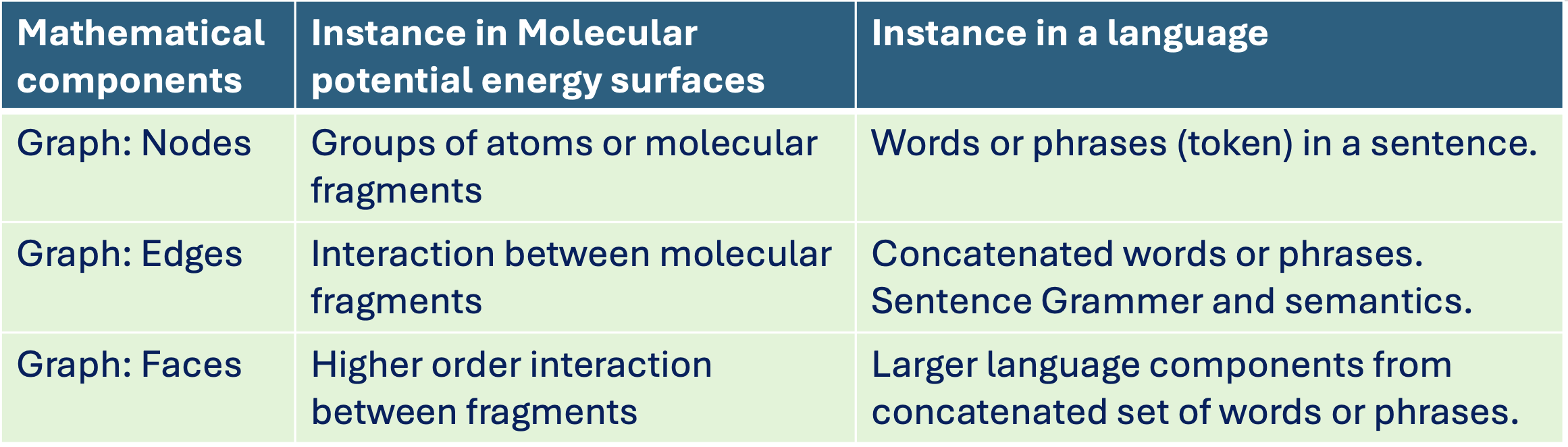}}
\caption{\label{Graphs-PES-LLM-components} The common constructs of graph based depiction of molecular potential energy surfaces used here and elements of a large language model (LLM). }
\end{figure}

\section{The graph abstraction connects molecular potential energy surfaces and natural language processing}\label{sec_compare}
Current methodologies in natural language processing (NLP) involve mapping linguistic components such as words or phrases in  a sentence to abstract vector spaces. 
The semantic relations between these words are then represented as distances 
between these vector representations of the primary linguistic components (which, as stated above could be words or phrases). This technique is used in large Generative AI models. In Figure \ref{Graphs-PES-LLM-components} we show how such an abstraction of natural language can be related to our graph theoretic depiction of molecular structure discussed in the previous section. The key similarity between both ideas is that the neighborhood of a word or phrase defines its context and semantics in natural language, much like the neighborhood of a molecular fragment, as captured from the graphical representation above. defines its stability. In the graph theoretic formalism, the neighborhoods are defined using graph components (nodes, edges, faces, etc, as described within the sets ${\bf V}_r$). 
Each component of a molecular system, such as a fragment, has many different possible energy values arising from multiple possible configurations, and the stability of each of these configurations depends on how the fragment adapts to its surroundings (as dictated by the edges, faces and higher order simplexes in a graph). Likewise, parts of a sentence, such as a word or a phrase, may carry different meanings with some probabilistic distribution, and the intended 
meaning arises given the presence of other language components (such as surrounding words and phrases) and provides a specific context. 
In our work, the weight of each molecular fragment has a critical role that arises completely from geometric considerations as enforced by graphical constraints, and allows us to compute the full system potential energy as a weighted sum of fragment energies as given by 
Eq. (\ref{eq_graph-ML-main}). 

In a complementary fashion, in natural language processing, the attention mechanism within transformers\cite{trans-attention} determines the importance, and contextual meaning of each word by analyzing the relationship between all  possible pairs of words within a sentence or a paragraph. In the transformer, each word or token is represented by an embedding vector. This embedding vector is then linearly transformed to three vectors: {\em query}, {\em key}, and {\em value}. The {\em query} vectors encode querying information about a specific word whereas the {\em key} vectors contain potential maps to other known words. Thus for the $i-th$ word in a sentence, there are three associated vectors to compute the attention for the sentence, namely $Q_i$ for the {\em query}, $K_i$ for the {\em key}, $V_i$ for the {\em value}. The dot product, $\left[ Q_i \cdot K_j \right]$, that is the dot product of the {\em query} vector for the $i$-th word and the {\em key} vector for the $j$-th word in the sentence, produces the relevance between the two words. 
Collecting all possible such inner products provides a cumulative relevance of a certain {\em query} vector, $\left[ Q_i K^T \right]$, where $K^T \equiv \left\{ K_1, K_2, \cdots, \right\}$ contains all {\em key} vectors. By extension, the quantity $\left[ Q K^T \right]$ simply projects all {\em query} vectors onto the {\em key} vector-space. One critical idea introduced in Ref. \onlinecite{trans-attention} which is purportedly responsible\cite{softmax_att,trans-attention} for the major generative AI revolution that we are witnessing at the moment, is to scale $\left[ Q K^T \right]$ using a $softmax$ function given by 
\begin{figure}
    \centering
    \includegraphics[width=\linewidth]{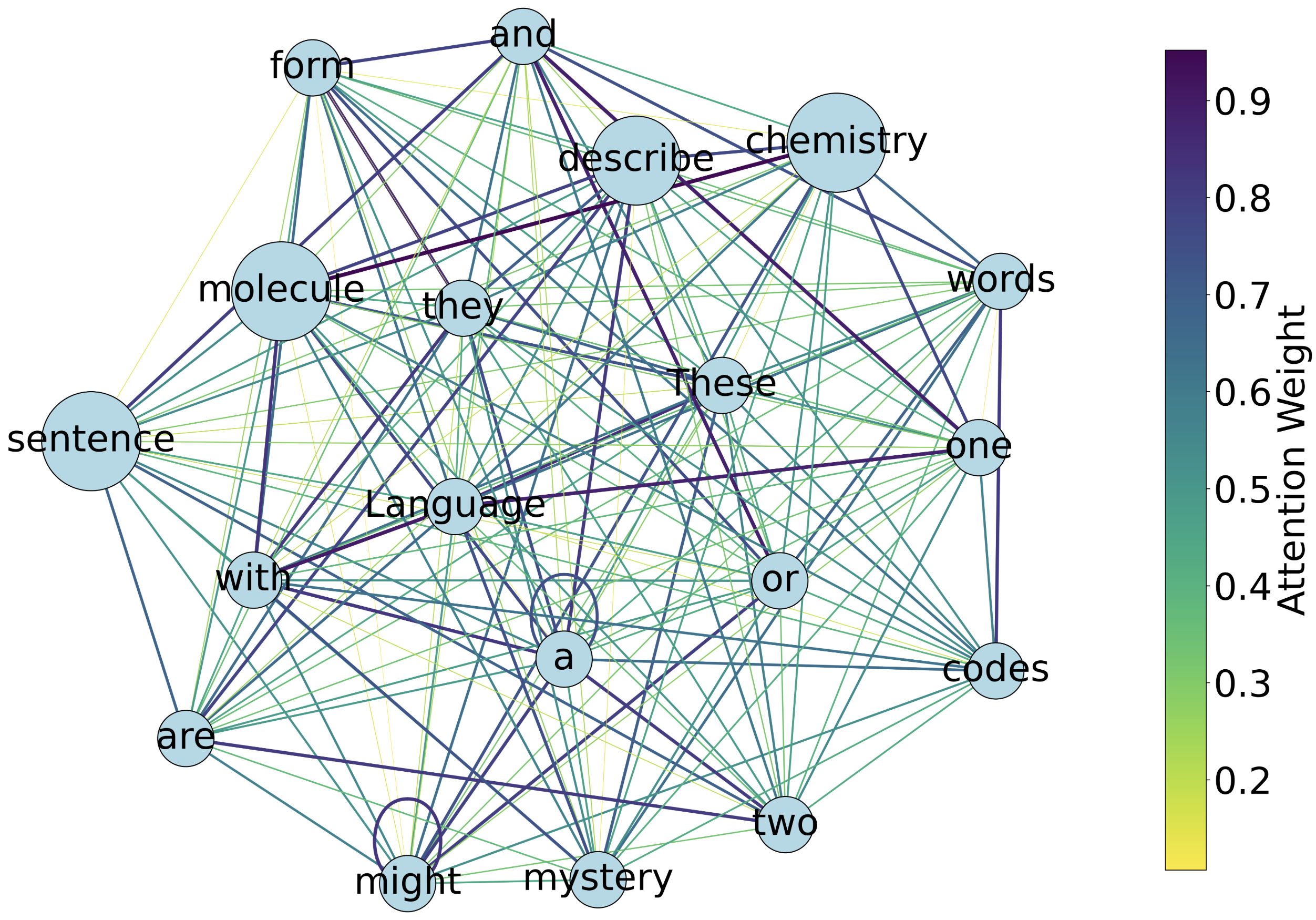}
    \caption{An attention mechanism network developed for the sentence: ``{\em These words might form a sentence, or they might describe a molecule. Language and chemistry are two codes with one mystery.}'' This network is generated using an AI agent in Microsoft Copilot\cite{microsoft2025copilot}. The various steps involved in this process are: Tokenization: The sentence was split into 15 lowercase tokens. Embedding Simulation: Each token was assigned a random 8-dimensional vector (simulating word embeddings). {\em Query}, {\em Key}, and {\em Value} Matrices: These were generated using random weight matrices in this case, for illustrative purposes. Attention Scores: Computed using the dot product of Query and Key matrices, scaled by the square root of the embedding dimension. Softmax: Applied to get attention weights. Attention Output: Calculated as the weighted sum of the Value vectors. Also see Figure \ref{fig:LLM-sentence}, where these algorithmic aspects are compared with those in the current molecular scheme.
}
    \label{fig:LLM-sentence}
\end{figure}
\begin{align}
\left[ softmax\left(\frac{Q K^T}{\sqrt{d}}\right)\right]_{i,j} = \frac{\exp{{Q_i\cdot K_j}/\sqrt{d}}}{\sum_{k}\exp{{Q_i\cdot K_k}/\sqrt{d}}}
\label{softmax-eq}
\end{align}
where $d$ is the dimensionality of each {\em query} vector and {\em key} vector. Thus, the $softmax$ function, in Boltzmann-style, simply weights the most significant {\em key} vectors that contribute to all queries. An illustration of this softmax function is provided in Figure \ref{fig:LLM-sentence}. Here we have presented the sentence: ``{\em These words might form a sentence, or they might describe a molecule. Language and chemistry are two codes with one mystery.}''. The nodes in the graph are the words from the sentence, and the colors of the edges represent the $softmax$ function value for the words connected by the respective edges. The way in which this graph is generated is described in the caption of Figure \ref{fig:LLM-sentence}. For example, in this sentence, the network has allowed us to discover that the words ``{\em molecule}'' and ``{\em language}'' have one of the highest $softmax$ function values. But this network was developed based on only one sentence, and clearly the networks developed using more data can provide us with far greater information on word correlations. 

The next important part of the transformer architecture\cite{trans-attention} is the {\em value} vector space. The {\em key}-{\em value} pairs together transform the queries to contextual information.
Thus\cite{trans-attention}, 
\begin{align}
    Attention(Q,K,V) = softmax\left(\frac{Q K^T}{\sqrt{d}}\right)V
    \label{eq_attention}
\end{align}
or more explicitly, 
\begin{align}
    [Attention(Q,K,V)]_{i,l} = \sum_{j} \frac{\exp{{Q_i\cdot K_j}/\sqrt{d}}}{\sum_{k}\exp{{Q_i\cdot K_k}/\sqrt{d}}} V_{j,l}
    \label{eq_attention-explicity}
\end{align}
where the elements $\left\{ V_{j,l} \right\}$ provide the contextual meaning of the $j$-th {\em key} vector. As a result, an attention network like that created in Figure \ref{fig:LLM-sentence} allows a {\em Query} $\rightarrow$ {\em Value} map based on the context within a sentence and helps decipher the meaning of a sentence. 

In the same manner, our graph-theoretic approach in Eq. (\ref{eq_graph-ML-main})
divides a molecule into {\em simplexes} or rank-$r$ objects and the weights of each such rank-$r$ object is given by ${\cal M}_{\alpha_r, r}^{\cal R}$. These quantities act as weights on the $\Delta E_{\alpha_r,r}({\bf {\bar x}}_{\alpha_r, r})$ {\em values}.

\begin{figure}
    \centering
    
    \subfigure[]{\includegraphics[width=0.48\linewidth]{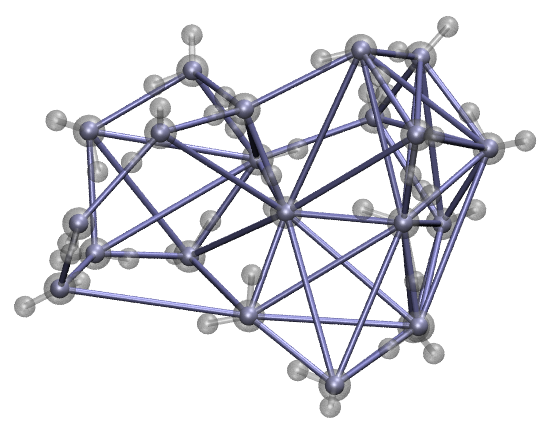}}
    \subfigure[]{\includegraphics[width=0.48\linewidth]{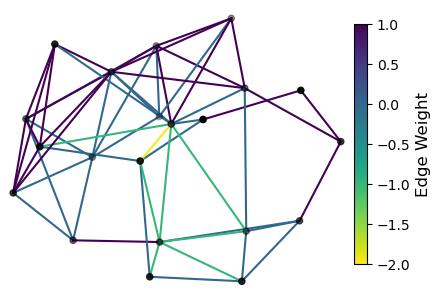}}
    \subfigure[]{\includegraphics[width=0.48\linewidth]{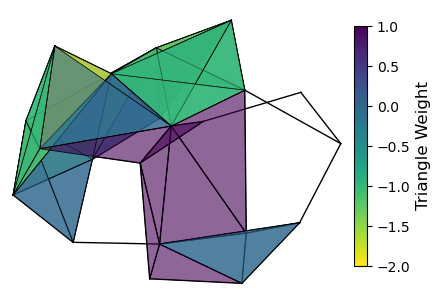}}
    \subfigure[]{\includegraphics[width=0.48\linewidth]{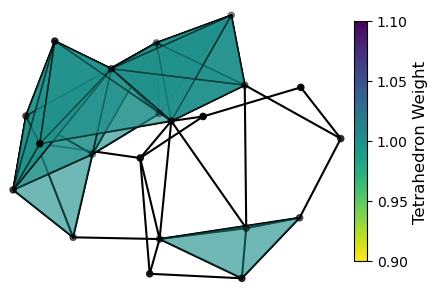}}
    \caption{For comparison with Figure \ref{fig:LLM-sentence}. Figure (a): Graph rendering of the protonated 21-water cluster. Figure (b): For the molecule in Figure (a) edge colors now represent ${\cal M}_{\alpha_r, r}^{\cal R}$ values. Compared with Figure \ref{fig:LLM-sentence}. Thus, these edges weights can be negative here unlike in Eq. (\ref{softmax-eq}). Furthermore, Eq. (\ref{eq_graph-ML-main}) allows higher rank objects too (and not just edges). 
    Similar to (b), the graph theoretic ${\cal M}_{\alpha_r, r}^{\cal R}$ values exists for all higher ranked objects, such as triangles in (c) and tetrahedrons in (d), with ${\cal M}_{\alpha_r, r}^{\cal R}$-weights are presented here. The white space in the graph is due to no triangle or tetrahedron formed at those positions as the nodes are not all mutually connected. The overall molecular energy (or stability) is given by the weighted sum of simplex energy values, $\Delta E_{\alpha_r,r}({\bf {\bar x}}_{\alpha_r, r})$ with corresponding simplex weights ${\cal M}_{\alpha_r, r}^{\cal R}$. Compare Eqs. (\ref{eq_graph-ML-main}) and (\ref{eq_attention-explicity}).
    \label{fig:LLM-molecule}}
\end{figure}

Based on the above discussion, we may make the connection that the $softmax$ function has a similar role as ${\cal M}_{\alpha_r, r}^{\cal R}$ in Eq. (\ref{eq_graph-ML-main}) where ${\cal M}_{\alpha_r, r}^{\cal R}$ also provides the importance of the $\alpha_r$-th, $r$-rank fragment (or simplex). Hence, we may state that
\begin{align}
{\text {Importance}} \equiv \left[ softmax\left(\frac{Q K^T}{\sqrt{d}}\right)\right]_{i,j} \leftrightarrow {\cal M}_{\alpha_r, r}^{\cal R}
\label{analogy-1}
\end{align}
Importantly, both terms in Eq. (\ref{analogy-1}) arise from purely geometric considerations. Geometry in the context of molecular system is the spatial spread of fragments, whereas geometry in the context of natural language processing is connectivity or conjunction of words or phrases.

Thus, it appears that Eqs. (\ref{eq_attention-explicity}) and (\ref{eq_graph-ML-main}) carry out importance-to-value projections 
where
\begin{align}
{\text {Intrinsic Value}} \equiv V_{j,l} \leftrightarrow \Delta E_{\alpha_r,r}({\bf {\bar x}})
\end{align}
As a comparison to drive home the message, we also provide Figure \ref{fig:LLM-molecule} which may be compared with Figure \ref{fig:LLM-sentence}. 

This connection can be made even more explicit by rewriting attention in bra-ket notation. In the absence of the softmax function, this map essentially writes the {\em query} vectors in the bi-orthogonal $\ket{K}\bra{V}$ basis and thus, in the absence of softmax, 
\begin{align}
    [Attention(Q,K,V)]_{\omega,\upsilon} = \sum_\kappa\bra{Q_{\omega}}\ket{K_\kappa}\bra{V_\upsilon}
    \label{eq_attention-braket}
\end{align}
and in the presence of softmax
\begin{align}
    [Attention(Q,K,V)]_{\omega,\upsilon} = \sum_{\kappa} f(\bra{Q_{\omega}}\ket{K_\kappa}) \bra{V_\upsilon}
    \label{eq_attention-braket-softmax}
\end{align}
Thus the projections, $\bra{Q_{\omega}}\ket{K_\kappa}$ or $f(\bra{Q_{\omega}}\ket{K_\kappa})$, act in an analogous fashion as the projections ${\cal P}_{\alpha_r, r}.$ in Eq. (\ref{eq_I_graph}).

\begin{figure} 
{\includegraphics[width=0.98\columnwidth]{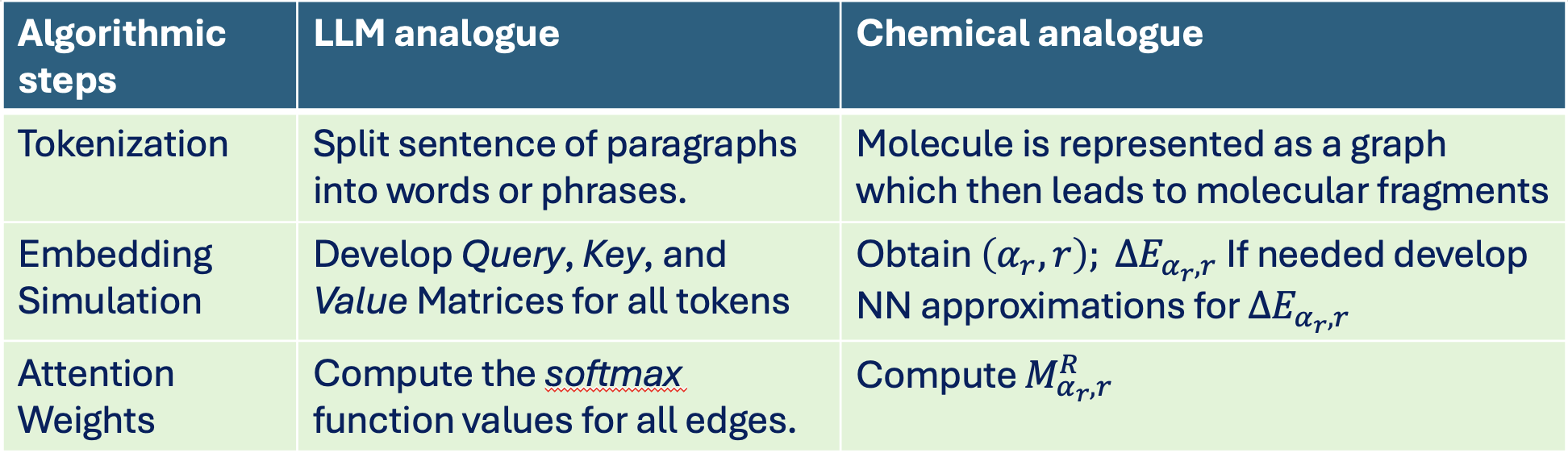}}
\caption{\label{Graphs-PES-LLM-process} The common algorithmic aspects involved in the NLP process and the graph theoretic process presented here. }
\end{figure}

To complement the discussion above, we present the various stages of LLM generation and graph-theoretic decomposition in Figure \ref{Graphs-PES-LLM-process} to make the connections between processes involved more explicit. A few distinctions between challenges that determine the complexity of chemical potential energy surfaces and creation of LLMs are discussed in Appendix \ref{sec_disctinction}.

As an illustration of our approach, we will show how simplex neural networks that were originally defined for obtaining accurate potential energy surfaces for the solvated Zundel system can now be generalized to obtain results for the much larger protonated 21 water cluster. Thus, this paper provides the first sub-kcal/mol level neural network description of larger protonated water clusters.

\section{potential energy surfaces for solvated Zundel using the family of neural networks introduced in Section  \ref{graphs}}
\label{solZ-results}

The solvated Zundel cation has a fundamental role in the study of proton transfer in water, as these contain two of the most critical substructures,  
the Eigen and Zundel moieties of water\cite{zundel2012hydration,cptuckerman3, h+oh-solv, schmittvothprotontransport1999, schmitt1998multistate}. The interconversion between the Eigen and Zundel systems offers valuable insight into proton transfer in a wide range of biological, atmospheric, material, and condensed phase systems\cite{HDMeyer-Zundel-1,protonwire1,atmosph-clusters1,atmosph-clusters2,pomesroux2,protonwire3,protonwire4,teeter,bio-clusters2,lipscombpeek,turowlett,Scott-proj,Harry-weighted-graphs}. 
However, due to the highly polarized nature of the Eigen and Zundel cations, 
accurately modeling these systems typically requires computationally prohibitive post-Hartree-Fock calculations\cite{H5O2+XCcomp,Schaefer:94,DFT-vDW-Michaelides,Truhlar-functional-review,yang-dft-review} along with detailed description of multi-dimensional quantum nuclear effects. This makes the construction of accurate potential energy surfaces and the associated quantum dynamical study extremely challenging\cite{Johnson-Jordan-Zundel-Science,Johnson-Jordan-Zundel-JCP,HDMeyer-Zundel-1}. In this paper, we choose the solvated Zundel as a foundational test case to (a) develop neural networks trained on molecular fragments of solvated Zundel that are then combined as per Eq. (\ref{eq_graph-ML-main-1}) to reliably reproduce the full system potential energies of solvated Zundel, and (b)  since the solvated Zundel system serves as a fundamental model for studying proton migration, we then probe whether the successful recovery of its potential energy surface provides a reliable starting point to use the same family of fragment neural network models to study more complicated 
and larger systems, such as the protonated 21 water clusters.

To construct a set of fragment neural networks for the solvated Zundel system, we follow a procedure similar to that in 
Ref. \onlinecite{frag-ML-Xiao}. Briefly, we begin by preparing an AIMD trajectory for the solvated Zundel system, by applying graph-theoretic fragmentation to obtain fragments from each structure, and create a data bank for each fragment type. These AIMD geometries use the energy expressions in Eq. (\ref{eq_graph-ML-main})
along with gradients  discussed in Refs. \onlinecite{fragAIMD,fragAIMD-elbo,fragAIMD-CC,CGAIMD,frag-BSSE-AIMD,frag-AIMD-multitop,fragPBC,fragIJQC-review,frag-AIMD-multitop-2,Harry-weighted-graphs,frag-TN-Anup,frag-PFOA}. The quantity $E^{Ref}$ is chosen as DFT with 6-31++g(d,p) basis, and  the $E^{target}$ level is CCSD with 6-31++g(d,p) basis.\cite{fragAIMD-CC} We considered at most rank 3 simplexes (${\cal R} = 3$ in Eq. (\ref{eq_graph-ML-main}) in these AIMD trajectories, and hence four-body interactions between the nodes) with maximum oxygen-oxygen distance (which defines the maximum edge length to define the graph) values set at 7.5\AA\ \cite{Harry-weighted-graphs}. Additionally, the hydrogen-oxygen distance cutoff value is set to be 1.4\AA\;  and defines the chemical bonding environment used to obtain the graphs and fragments. These choices are based on our previous studies\cite{Harry-weighted-graphs,fragAIMD-CC,fragAIMD,fragAIMD-elbo}, and help determine the graph and associated fragments for any given structure. Our library is a in-house Python code that uses NumPy and SciPy functionalities. The number of fragments of each kind obtained during
these trajectories are summarized in Table \ref{tab_num_of_geo}.

We then apply a mini-batch-k-means clustering algorithm, which was discussed in Ref. \onlinecite{frag-ML-Xiao} and is summarized further below in this section and in more detail in Appendix \ref{sec_kmeans_nn}. This helps us select 10\% of the data for each type of fragment to be used for training, and construct separate neural network models to obtained the corresponding fragment energy surfaces. 
Since there are multiple protons present inside each fragment, we include permutation symmetry into our treatment for all fragments, 
as explained in Appendix \ref{sec_descriptor}. We have also adopted Gaussian functions as the activation functions to improve the smoothness and differentiability of the model predictions. 
At the end of this training cycle, we determine the fragment energy surfaces, $\Delta E^{ML,1 }_{\alpha_r,r}({\bf {\bar x}_{\alpha_r,r}})$ needed in Eq. (\ref{eq_graph-ML-main-1}). This helps develop a model of the full potential energy surface as indicated on the left side of Eq. (\ref{eq_graph-ML-main}).

\begin{table}[h!]
\centering
    \begin{tabular*}{\columnwidth}{@{\extracolsep{\fill}}l|cc|c}
\hline  \hline
Fragment&	
\multicolumn{2}{c|}{$(H_2O)_6H^+$ data\footnote{Number of fragments from the $(H_2O)_6H^+$ trajectory.}} &	
$(H_2O)_{21}H^+$ data\footnote{Number of fragments from the $(H_2O)_{21}H^+$ trajectory.}  \\ \hline 
 & Full data & Training dataset & Target data \\
& &  size\footnote{Neural network training data size. These data points are obtained from a multi-dimensional k-means clustering algorithm\cite{frag-ML-Xiao,kmeans} of the Full $(H_2O)_{6}H^+$ data and represent 10\% of the data set.} & \\
\hline 
$H_2O$ &	38910&	3891 & 559877\\ 
$H_3O^+$ &	17046&	1704 & 34297\\
$H_4O_2$&	62380&	6238 & 1356707\\
$H_5O_2^+$&	77510&	7751 & 190727\\
$H_6O_2^{++}$\footnote{Very low probability doubly protonated fragments found in the larger 21-water cluster but not in solvated Zundel.} &	0&	0 & 99\\
$H_6O_3$ &	46940&	4694 & 1014960\\
$H_7O_3^+$ &	139580&	13958 & 257260\\
$H_8O_3^{++}$\footnotemark[4] &	0&	0 & 256\\
$H_8O_4$ &	15750&	1575 & 253889\\
$H_9O_4^+$ &	124140&	12414 & 108123\\
$H_{10}O_4^{++}$\footnotemark[4] &	0& 0 &	71\\
\hline  \hline
\end{tabular*}
\caption{Number of fragments in the solvated Zundel library, the corresponding number of data points in a training library and the library size for the protonated 21 water cluster system used for testing predictions.}
\label{tab_num_of_geo}
\end{table}

\begin{figure}
    \centering
    \subfigure[Using graph-based NN models for solvated-Zundel, ${\cal R}$=0]{\includegraphics[width=0.485\columnwidth]{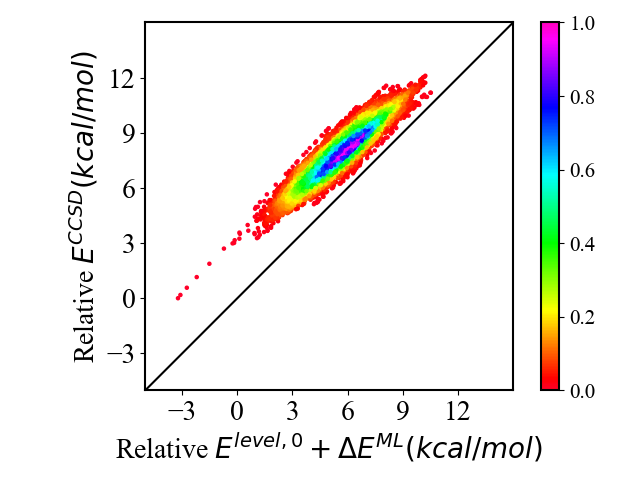}}
    \subfigure[Using graph-based NN models for solvated-Zundel, ${\cal R}$=1]{\includegraphics[width=0.485\columnwidth]{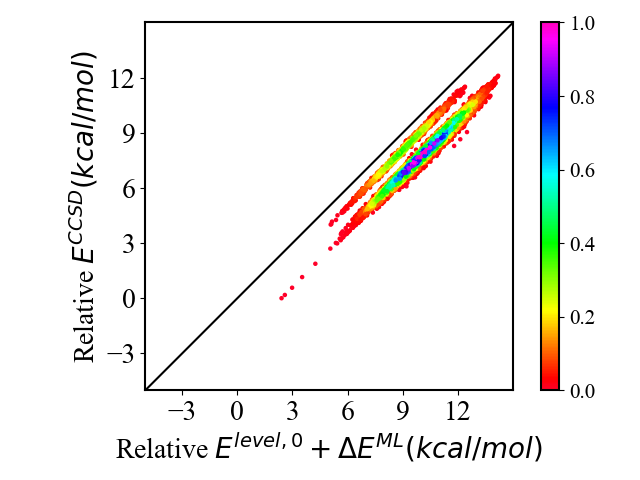}}
    \subfigure[Using graph-based NN models for solvated-Zundel, ${\cal R}$=2]{\includegraphics[width=0.485\columnwidth]{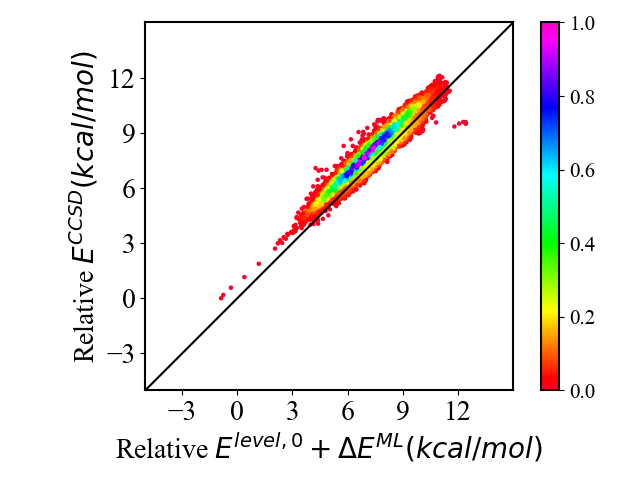}}
    \subfigure[Using graph-based NN models for solvated-Zundel, ${\cal R}$=3]{\includegraphics[width=0.485\columnwidth]{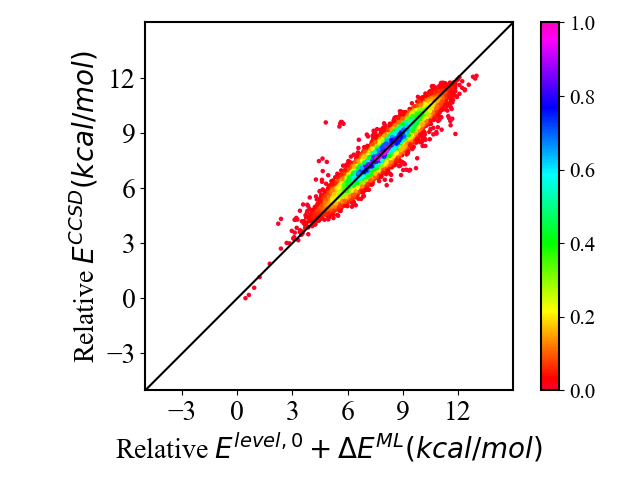}}
    \caption{The solvated Zundel potential energy absolute errors. All energies are plotted relative to the minimum potential energy of solvated Zundel in the trajectory.}
    \label{fig_sz-sz}
\end{figure}

In Figure \ref{fig_sz-sz}, we demonstrate the accuracy of this ML architectures for full system potential energy surface of the solvated Zundel problem  (left side of Eq. (\ref{eq_graph-ML-main-1})). The axis labels in Figure \ref{fig_sz-sz}, are shifted relative to the lowest energy configuration found in the solvated Zundel trajectory. The horizontal axis represents the ML result (right side of Eq. (\ref{eq_graph-ML-main-1})) and the vertical axis represents the full CCSD direct calculation (${\cal O}(N^6)$ scaling of cost for vertical axes). The error (spread of the distribution away from the solid diagonal line) distribution is also shown in Figure \ref{fig_sz-sz}. 
Clearly, the neural networks have been well-trained for solvated Zundel with 10\% training at rank 3. We arrive at a full system MAE of only 0.36 kcal/mol for the potential energies for the solvated Zundel trajectory.

As noted above, the $\Delta E^{ML,1 }_{\alpha_r,r}({\bf {\bar x}_{\alpha_r,r}})$  needed to compute the left side in Eq. (\ref{eq_graph-ML-main-1}) are obtained through a mini-batch k-means procedure. 
We begin with the training a set of neural networks following Appendix \ref{sec_kmeans_nn} for the fragments in Table \ref{tab_num_of_geo} using the solvated Zundel data. To develop these neural networks, 
we generate training sets using the k-means clustering\cite{kmeans,frag-ML-Xiao} method. We perform k-means clustering\cite{kmeans} calculations on the database of fragment structures obtained from solvated Zundel AIMD trajectory to create a representative training set for the fragments in the solvated Zundel system. 
Using k-means we divide or 
tessellate the data space for each fragment type in Table  \ref{tab_num_of_geo} into regions, and these regions are called clusters. Each cluster is represented by a centroid that is computed as the arithmetic mean of all data points in that specific cluster. 
This primitive training set (that is the set of centroids obtained from k-means) is chosen to be 10\% in size (as compared to the full data set for the solvated Zundel system in Table \ref{tab_num_of_geo}) and is used to construct neural networks and fragment energies needed in Eq. (\ref{eq_graph-ML}). 
Examples of such neural networks are shown in Figures \ref{NN-node} and \ref{NN-edge}. 
These neural networks are used to then compute energy values for the data from the full trajectory for solvated Zundel as well as the 21-mer 
with fragment energy errors for the target level of the theory given in Table \ref{tab_initial_error} and the distribution of errors for full system geometries for both systems being given in Figure \ref{fig_init_compare}. 

\subsection{The challenge of transferring fragment neural networks across molecular systems}
\label{sec_sys_diff}
Next, we assess the quality of predictions from these fragment neural network models in obtaining  the protonated 21 water clusters. 
Figure \ref{fig_graph_complex} is used to show the level of complexity of the graphical representation as one moves from the solvated Zundel system to the protonated 21-water cluster. Specifically, assuming $O(N^{6})$ complexity for the CCSD electronic structure values used here, each electronic structure energy for the system in Figure \ref{fig_graph_complex}(b) is 1762 times more expensive to compute than for that in Figure \ref{fig_graph_complex}(a). When one  accounts for the fact that the system in Figure \ref{fig_graph_complex}(b) has 186 independent degrees of freedom, whereas system in Figure \ref{fig_graph_complex}(a) contains {\em only} 51 degrees of freedom, the prohibitive exponential complexity associated with training for the potential energy surface for the system in Figure \ref{fig_graph_complex}(b) is apparent.

\begin{figure}
    \centering
    \subfigure[$H_{13}O_{6}^+$]{\includegraphics[width=0.48\columnwidth]{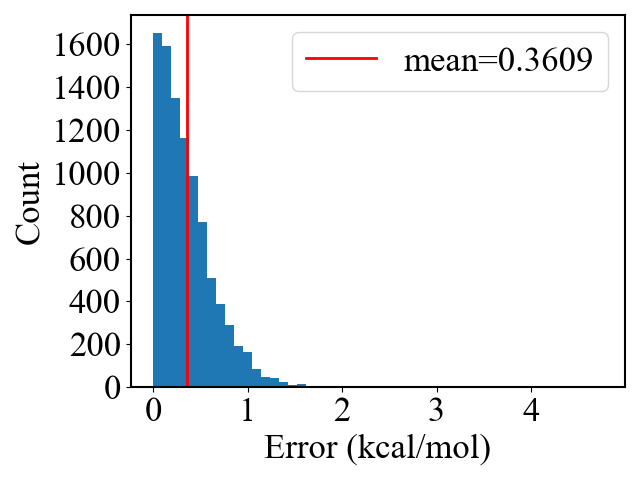}}
    \subfigure[$H_{42}O_{21}^+$]{\includegraphics[width=0.48\columnwidth]{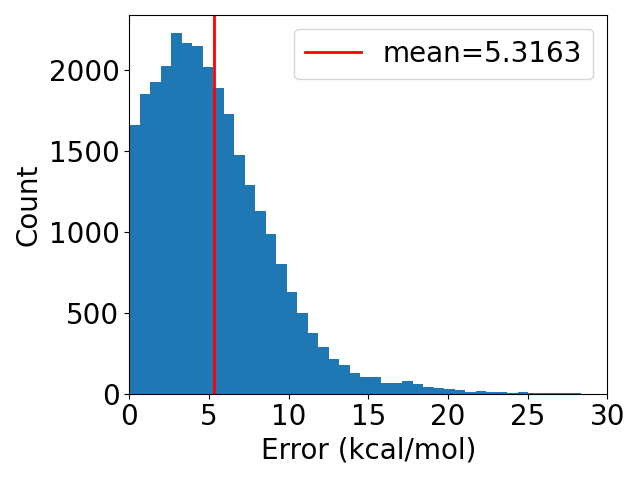}}
    \caption{Subfigure (a) complements Figure \ref{fig_sz-sz} subfigure (b) complements Figure \ref{fig_final-main}. The absolute error distribution (blue) and MAE (red) for solvated Zundel (a) and protonated 21 water cluster (b) full system potential energies. 
     }
    \label{fig_init_compare}
\end{figure}

\begin{figure}
    \centering
    \subfigure[Using graph-based NN models for solvated-Zundel, ${\cal R}$=0]{\includegraphics[width=0.485\columnwidth]{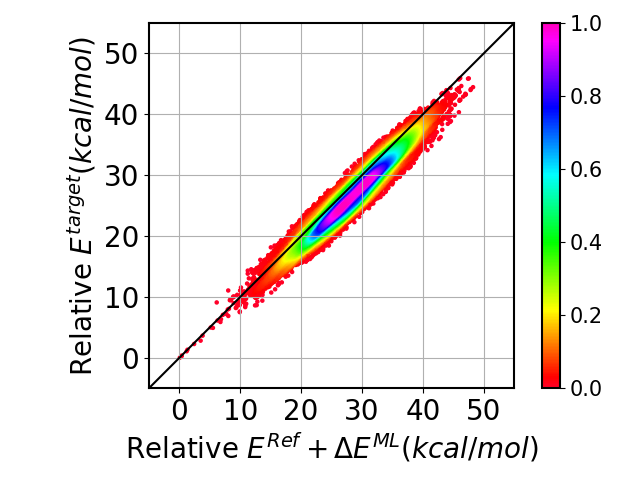}}
    \subfigure[Using graph-based NN models for solvated-Zundel, ${\cal R}$=1]{\includegraphics[width=0.485\columnwidth]{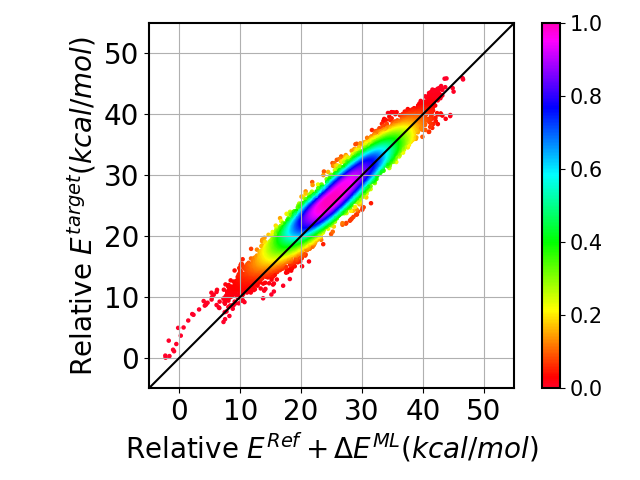}}
    \subfigure[Using graph-based NN models for solvated-Zundel, ${\cal R}$=2]{\includegraphics[width=0.485\columnwidth]{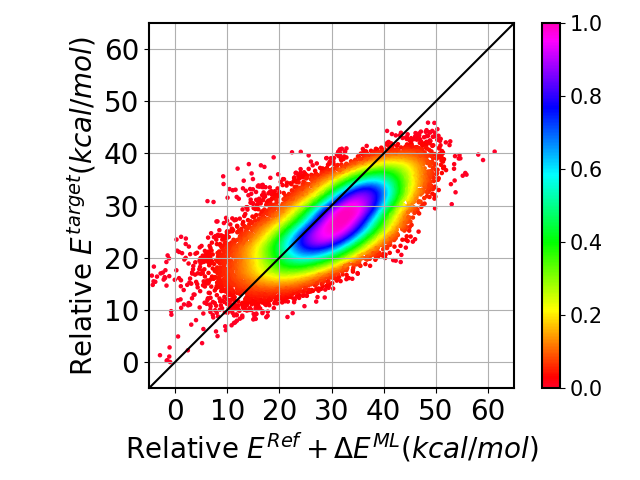}}
    \subfigure[Using graph-based NN models for solvated-Zundel, ${\cal R}$=3]{\includegraphics[width=0.485\columnwidth]{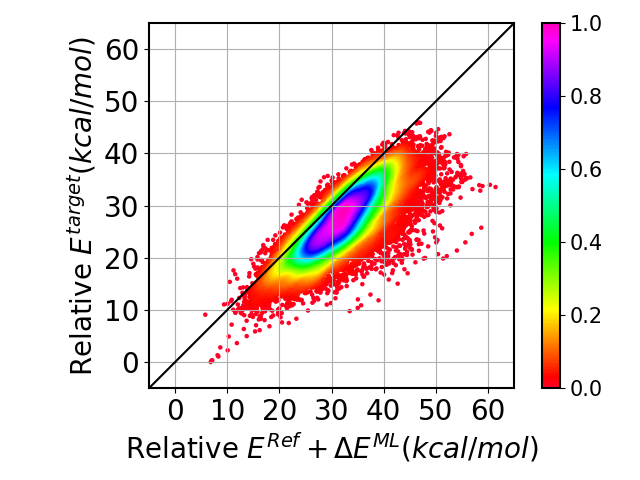}}
    \caption{Similar to Figure \ref{fig_sz-sz}, for the protonated 21 water cluster energies from NN models for the solvated Zundel system. Clearly the error increases with increasing rank, since the potential energy surface space for the larger clusters in the 21-mer is not well-represented in the original solvated Zundel data.}
    \label{fig_final-main}
\end{figure}

\begin{figure}[tbp]
    \centering
    {\includegraphics[width=0.6\columnwidth]{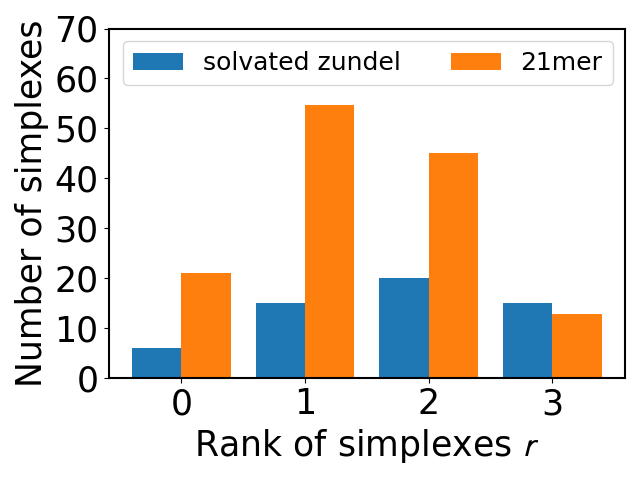}}
    \caption{The average number of simplexes for each rank obtained from the library of solvated Zundel and protonated 21 water cluster structures in the AIMD trajectories. The solvated Zundel trajectory library contains 9326 different geometries created using edge length cutoff 7.5\AA. The $r=3$ fragment calculations are roughly $4^6 \equiv 4096$ times more expensive as compared to the $r=0$ fragment calculations for CCSD accuracy. The protonated 21 water cluster (21-mer) AIMD library has 28294 geometries but with edge cutoff 4.5\AA, so the average number of rank 3 simplexes is fewer. Simplexes with zero weights are not presented in the figure.}
    \label{SZ-21-rank-number}
\end{figure}
As before, to gauge the accuracy of the solvated Zundel neural network fragments in being able to produce accurate protonated 21-water cluster energies, we first develop a library of protonated 21-water cluster structures. Thus, in addition to the solvated Zundel AIMD trajectory noted earlier, we also generate a new trajectory for the protonated 21 water cluster including 28294 frames. The 
corresponding library of structures include 
9326 geometry frames from the solvated Zundel described in the previous section and 28294 frames from the 
protonated 21 water cluster trajectory. These AIMD geometries (as for the solvated Zundel trajectories previous noted) are obtained using the graph theoretic fragmentation energy expressions (Eq. (\ref{eq_graph-ML-main}))
along with gradients as discussed in Refs. \onlinecite{fragAIMD,fragAIMD-elbo,fragAIMD-CC,CGAIMD,frag-BSSE-AIMD,frag-AIMD-multitop,fragPBC,fragIJQC-review,frag-AIMD-multitop-2,Harry-weighted-graphs,frag-TN-Anup,frag-PFOA}). The levels of theory, basis, maximum graph rank are the same as the in case of solvated Zundel. 
The maximum edge length to define the graph is set at 7.5\AA\; for the solvated Zundel and 4.5\AA\; for the protonated 21 water cluster respectively\cite{Harry-weighted-graphs}. Additionally, as before, the hydrogen-oxygen distance cutoff value is set to be 1.4\AA\;  and defines the chemical bonding environment used to obtain the graphs and fragments. These choices are based on our previous studies\cite{Harry-weighted-graphs,fragAIMD-CC,fragAIMD,fragAIMD-elbo}. The number of fragments of each kind obtained during both trajectories is summarized in Table \ref{tab_num_of_geo}. 

However, when the fragment neural network models trained from data obtained in the solvated Zundel AIMD trajectory 
are used to compute the potential energy surface for the protonated 21-water cluster,  an increase in error is noted for larger values of rank, ${\cal R}$, as seen in Figure \ref{fig_final-main}(a)-(d). 
Also see Figure \ref{fig_init_compare}.
Unlike the solvated Zundel, the full system CCSD energy calculations are prohibitive. Hence, the vertical axis in Figure \ref{fig_final-main}(a)-(d) are $E^{target}$ from Eq. (\ref{eq_graph-ML-main}) and the error is essentially the error in ML predictions as given by Eq. (\ref{eq_graph-ML-error}). 
As can be seen from Figure \ref{fig_final-main}(a)-(d), when the primitive solvated Zundel neural network models ($\Delta E^{ML,1 }_{\alpha_r,r}({\bf {\bar x}_{\alpha_r,r}})$ in Eq. (\ref{eq_graph-ML-main-1})) are applied the errors get worse with increasing rank. 
We drill down into the reasons behind the errors in Figure \ref{fig_final-main}(a)-(d) in Section \ref{sec_sys_diff}. 

But there are two fundamental reasons for these differences: 

First, as the rank increases, the number of fragments as well as the number of $\Delta E_{\alpha_r,r}^{ML}({\bf {\bar x}_{\alpha_r,r}})$ terms increases in Eq. (\ref{eq_graph-ML-main}). This can be seen from Figure (\ref{SZ-21-rank-number}). Additionally, the error is bounded by Eq. (\ref{eq_graph-ML-error}), and the bound grows with the number of fragments of a certain size. 
In Table \ref{tab_initial_error}, we also present the quantity relative significance of all rank-$r$ fragments in Eqs. (\ref{eq_graph-ML-main}). Towards this we first introduce, 
\begin{align}
  \omega_{r,{\cal R}} = \frac{\sum_{\alpha_r  \in {\bf V}_r}\left\vert {\cal M}_{\alpha_r, r}^{\cal R} \right\vert}{\sum_{r'=0}^{\cal R}  { \sum_{\alpha_r  \in {\bf V}_{r'}}\left\vert {\cal M}_{\alpha_r, r'}^{\cal R} \right\vert }}
\end{align}
which is the relative significance of all rank-$r$ fragments in Eqs. (\ref{eq_graph-ML-main}) and (\ref{eq_graph-ML}). Additionally, a specific kind of fragment of rank-$r$, $f_r \subseteq {\bf V}_r$, may have significance given by
\begin{align}
  \omega_{f_r,{\cal R}} = \frac{\sum_{\alpha_r  \in f_r}\left\vert {\cal M}_{\alpha_r, r}^{\cal R} \right\vert}{\sum_{r'=0}^{\cal R}  { \sum_{\alpha_r  \in {\bf V}_{r'}}\left\vert {\cal M}_{\alpha_r, r'}^{\cal R} \right\vert }}
\label{eq_weight}
\end{align}
This last quantity gauges the significance of each fragment in the overall $target$ energy and is presented in Table \ref{tab_initial_error}. We clearly see that some of the larger sized fragments are more significant in the 21-water cluster than they were in the solvated Zundel system which could be expected to have an important role on the overall error.
Hence, the overall full system error accumulates and makes the quality of predictions unstable. 

Second, higher rank simplexes introduce larger fragments into the graph representation. These large fragments have more degrees of freedom and represent more complicated interactions within the molecular systems. More importantly, as we will discuss later in Appendix \ref{sec_H_path} and in Section \ref{sec_sys_diff}, large fragments from the 21-mer occupy substantially different regions of the configuration space compared to those from the solvated Zundel. 
This difference arises from the greater flexibility for hydrogen transfer and geometry deformation in the larger cluster. 
Therefore, there is a need to adapt the solvated Zundel family of neural networks to an orthogonal to the original solvated Zundel data. 
We introduce such a scheme in the next section. 


\subsection{Incremental transformation of fragment network models}
As can be seen from Table \ref{tab_initial_error} and Figure \ref{fig_init_compare}, for the 10\% training, we can construct highly accurate models for solvated Zundel fragment energies and obtain full system potential energies with an MAE of 0.36 kcal/mol. Additionally, most of the structures are in the sub-kcal/mol range. However, when these same neural network models are used to evaluate the $target$ energies for the structures in the larger protonated 21 water clusters, the full system error has raised to 5.32 kcal/mol, with a substantial population of structures with errors greater than 10 kcal/mol. This is because the larger fragments, beginning with the protonated water dimer systems, contain geometries that are very different in the protonated 21-water cluster trajectory as compared to the solvated Zundel trajectory. These differences in fragment structures are responsible for the larger errors in fragment energies in Table \ref{tab_initial_error} and Figure \ref{fig_init_compare} and as a consequence the full system error accumulates rapidly with fragment errors according to Eq. (\ref{eq_graph-ML}). 
Our goal is to lay the ground work to transform, as needed, and transfer the neural networks computed from a solvated Zundel database to the larger protonated 21 water cluster database. 

\begin{figure}
\centering
    \subfigure[]
    {\includegraphics[width=0.475\linewidth]{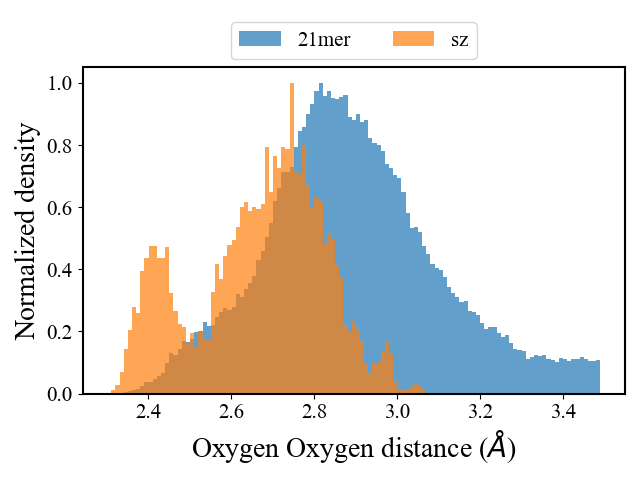}}
    \subfigure[]
    {\includegraphics[width=0.475\linewidth]{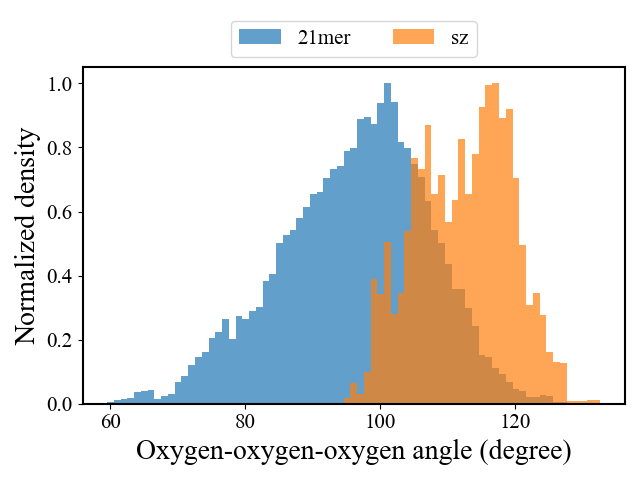}}
\caption{\label{fig_full_rdf}Oxygen-oxygen distance distribution and oxygen-oxygen-oxygen angle distribution for the solvated Zundel and protonated 21 water cluster trajectories.}
\end{figure}

The range of fragment structures that significantly contribute to the solvated Zundel system could be very different from those that contribute to the protonated 21 water cluster system, and this may contribute to error trends noted in Figure \ref{fig_final-main}. 
To gauge the extent to which the range of structures differ between the two libraries, we first examine the differences in distributions of fragment geometries as they appear in a solvated Zundel system with that obtained for the larger protonated 21 water cluster system. The larger system allows each fragment to, in general, sample a greater portion of the potential energy space. For example, this can be seen from the oxygen-oxygen (OO) radial distribution functions for the full systems as shown in Figure \ref{fig_full_rdf}(a), and the hydrogen bonding oxygen-oxygen-oxygen (OOO) angles as shown in Figure \ref{fig_full_rdf}(b). For the solvated Zundel trajectory, OO distances display two distinct peaks. The smaller one centers at 2.4\AA\; corresponding to the Zundel structures and the other around 2.7\AA\; associates with the eigen structures or the solvation water molecules surrounding the Zundel ions. By contrast, the protonated 21 water cluster displays a wider OO radial distribution. A similar trend is observed for the OOO angles, where the larger systems permit greater configurational flexibility and hence produce a wider range of angle distributions. The OO radial, and OOO angle distributions have a critical role in proton transfer pathways. Also see Appendix \ref{sec_H_path}.

\begin{table}[h!]
\centering
\begin{tabular*}{\columnwidth}{@{\extracolsep{\fill}}lcclcc}
\hline  \hline
Fragments& \multicolumn{2}{c}{$(H_2O)_6H^+$}  & \multicolumn{2}{c}{$(H_2O)_{21}H^+$} \\
 &$\omega_{f_r,{\cal R}}$\footnote{For fragments found in the $(H_2O)_6H^+$ AIMD trajectory at $R=3$ from Eq.14
 }
 & MAE\footnote{MAE for fragments found in the $(H_2O)_6H^+$ AIMD trajectory.}& $\omega_{f_r,{\cal R}}$\footnote{For fragments found in the $(H_2O)_{21}H^+$ AIMD trajectory at $R=3$ from Eq.14
 }
 & MAE\footnote{MAE for fragments found in the $(H_2O)_{21}H^+$ AIMD trajectory.}\\
 \hline  
$H_2O$ &	0.13&0.00&	0.14&0.00\\ 
$H_3O^+$ &	0.06&0.00&	0.01&0.00\\
$H_4O_2$&	0.16&0.01&0.30	&0.06\\
$H_5O_2^+$&	0.20&0.01&0.04&0.15\\
$H_6O_3$ &	0.08&0.01&0.28	&0.31\\
$H_7O_3^+$ &	0.24&0.02&0.06	&0.45\\
$H_8O_4$ &	0.01&0.04&	0.12&0.59\\
$H_9O_4^+$ & 0.11 &0.05&	0.05&0.88\\
\hline  \hline
\end{tabular*}
\caption{Mean absolute error(MAE) in kcal/mol from initial models trained on 10\% solvated Zundel fragments (see Table \ref{tab_num_of_geo}) and predictions for both solvated Zundel fragments and protonated 21 water cluster fragments.}
\label{tab_initial_error}
\end{table}

\begin{figure}
    \centering
    {\includegraphics[width=0.75\columnwidth]{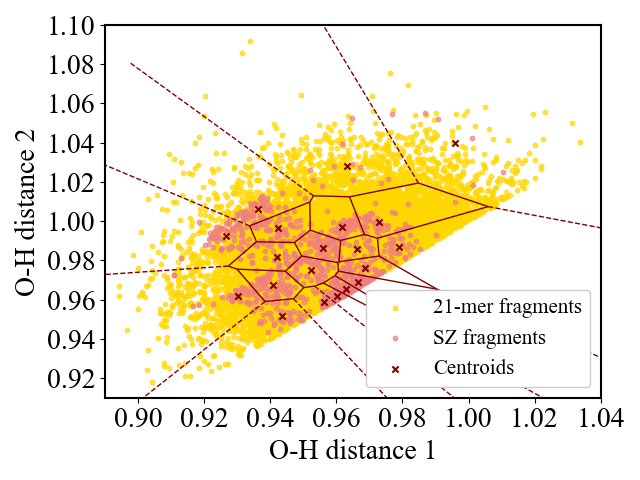}}
    \caption{Distribution of data for the water fragments, and corresponding centroids obtained from k-means. The centroids were derived from the solvated Zundel data (see Table \ref{tab_num_of_geo}). Note that all data lies above the line where the two $O-H$ distances are equal, since ``OH distance 1'' is chosen to be the shorter distance. Clustering is conducted only on the two $O-H$ distances for demonstration.
    }
    \label{fig_voro}
\end{figure}

\begin{figure}
    \centering
    \subfigure[$H_2O$]{\includegraphics[width=0.48\columnwidth]{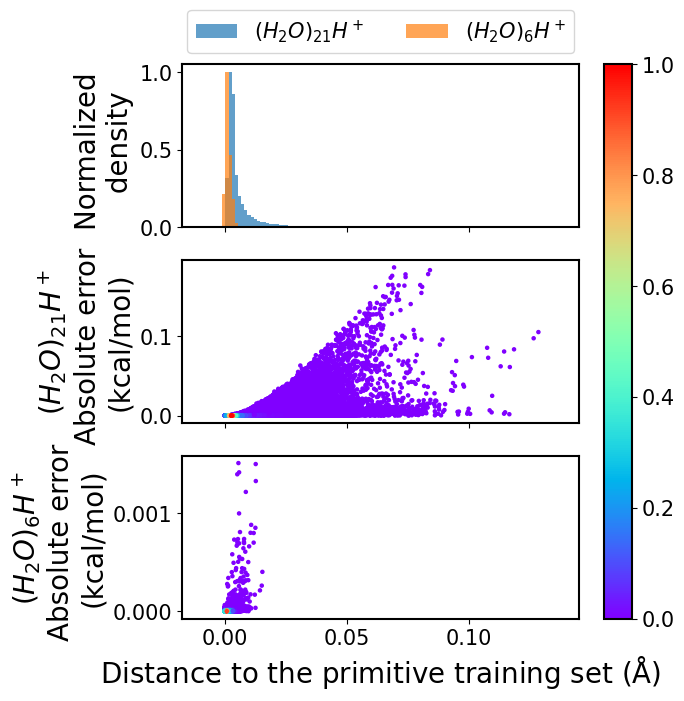}}
    \subfigure[$H_3O^+$]{\includegraphics[width=0.48\columnwidth]{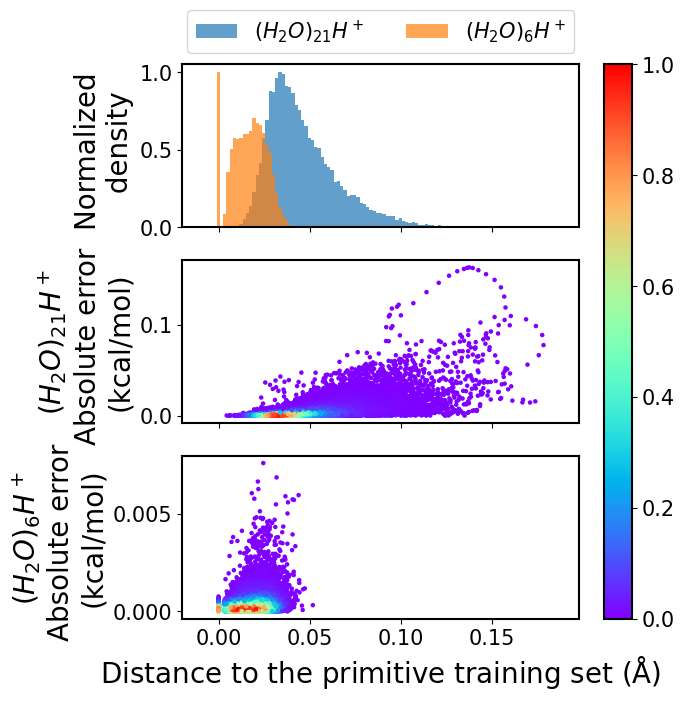}} 
    \subfigure[$H_4O_2$]{\includegraphics[width=0.48\columnwidth]{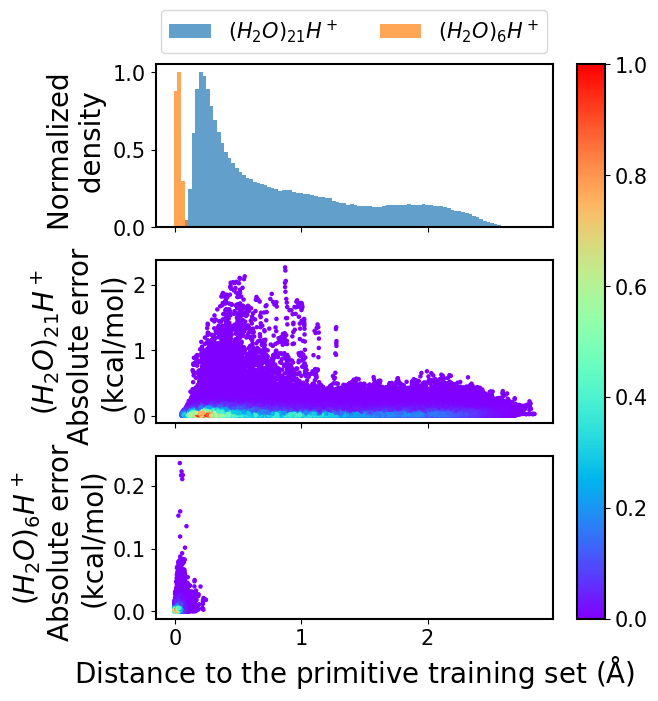}}
    \subfigure[$H_5O_2^+$\label{fig_init_d}]{\includegraphics[width=0.48\columnwidth]{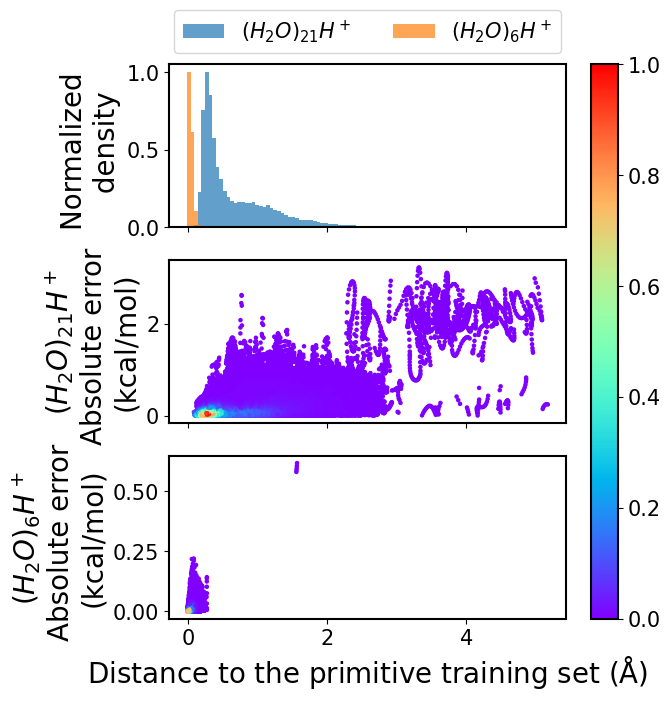}}
    \subfigure[$H_6O_3$]{\includegraphics[width=0.48\columnwidth]{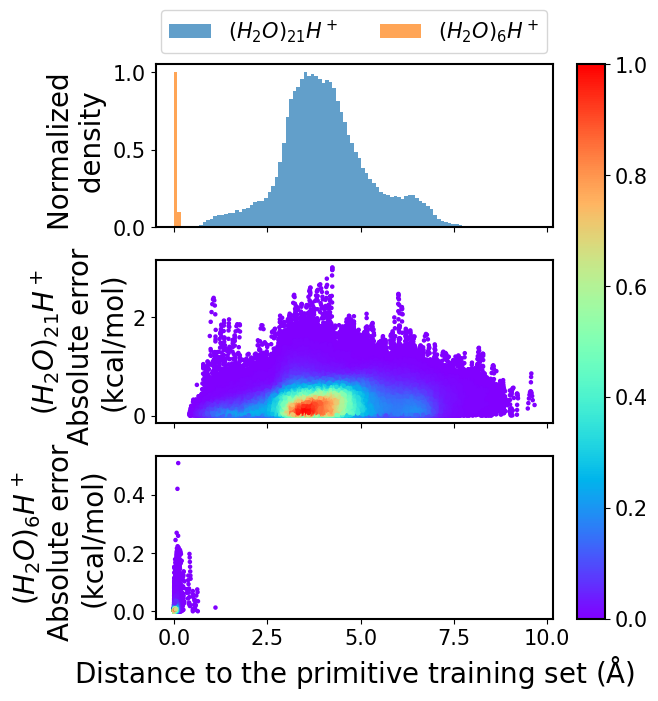}}
    \subfigure[$H_7O_3^+$]{\includegraphics[width=0.48\columnwidth]{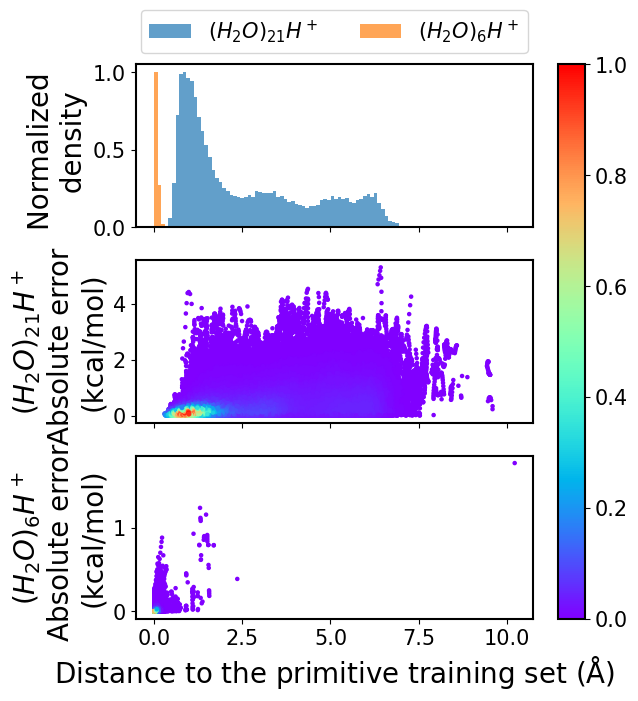}}    \subfigure[$H_8O_4$]{\includegraphics[width=0.48\columnwidth]{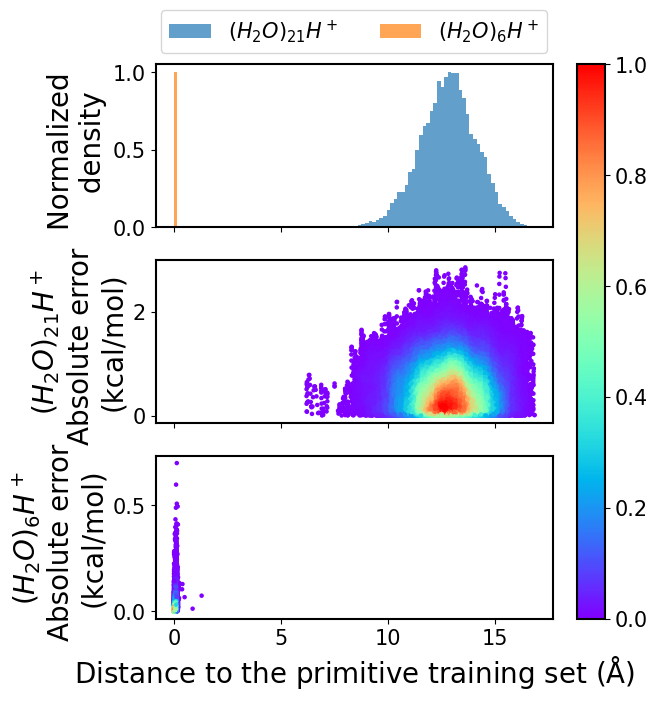}}
    \subfigure[$H_9O_4^+$]{\includegraphics[width=0.48\columnwidth]{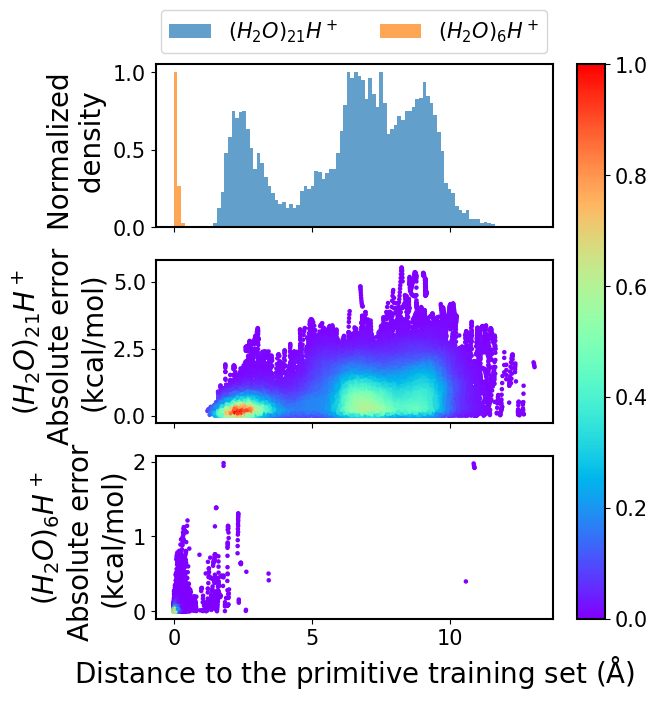}}
    \caption{Fragment data proximity from nearest training point (horizontal axis) and respective distributions (top panel). 
    The absolute error is computed from the neural network predictions trained on solvated Zundel fragments. Color bars represent relative density.
    }
    \label{fig_init}
\end{figure}

To quantify the degree to which the fragment structures differ between the two systems, we first gauge the distribution of data around the training set centroids. A simplified heuristic for such a data distribution is provided in Figure \ref{fig_voro} for water fragments found within both the solvated Zundel as well as the 21-water cluster. In Figure \ref{fig_voro}, a significant difference in water fragment geometries can be observed by comparing the distribution of red and yellow dots. To further probe this difference, we consider the Cartesian distance between the descriptor vector for the k-means centroids and the data points from the two systems, that is, in some sense the level of confidence for a given data point given its proximity to a training point. 
We then compute the distribution of these Cartesian distances from each fragment to their closest training geometries (obtained from the k-means centroid geometries) 
and these are shown along the horizontal axes in Figures \ref{fig_init}. The distributions of distances for fragments from the two systems are displayed using two histograms in the top panels in Figure \ref{fig_init}. From the top panels of each figure in Figure \ref{fig_init}, a clear distinction between the distributions for the two systems begins to appear from $H_4O_2$, which coincides with the pattern of MAE differences in Table \ref{tab_initial_error}. As the size of the fragment increases, the difference in geometry distribution also increases as shown by the right-shifting peaks of blue histograms on the top panels of Figure \ref{fig_init}. 

In the middle and bottom panels of each figure of Figure \ref{fig_init} we present the error in $target$ energy for 21-mer fragments and the error in CCSD energy for the solvated Zundel fragments respectively, which results from the difference in the distribution of data noted above. 
From the scale of the vertical axes in the middle and bottom panels, the effect of this perturbation in distribution is already clear. 
This strongly indicates that for geometries with large distances to the primitive training set, the predictive accuracy is generally low. Therefore, an amended transferring process needs to be designed to expand the valid training set regions to build confidence in the accuracy of predictions for all geometries. 

\subsection{The corresponding LLM challenge}
The difficulty in extrapolating the protonated 21-water cluster surface, from the learning process set in place using solvated Zundel data, can also be understood from the LLM perspective presented in Section \ref{sec_compare}. As noted in Figure \ref{Graphs-PES-LLM-process}, the attention weights are limited by the breadth of context presented within the training data, as discovered during the tokenization and embedding processes. 

To make the analogy much more precise, we posed the following question to the AI Chatbot, Google Gemini\cite{GoogleGemini_YYYYMMDD}: ``We wish for you to write a 186 word length paragraph, using only 51 words. However, to train you, we can only provide you lines that are precisely 51 words long.'' 
Note that the number 51 and 186 in the above query arise from the fact that the solvated Zundel has 51 nuclear dimensions whereas the protonated 21-water cluster as 186 nuclear dimensions. 

Google Gemini provided the following response to this query.
``This is a very specific and unusual constraint for training a language model, as typical training involves a wide variety of sentence lengths and structures. Here is what that restriction implies for how I would be `trained' or what I could `learn' from such data:
(a) Limited Grammatical Diversity: If every sentence is precisely 51 words, I will not be exposed to the natural variations in sentence length and complexity common in human language. This could make it difficult for me to generate shorter, longer, or more nuanced sentences.
(b) Focus on Word Choice within Constraint: My `learning' would heavily focus on how to use words effectively to reach that 51-word count, rather than focusing on natural flow or conciseness.
(c) Potential for Repetitive Structures: To meet the word count consistently, the provided training sentences might employ similar grammatical structures or rely on a lot of descriptive phrases to extend length. I might then mimic these patterns in my own output.
(d) Challenges in Generalization: If I am only trained on 51-word sentences, my ability to understand and generate text outside of that specific length might be severely hampered. I might struggle with prompts that require very brief answers or very detailed, extended explanations.
In essence, while I could still process and analyze the patterns within those 51-word sentences, my understanding and generation capabilities would be highly specialized to that very specific input format. It would be like learning to play a song with only one instrument and then being asked to perform a whole orchestra.''

Clearly from an LLM constraint, this is a severe restraint, however there is additional information that we possess making it possible for us to take such as a step in this paper, as we will see in the next section.

\section{A transformer learning platform to refine the fragment neural networks towards the $E^{target}$}
\label{transformer-SI}

\begin{figure}
    \centering
    \subfigure[Training regions as the union of circles centered at centroids]{\includegraphics[width=0.475\columnwidth]{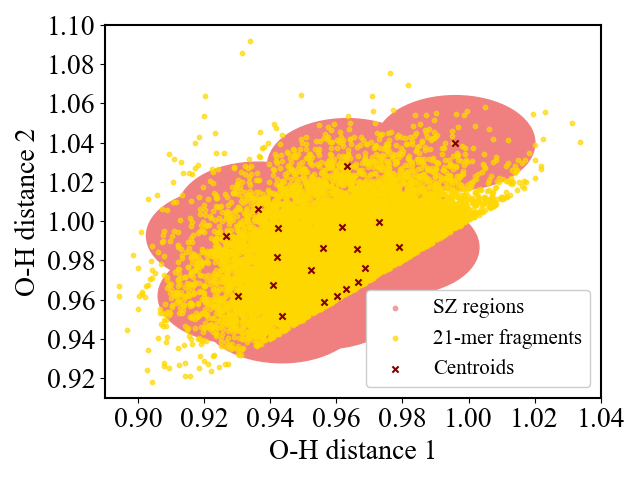}}
    \subfigure[Slicing of the space around training regions for transferring]
    {\includegraphics[width=0.475\columnwidth]{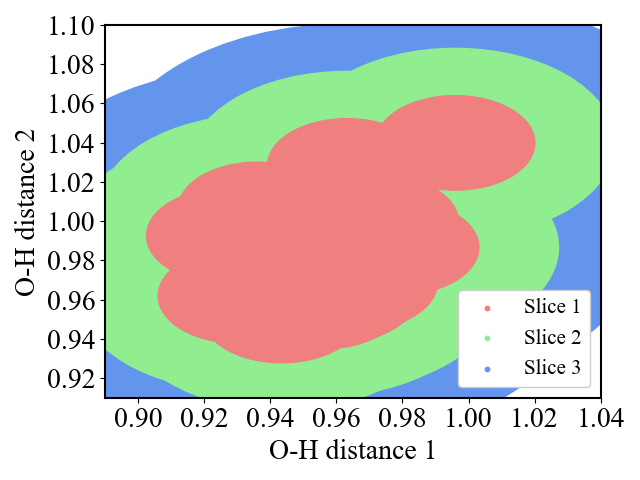}}
    \subfigure[Centroid positions]
    {\includegraphics[width=0.475\columnwidth]{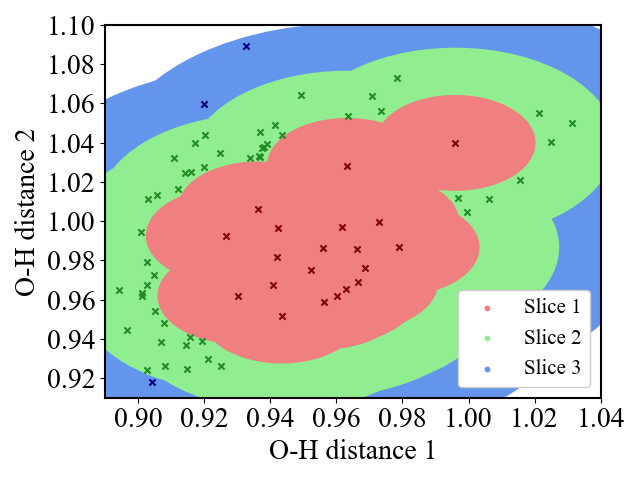}}
    \caption{(a) We place a circle with on each centroid(same as in Figure\ref{fig_voro}) with radius computed from the maximum sample-to-centroid distance for SZ fragments to define training regions. These regions imply that any data involved can be accurately predicted by SZ models. (b) To include data outside the training regions, we further slice the space based on their distance to the closest centroid and create an incremental transfer process. (c) Centroid positions for slices 2 and 3 are obtained from the recursive mini-batch-k-means.}
    \label{fig_voro10}
\end{figure}

As discussed above, 
it is necessary to expand the previously defined training set regions. This is because the fragments sampled from the 21-mer trajectory include a much broader and more diverse range of geometries compared to those from the solvated Zundel trajectory. Also see discussion in Appendix \ref{sec_H_path} in this regard. The difference in range of geometries sampled can be seen by comparing the geometry distribution of the water molecules represented by red and yellow dots in Figure \ref{fig_voro}. 
Therefore, we expand the training sets using the portion of the 21-mer data that are not visited in solvated Zundel trajectory. 
The process produces an adapted family of neural networks that apply to both the solvated Zundel data and 21-mer geometries. 
For this purpose, we introduce a systematic process to incrementally expand the domain of training beginning from the data space of fragment geometries generated using the solvated Zundel AIMD trajectory. The process is illustrated in Figures \ref{fig_voro10}.

We first imagine the full data space for each fragment to be divided into two orthogonal dimensions. One dimension is well sampled during the solvated Zundel trajectory, but configurations in the second dimension are only sampled during an AIMD trajectory for the larger water cluster system. In summary, our goal is to facilitate the sampling of this second orthogonal dimension, since the first dimension is already well-sampled during the solvated Zundel study. See Figures \ref{fig_sz-sz} and \ref{fig_final-main}.
This is achieved by first interpreting the entire orthogonal data space as one single effective dimension and tessellating this dimension into a series of slices based on the distance to the closest reference points from the solvated Zundel training set. 
For example, in Figure \ref{fig_voro10}(a) and \ref{fig_voro10}(b), the red region represents the data span of the H$_2$O fragments obtained from the original solvated Zundel simulation (compare with Figure \ref{fig_voro}). Expanding this space into the orthogonal 21-mer data space involves creating multiple additional layers or ``shells" in that data space surrounding the solvated Zundel core displayed in red. 
The additional slices incrementally added are shown in green and blue in Figure \ref{fig_voro10}(b). These slices facilitate an incremental learning procedure to go beyond the neural networks obtained from the solvated Zundel systems based on the proximity of data to inner layers. In Figure \ref{fig_voro10}, the slice-width (that is the width of the green and blue regions) is equal to the radius of the red circles, which represent the maximum sample-to-centroid distance for each training sample in the solvated Zundel data. (Further details on this geometric tesselation procedure of the data space is given in Appendix \ref{sec_kmeans_nn}.) Following this, for each new slice of the space (that is, for example the green and blue slices in Figure \ref{fig_voro10}), we apply recursive mini-batch-k-means algorithms\cite{kmeans} (a geometric tesselation procedure for higher dimensional spaces) on geometries within the specific slice to identify a set of geometries with desired mutual distances to represent the slice of fragment energy space. This algorithm aims to perform k-means clustering recursively to reduce complexity and automatically derive an appropriate number of clusters and training samples to facilitate description of the protonated 21-mer surface. More details about the recursive mini-batch-k-means clustering algorithm can be found in the Appendix \ref{slicing}. Figure \ref{fig_voro10}(c) shows additional centroids in each slice. By comparing Figure \ref{fig_voro10}(c) and \ref{fig_voro}, we note that the additional centroids in the green and blue slices in Figure \ref{fig_voro10}(c) effectively sample the region in yellow in Figure \ref{fig_voro} 
and generate additional training samples to adapt the neural networks from the solvated Zundel subspace to those seen in larger clusters. 

After identifying the additional centroids using the recursive k-means clustering algorithms discussed in detail in Appendix \ref{sec_kmeans_nn} and \ref{slicing}, an additional training set is formed from the geometries closest to each centroid to adapt the neural network models. Once the neural networks are trained on the additional training set, they are retrained with geometries closest to all existing centroids as a fine-tuning process. The complexity of neural network training is reduced here because the transfer process is conducted based on the proximity of the data to the primitive training set and then adapted gradually to new regions. In the next section, we demonstrate the efficacy of this algorithm by gauging the error in the protonated 21-mer surfaces when these algorithms are implemented. 

\section{Protonated 21 water cluster potential energy predictions after incremental transformations in Section \ref{transformer-SI}}

\subsection{Fragment accuracy improves during the transfer learning protocol introduced in Section \ref{transformer-SI}}
\label{sec_h5o2}

\begin{figure}
    \centering
    \subfigure[H$_5$O$_2^+$]{\includegraphics[width=0.48\columnwidth]{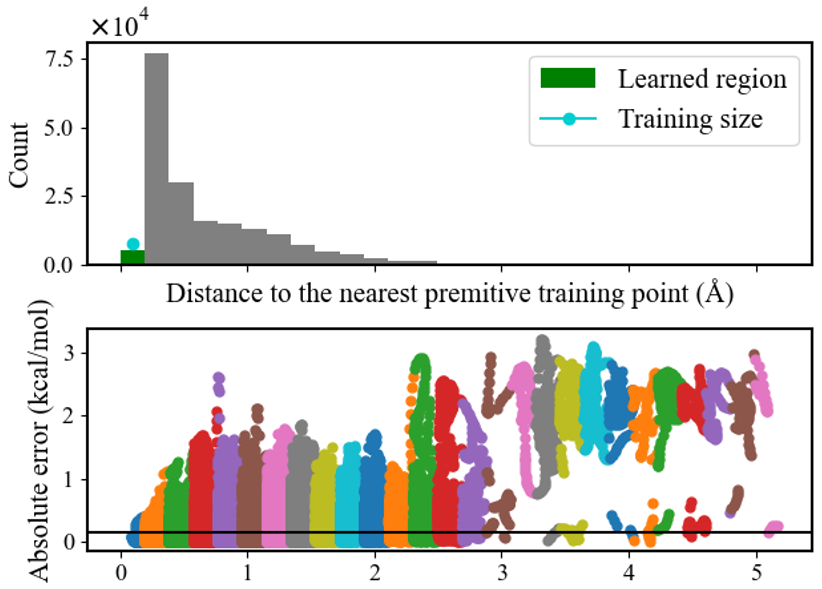}}
    \subfigure[H$_4$O$_2$]{\includegraphics[width=0.48\columnwidth]{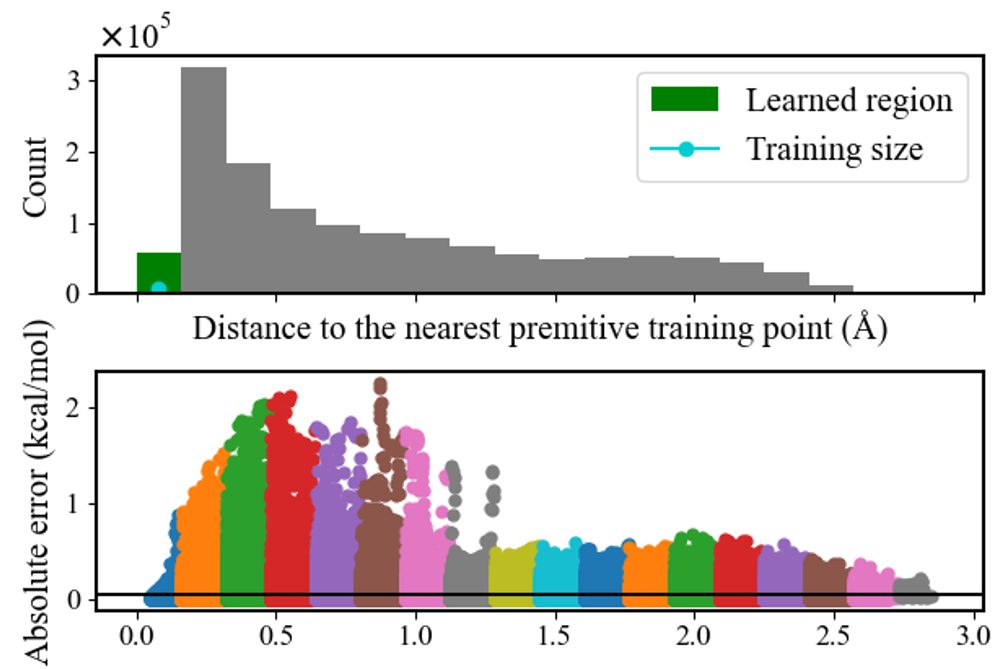}}
    \subfigure[H$_6$O$_3$]{\includegraphics[width=0.48\columnwidth]{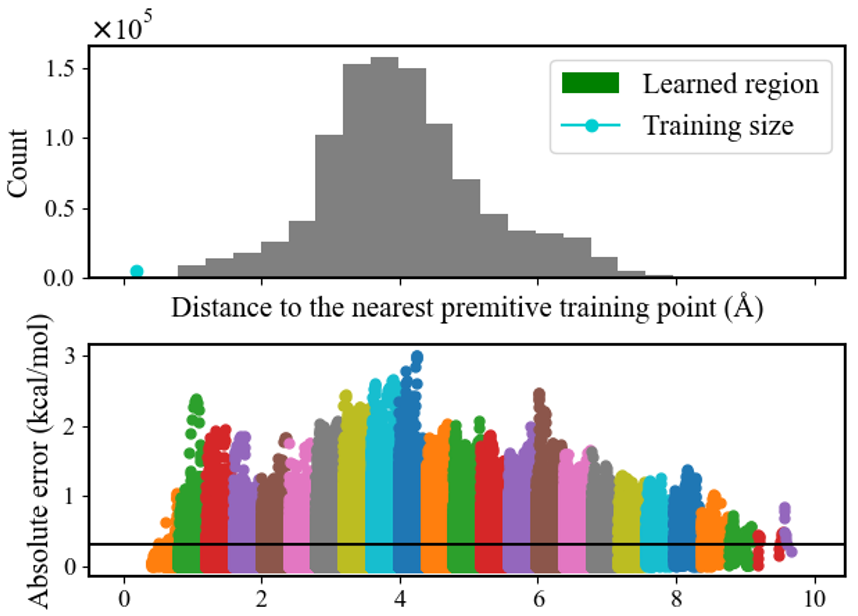}}
    \subfigure[H$_7$O$_3^+$]{\includegraphics[width=0.48\columnwidth]{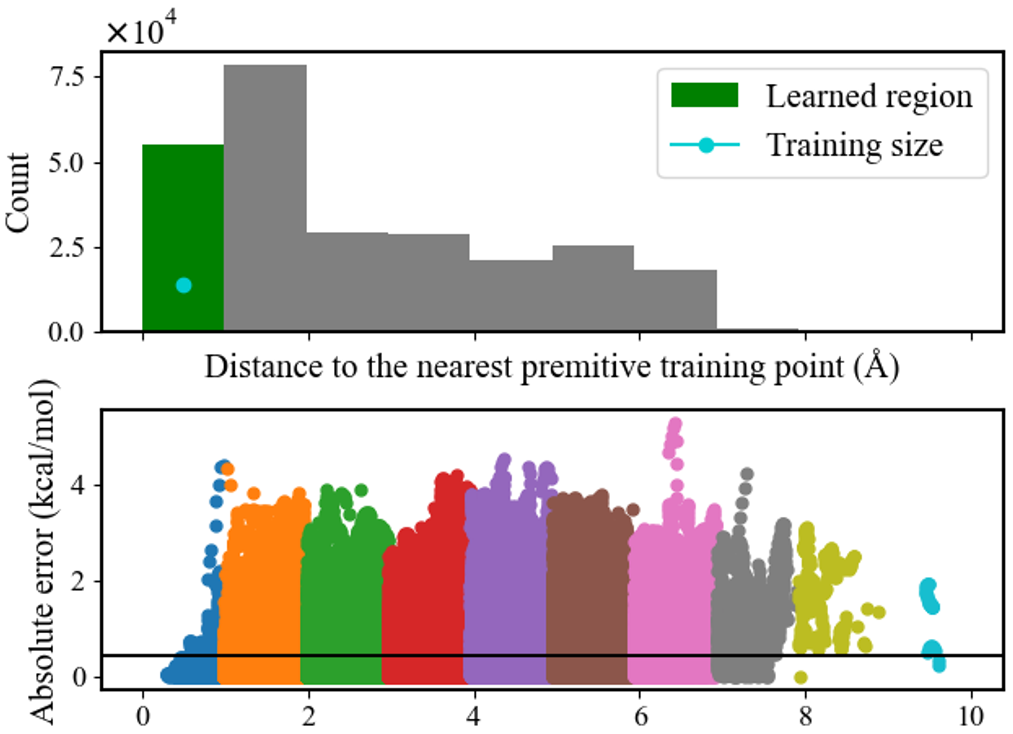}}
    \subfigure[H$_8$O$_4$]{\includegraphics[width=0.48\columnwidth]{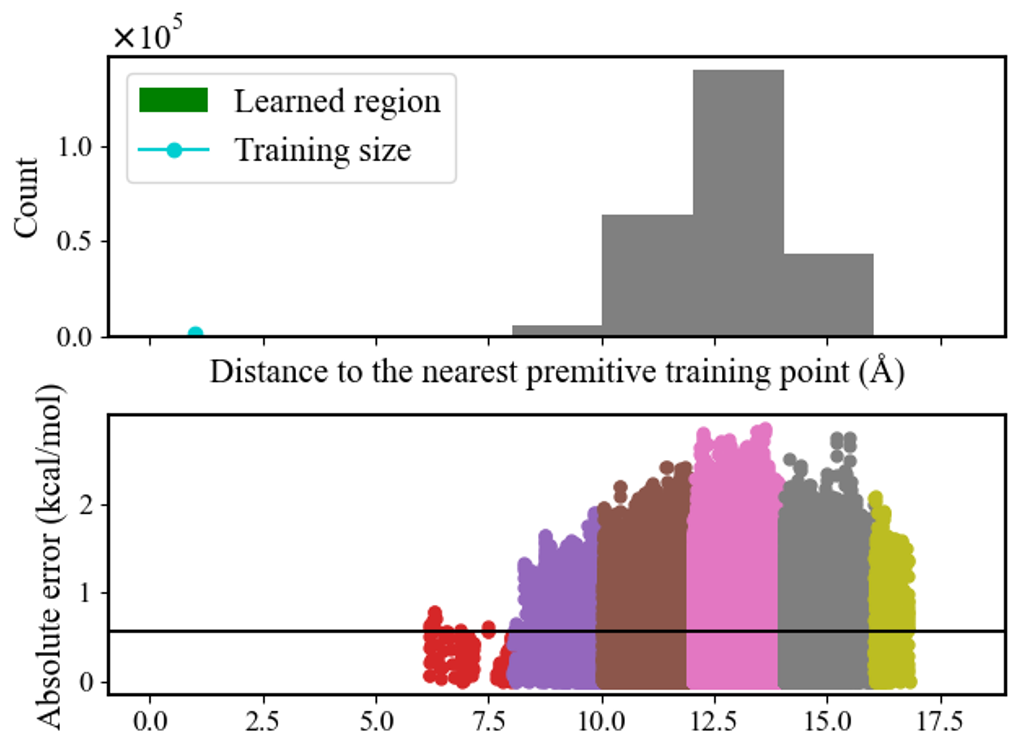}}
    \subfigure[H$_9$O$_4^+$]{\includegraphics[width=0.48\columnwidth]{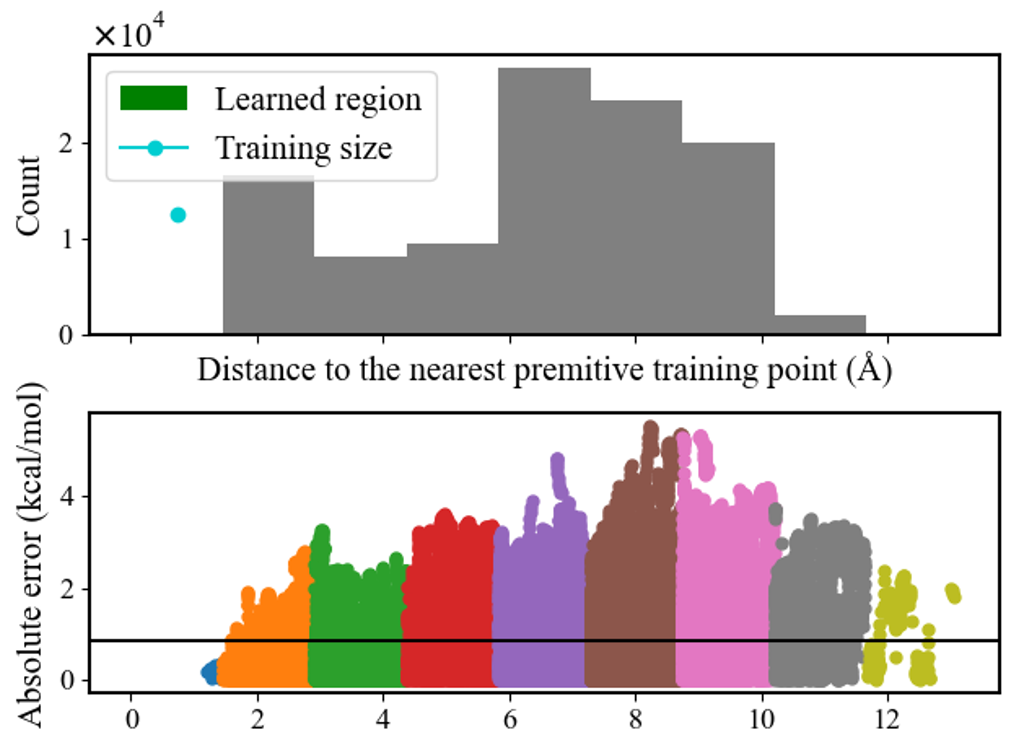}}
    \caption{Geometry distributions (top panels) and error distributions (bottom panels)  for (a) H$_5$O$_2^+$,  (b) H$_4$O$_2$, (c) H$_6$O$_3$, (d) H$_7$O$_3^+$, (e) H$_8$O$_4$ and (f) H$_9$O$_4^+$. Fragment structures deviate significantly between the solvated Zundel and 21-mer trajectories necessitating the transformer algorithm in Section \ref{transformer-SI}. The mean absolute errors (MAE) are note using dark horizontal lines in the bottom panels and these are also reported in Table \ref{tab_all} under the ``MAE before transfer'' column. Compare these MAEs with the solvated Zundel MAEs in Table \ref{tab_initial_error} along with the increased probabilities of these fragments in 21-mer as seen from the $\omega_{f_r,{\cal R}}$ in Table \ref{tab_initial_error}.}
    \label{fig_slicea}
\end{figure}

\begin{figure}
    \centering
    \subfigure[1 slices learned]{\includegraphics[width=0.48\columnwidth]{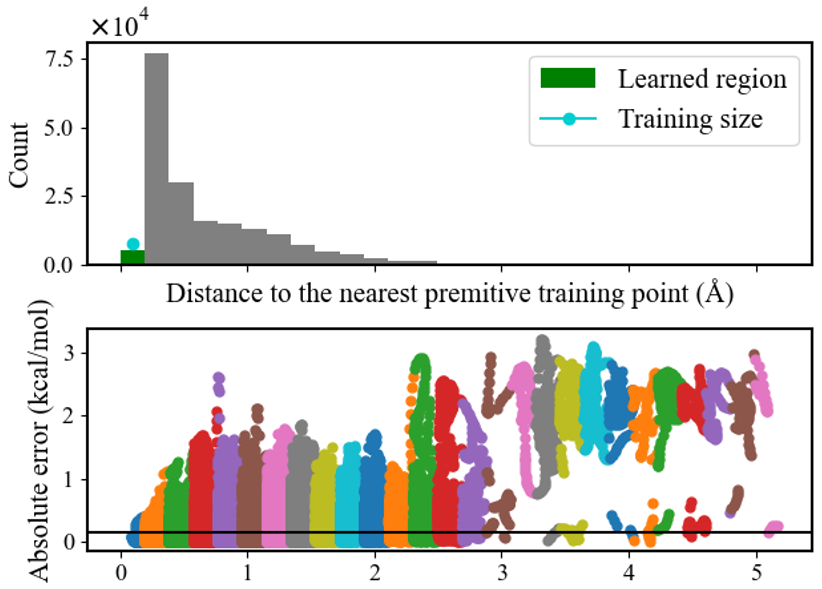}}
    \subfigure[10 slices learned]{\includegraphics[width=0.48\columnwidth]{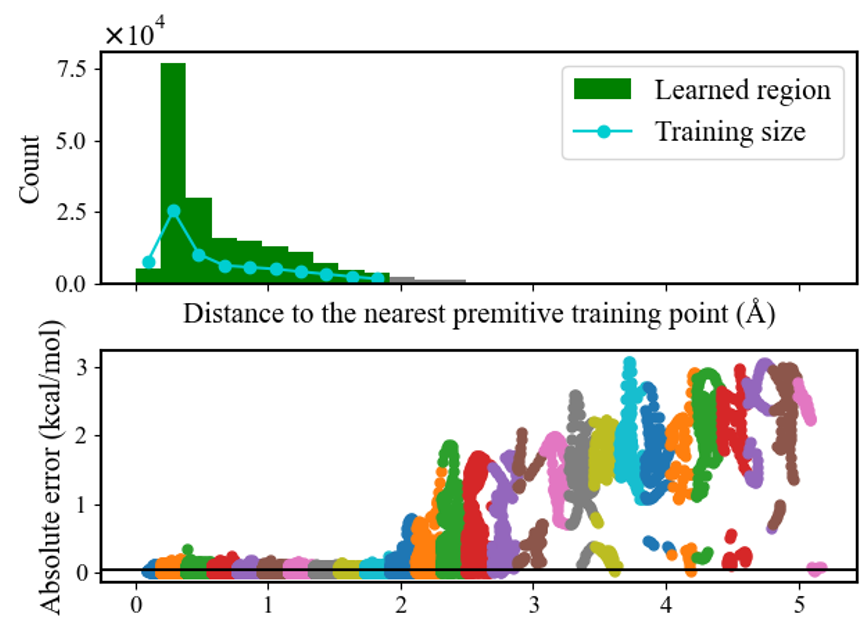}}
    \subfigure[20 slices learned]{\includegraphics[width=0.48\columnwidth]{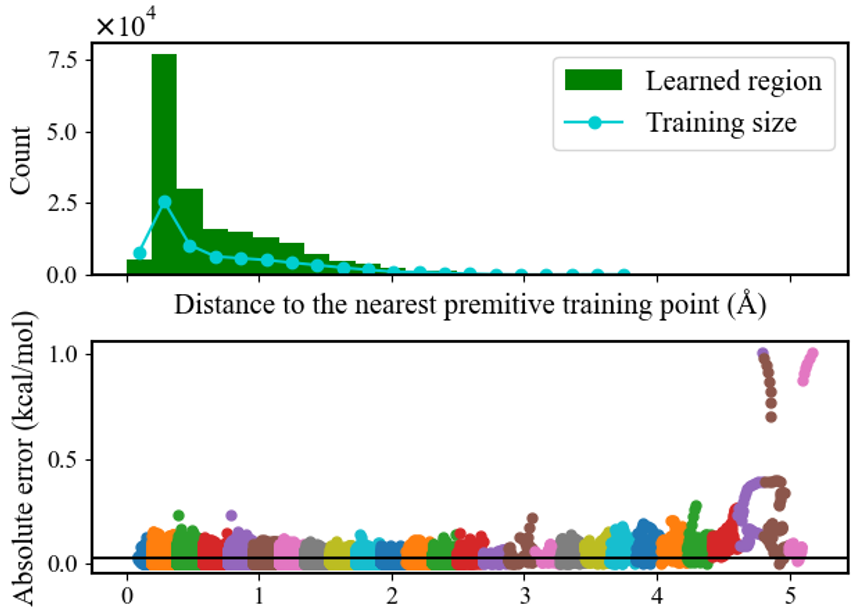}}
    \subfigure[All 27 slices learned]{\includegraphics[width=0.48\columnwidth]{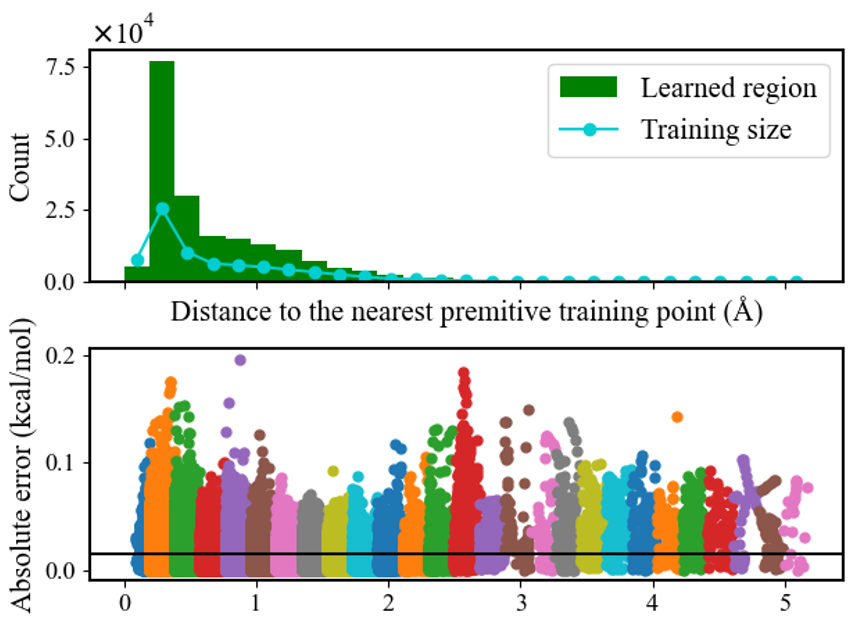}}
    \subfigure[MAE changes with number of slices transferred]{\includegraphics[width=0.48\columnwidth]{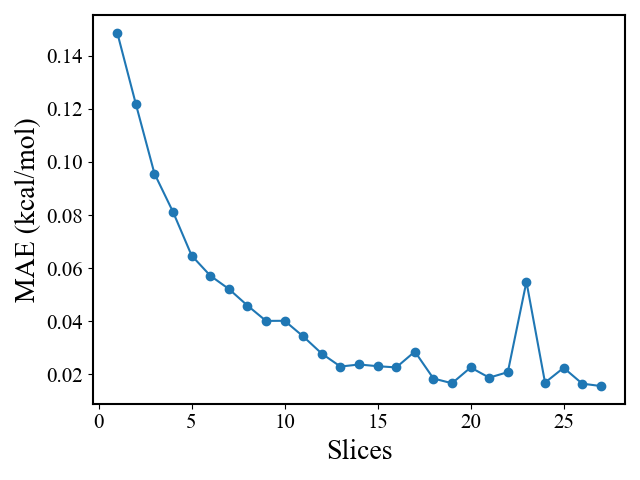}}
    \caption{Error distribution change with the number of slices trained for $H_5O_2^+$.}
    \label{fig_slice}
\end{figure}


To improve our neural network models for protonated 21-mer, in Section \ref{transformer-SI}, we discussed how the full data space is divided into two dimensions. One of these dimensions contain fragment geometries in close proximity to training data obtained from  the solvated Zundel system, 
and the other includes fragment geometries that are farther away from these original training data. We presented these as two orthogonal learning dimensions in Section \ref{transformer-SI}. We also discussed how the second dimension is partitioned, incrementally, into a series of slices that forms subspaces as shown in Figure \ref{fig_voro10}(b). We now apply this partitioning scheme for all fragments. 
Figures \ref{fig_slicea} and \ref{fig_slice} complement our discussion here, with additional data for all other fragments in Appendix \ref{sec_com_fig}. The said partitioning of the orthogonal dimension 
is presented in Figure \ref{fig_slicea}, top panels, 
where the horizontal axes refers to the distance to the nearest training set data point (which was obtained as above from the solvated Zundel data). It is this horizontal axis that is taken as the orthogonal dimension as referred to in Section \ref{transformer-SI}, and sliced into smaller regions where additional mini-batch k-means clustering calculations are performed. The vertical axis in Figure \ref{fig_slicea}, top panel
shows the number of geometries inside the corresponding slice.  
In gray we show the distribution of geometries obtained from the protonated 21 water cluster trajectory and in green we show the data space that was already visited in the solvated Zundel trajectory. Clearly, even for H$_5$O$_2^+$, there is a large deviation of structures obtained from the 21-mer trajectory, as compared to those obtained from the solvated Zundel trajectory, and these new structures contribute significantly to the overall errors as seen from the bottom panels of Figure \ref{fig_slicea}. Comparison of Figure \ref{fig_slicea}, top panels, to Figure \ref{fig_voro} shows how dramatically the structures deviate as the fragment sizes grow. 
In Appendix \ref{sec_H_path} we analyze these deviations in structures sampled between the solvated Zundel and 21-mer data. 
The goal of the incremental slicing algorithm presented in Section \ref{transformer-SI} is to improve on these errors, as we discuss below.

Details regarding the construction of  the histogram error plots are as follows. The slice widths, or bin sizes in Figures \ref{fig_slicea},  \ref{fig_slice} and complementary figures in Appendix \ref{sec_com_fig}, are determined from the maximum sample-to-centroid distances from the 10\% mini-batch-k-means clustering. This is identical to the sampling process to generate the primitive training set, on solvated Zundel data sets as we discussed in Section \ref{solZ-results}. This distance approximates the farthest sample that can be accurately predicted by the model, thus serving as a boundary between the trained and untrained regions of the potential energy data space. 
Consequently, we plot the regions trained from solvated Zundel in green in Figures \ref{fig_slicea} and   \ref{fig_slice}. Each bin in the histogram plots in the top panels also includes a cyan dot to indicate the number of training samples used for that specific slice in the incremental training process. This helps to assess the extent of training needed as compared to the total number of samples in each slice (i.e., the respective bin heights). Notably in the first bin, the number of training samples exceeds the number of fragments from the protonated 21 water cluster trajectory, which arises from the limited fragment geometric overlap between the solvated Zundel and the protonated 21 water cluster trajectory. 

\begin{table}[tbp]
\centering
\begin{tabular}{|l|r|r|r|c|c|c|r}
\hline  \hline
&	\multicolumn{1}{c|}{H$_{13}$O$_6^+$} & \multicolumn{1}{c|}{Transfer} & H$_{43}$O$_{21}^+$ & & MAE  &MAE  \\ 
 &	training & \multicolumn{1}{c|}{size\footnote{Numbers show the order of magnitude of data.}} & data-set & & before &after\\ 
 & set size &  & \multicolumn{1}{c|}{size} & & transfer\footnote{in kcal/mol}  & transfer\footnotemark[1] \\ \hline
$H_2O$ &	3891 & \textasciitilde 16000& 559877& 3\% &	0.00&0.00\\ 
$H_3O^+$ &1704	&\textasciitilde4000&	34297& 11\% & 0.00&0.00\\
$H_4O_2$&	6238&\textasciitilde480000&	1356707& 35\% & 0.06&0.00\\
$H_5O_2^+$&	7751&\textasciitilde76000&	190727& 40\% & 0.15&0.02\\
$H_6O_3$ &	4694&\textasciitilde470000&	1014960& 46\% & 0.31&0.02\\
$H_7O_3^+$ &	13958&\textasciitilde90000& 257260& 35\% & 0.45&	0.07\\
$H_8O_4$ &	1575&\textasciitilde134000&	253889& 53\% & 0.59&0.05\\
$H_9O_4^+$ &	12414&\textasciitilde65000&	108123& 60\% & 0.88&0.10\\
\hline  \hline
\end{tabular}
\caption{The third column shows the number of fragments from the protonated 21 water cluster trajectory used for improving the accuracy of the original neural networks developed using the solvated Zundel trajectory data-set. The fragment mean absolute errors before and after transferring are also shown. Note the effect on the larger clusters. }
\label{tab_all}
\end{table}

Hence, we incrementally create additional training samples from each slice of the space (as discussed in Section \ref{transformer-SI}) and adapt the neural networks. The resulting improvement on prediction error can be observed by comparing Figure \ref{fig_slice}(a) and \ref{fig_slice}(b). In the bottom panel of Figure \ref{fig_slice}(b), a clear distinction on error distribution emerges between trained and untrained slices. Predictions remain accurate across the 10 learned slices, while errors in the untrained slices remain high. 
As the incremental transfer process continues, the error 
in systematically reduced, from a maximum value of approximately 3 kcal/mol in Figure \ref{fig_slice}(b), to a maximum value of 1 kcal/mol in Figure \ref{fig_slice}(c), and finally to 0.2 kcal/mol in Figure \ref{fig_slice}(d). This is also the case for larger fragments as seen from the figures in Appendix \ref{sec_com_fig}. 
The systematic reduction in mean absolute errors due to this incremental learning process 
is summarized in Figure \ref{fig_slice}(e) and the corresponding final figures in Appendix \ref{sec_com_fig}.  
These final errors are also summarized in Table \ref{tab_all} and are comparable to the errors from primitive neural networks trained only on solvated Zundel fragments, and reported in Table \ref{tab_initial_error}. At this point, the adapted neural network models have successfully evolved to make accurate predictions for both solvated Zundel and protonated 21 water cluster systems.
In the next section, we integrate all adapted models to perform full system predictions and analysis for the protonated 21 water clusters.

\subsection{Full system potential energy prediction accuracy after transferring all fragment models}
\begin{figure}
    \centering
    \subfigure[Using transformed  NN models, ${\cal R}$=0]{\includegraphics[width=0.45\columnwidth]{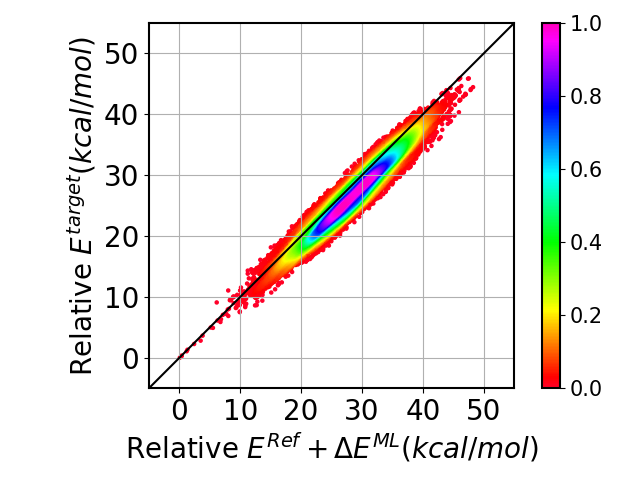}}
    \subfigure[Using transformed NN models, ${\cal R}$=1]{\includegraphics[width=0.45\columnwidth]{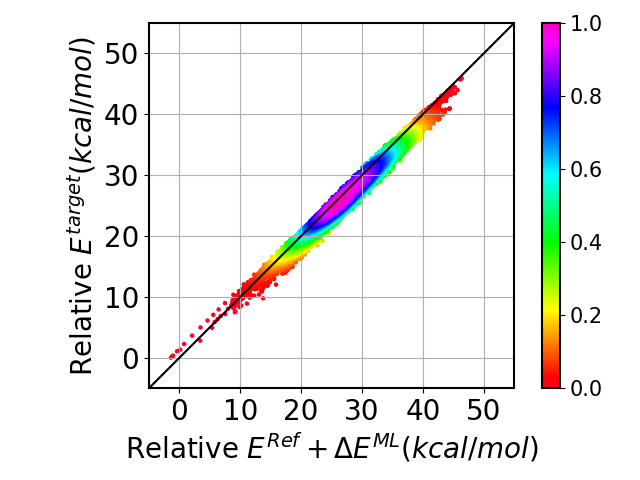}}
    \subfigure[Using transformed NN models, ${\cal R}$=2]{\includegraphics[width=0.45\columnwidth]{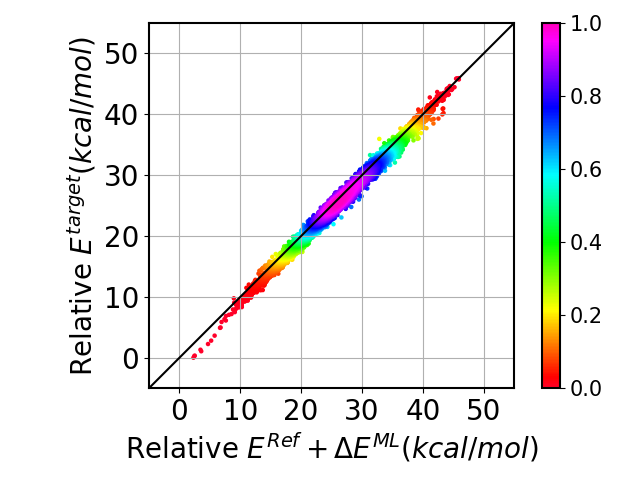}}
    \subfigure[Using transformed NN models, ${\cal R}$=3]{\includegraphics[width=0.45\columnwidth]{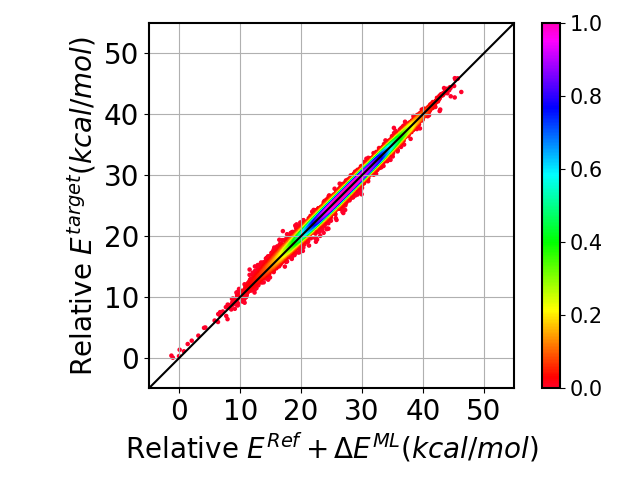}}
    \caption{The protonated 21 water cluster potential energy prediction accuracy from transformed NN models. }
    \label{fig_final-main2}
\end{figure}

\begin{figure}
    \centering
    \subfigure[Before transferring]{\includegraphics[width=0.48\columnwidth]{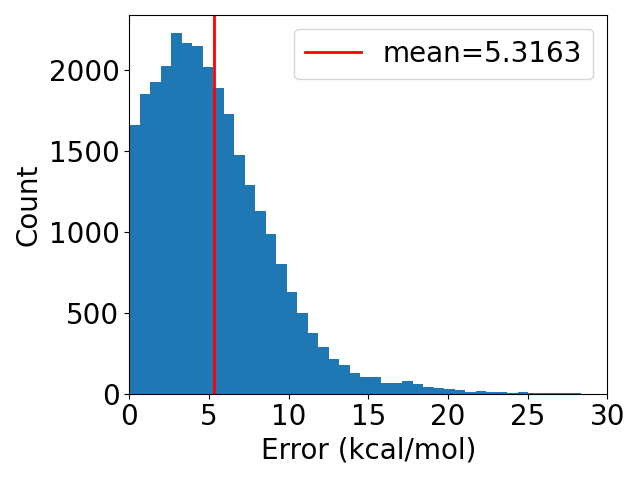}}
    \subfigure[After transferring]{\includegraphics[width=0.48\columnwidth]{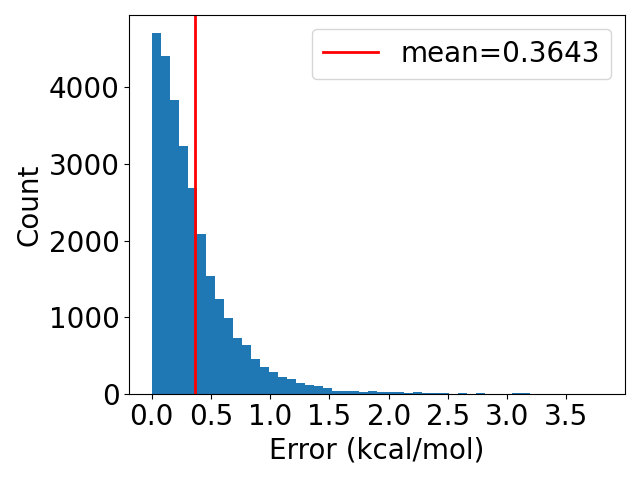}}
    \caption{The protonated 21 water cluster full system potential energy absolute error predicted from models before and after transferring. Subfigure (a) recasts the same data in Figure \ref{fig_final-main}(d) and Subfigure (a) recasts the same data in Figure \ref{fig_final-main2}(d) for easy comparison.}
    \label{fig_final}
\end{figure}

\begin{figure}
{\includegraphics[width=0.95\columnwidth]{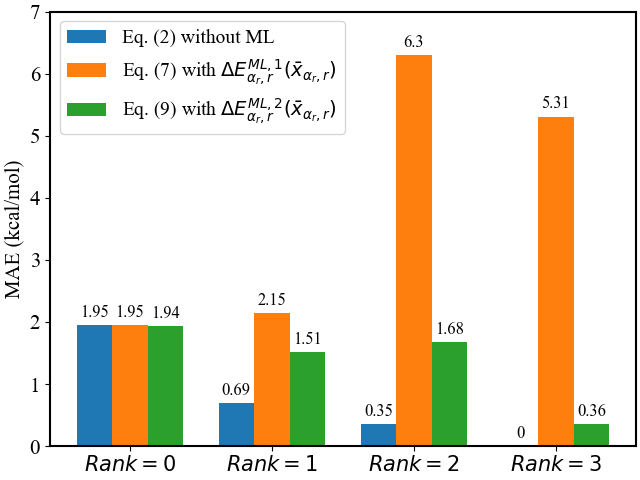}}
    \caption{Improved performance of the transformed neural networks for protonated 21 water cluster potential energy predictions. Blue histogram represents Eq. (\ref{eq_graph-ML-main}) without the use of machine learning. The orange histogram represents Eq. (\ref{eq_graph-ML-main-2}) but with with $\left\{ \Delta E_{\alpha_r,r}^{ML, 1}({\bf {\bar x}_{\alpha_r,r}}) \right\}$, that is the solvated Zundel network models used to obtain $E_{{\text{system,2}}}^{target}({\bf {\bar x}})$, and the 
green histogram depicts Eq. (\ref{eq_graph-ML-main-2}) as written that is $\left\{ \Delta E_{\alpha_r,r}^{ML, 2}({\bf {\bar x}_{\alpha_r,r}}) \right\} \rightarrow  E_{{\text{system,2}}}^{target}({\bf {\bar x}})$. All MAEs are computed with reference to graph energy at rank 3.
}
    \label{fig_final-main-summary}
\end{figure}

Upon incremental modification of all fragment neural networks, 
the overall error in the protonated 21-water cluster reduces from 5.32 kcal/mol (the error when the solvated Zundel models are directly used, that is $\left\{ \Delta E_{\alpha_r,r}^{ML, 1}({\bf {\bar x}_{\alpha_r,r}}) \right\} \rightarrow  E_{{\text{system,2}}}^{target}({\bf {\bar x}})$, Figure \ref{fig_final-main}) to 0.36 kcal/mol (when the modified, incremental, and directed learning process is used, that is $\left\{ \Delta E_{\alpha_r,r}^{ML, 2}({\bf {\bar x}_{\alpha_r,r}}) \right\} \rightarrow  E_{{\text{system,2}}}^{target}({\bf {\bar x}})$). The corresponding distribution of errors is seen in Figure \ref{fig_final-main2} and a summary  for ${\cal{R}}=3$ before and after transferring is presented in Figure \ref{fig_final}. Importantly, the distribution of errors shrinks towards the main diagonal with increase in ${\cal R}$, in Figure \ref{fig_final-main2} unlike the case in Figure \ref{fig_final-main}, and the predictions remain accurate for the entire range of structures. This can also be seen in Figure \ref{fig_final}. In Figure \ref{fig_final}(a), the error distribution before transferring exhibits a peak near 3 kcal/mol, corresponding to the purple region representing the highest density in Figure \ref{fig_final-main}(d). After neural network modification, 
the peak shifts 
reflect the improved accuracy displayed in Figure \ref{fig_final-main2}(d) where the purple region is strictly along the diagonal.

 We summarize all mean absolute errors 
 in Figure \ref{fig_final-main-summary}. These 
 are computed with respect to the graph energy at ${\cal R}$=3 and hence 
 the blue bar on the right of Figure \ref{fig_final-main-summary} is identically zero for ${\cal R}$=3. 
 The orange ML error 
 very clearly shows the deviations predicted in Figure \ref{fig_final-main} that are then corrected after modifying the fragment neural networks as seen from the green histograms. We also compare the corresponding computational cost for constructing these models with direct computation of CCSD energies in Section \ref{sec_cost}.
 

As per Table \ref{tab_all},  
one may expect major improvements in the protonated 21 water cluster potential energy predictions by including 
the larger ${\cal R}$=2 and ${\cal R}$=3 fragments, $H_6O_3$, $H_7O_3^+$, $H_8O_4$, $H_9O_4^+$. 
\begin{figure}
    \centering
    \includegraphics[width=0.6\linewidth]{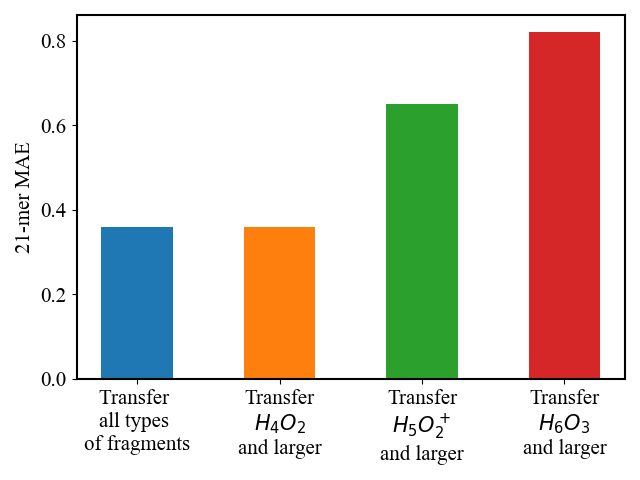}
    \caption{The protonated 21-water cluster MAE if the transformation process is conducted only on fragments larger in size than those listed. This shows that nodes do not need to be transformed and accuracy may be maintained even when all edges are not transformed.}
    \label{fig_combined}
\end{figure}
\begin{figure}
    \centering
    \subfigure[Transfer $H_5O_2^+$ and larger]{\includegraphics[width=0.45\columnwidth]{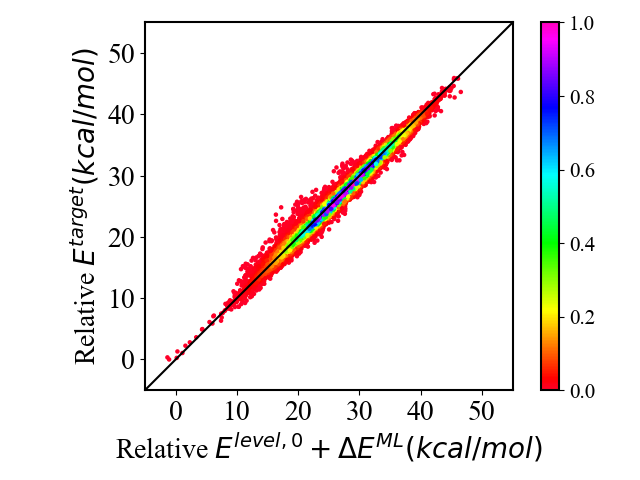}}
    \subfigure[Transfer $H_6O_3$ and larger]{\includegraphics[width=0.45\columnwidth]{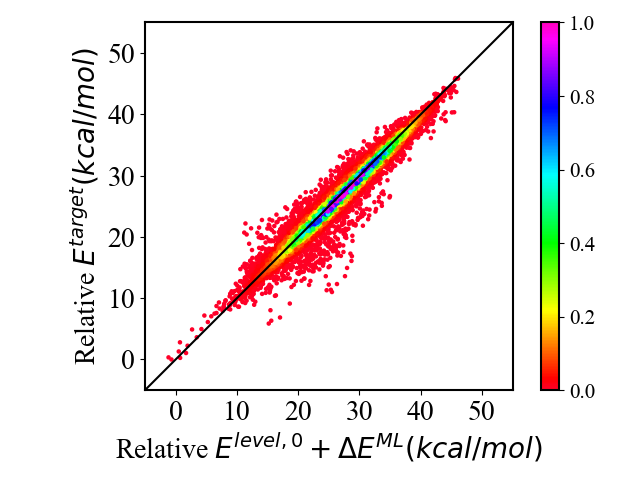}}

    \caption{The protonated 21 water cluster potential energy error distribution similar to Figure \ref{fig_final-main2} but only larger fragments are transferred as discussed in Figure \ref{fig_combined}.}
    \label{fig_com_dis}
\end{figure}
Therefore,
we examine the potential energy predictions for the protonated 21-water cluster when the transformed neural networks are 
only used 
for larger fragments. The error is displayed in Figure \ref{fig_combined} where the first blue bar corresponds to 
the MAE from the green bar for 
${\cal R}$=3 in Figure \ref{fig_final-main-summary}. As seen from comparing the blue and orange bars in Figure \ref{fig_combined}, omitting $H_2O$ and $H_3O^+$ fragments from the modification process leads to negligible differences in MAE. 
This result is also consistent with Table \ref{tab_initial_error}. However, excluding $H_4O_2$ results in an increase in MAE 
from 0.36 kcal/mol to 0.65 kcal/mol as shown by the green bar in Figure \ref{fig_combined}. This is because, although $H_4O_2$ has a low error of 0.06 kcal/mol when the solvated Zundel fragment neural networks are used, the population, $\omega_{f_r,{\cal R}}$ values in Table \ref{tab_initial_error} are large. Thus, excluding these fragments in  Figures and \ref{fig_combined} has a significant effect. A similar behavior results 
when $H_5O_2^+$ is excluded, as shown by the red bar in Figure \ref{fig_combined} due to the larger error from these fragments as seen in Table \ref{tab_initial_error}, that is significantly reduced due to the modification of neural networks for  these fragments as seen from Table \ref{tab_all}. 

To further investigate the contribution of $H_4O_2$ and $H_5O_2^+$ to the total error in Figure \ref{fig_com_dis}, we provide 
the 
distribution of energies for the protonated 21 water cluster when these two species are excluded from the transferring process. The corresponding mean absolute errors are represented by the green and red bars in Figure \ref{fig_combined}. A comparison between Figure \ref{fig_com_dis} and Figure \ref{fig_final-main2}(d) reveals that the 
error distribution has grown broader and more diffuse in Figure \ref{fig_com_dis}. At the same time, the broadened error distribution displays a symmetric nature along the central diagonal. This observation suggests that the errors associated with $H_4O_2$ and $H_5O_2^+$ may have a probabilistic nature, and thus perhaps less 
complicated as compared to 
off-axis shifts and deformations in Figure \ref{fig_final-main}(d). Thus, future work may benefit from stochastic methods 
to further reduce the amount of training samples required for such fragments.

\subsection{Computational cost associated with graph theoretic  neural network models for post-Hartree-Fock accuracy}\label{sec_cost}
\begin{figure}
    \centering
    \includegraphics[width=0.8\columnwidth]{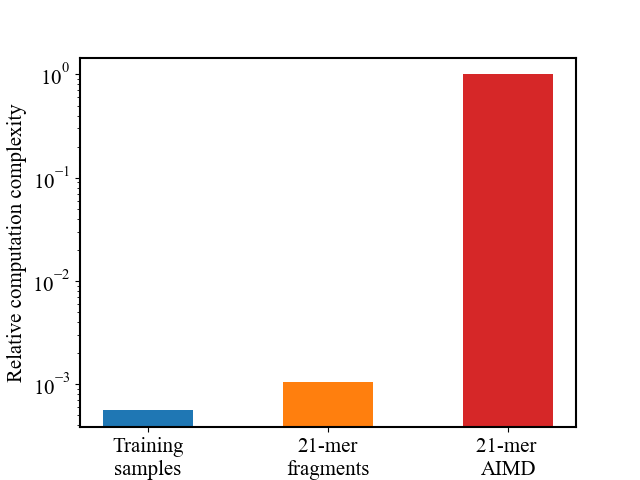}
    \caption{Relative computational cost for CCSD energies for all training samples, fragments from the protonated 21 water cluster trajectory, and the full system potential energies of protonated 21 water clusters. 
    Training samples shown in blue bar involve both primitive and additional training sets. Fragments shown by the orange bar are produced from the protonated 21-water cluster trajectories.}
    \label{fig_cost}
\end{figure}
Here we summarize the total cost for constructing neural network potential energy surfaces and compare it with direct evaluation of fragment CCSD energies and full system CCSD energies. The total training cost reported here includes both fragments obtained from the solvated Zundel trajectory as well as those appended through incremental  orthogonal training as dictated by the geometries found in the 21-mer trajectory. The resultant training data represents approximately 32\% of all fragment geometries obtained from the solvated Zundel and protonated 21 water cluster trajectories. 

Since the goal is to reduce the computational effort to allow higher quality electronic structure, we discuss the formal computational complexity.
In order to quantify the discussion, we will assume that the target molecular system is divided into ${\cal N}$ identical nodes each with basis set size of ${\cal F}_0$ which determines the computational complexity for the molecular fragment associated with each node.
Thus the size of basis set for the full system is roughly $[{\cal N}*{\cal F}_0]$.
A computationally sequential (or serial) treatment of the problem leads to evaluation of each rank-$r$ term at two levels of theory. If we further assume that the $Ref.$ calculation scales as ${\cal O} \left( N^{L_0} \right)$ and $target$ scales as ${\cal O} \left( N^{L_1} \right)$, then the overall $target$ calculation scales as ${\cal O}\left( ({\cal N}{\cal F}_0)^{L_1} \right)$. The fragment calculations then scale as 
\begin{align}
{\text{Scaling}}_{Eq. (\ref{eq_graph-ML-main-delta})} &\approx 
\sum_{r=0}^{\cal R}  { \sum_{\alpha_r  \in {\bf V}_r} {\cal M}_{\alpha_r, r}^{\cal R} \;  {\cal O} \left( (r{\cal F}_0)^{L_1} \right)
} \nonumber \\ &= \sum_{r=0}^{\cal R}  {\cal O} \left( (r{\cal F}_0)^{L_1} \right)
\left[ \sum_{\alpha_r  \in {\bf V}_r} {\cal M}_{\alpha_r, r}^{\cal R} \right]
\nonumber \\ &\approx {\cal O} \left( ({\cal R}{\cal F}_0)^{L_1} \right) N_{\cal R}
\label{graph-scaling}
\end{align}
where $N_{\cal R}$ is the number of rank-${\cal R}$ fragments. The calculations here are performed in an asynchronously parallel fashion\cite{Harry-weighted-graphs} and hence the scaling in Eq. (\ref{graph-scaling}) is further reduced. 
The serial version of the expression in Eq. (\ref{graph-scaling}) is shown using the orange histogram in Figure \ref{fig_cost}, whereas the quantity, ${\cal O}\left( (NM)^{L_1} \right)$ is represented in red for the protonated 21-water cluster system. As can be seen there is significant reduction in computational effort even in the absence of exploiting parallelism. 
The histogram in blue represents the portion of the orange histogram that is used for training here in constructing the machine learning models. As noted above this is roughly 32\% of the orange histogram and also does not represent the available parallelism in developing neural networks simultaneously for multiple fragments. 

While the discussion above provides general guidelines and argue the power of our approach, we next provide more precise scaling costs that accurately present the systems treated here. 
To estimate and compare the costs, we use CCSD scaling cost\cite{Schlegel-Bottlenecks} of $N^6$ for $N$ electrons, and the represents the item shown in red in Figure \ref{fig_cost} for all 21-mer data.  
Based on this scaling, the protonated 21 water cluster has 211 electrons, leading to a total computational cost of $211^{6}*28294=2.5*10^{18}$ for all 28294 geometries in the trajectory. This value is represented by the red bar in Figure \ref{fig_cost} and normalized as unity for ease of comparison. 

To evaluate the cost for all fragment energies obtained from the protonated 21 water cluster trajectories, we take the number of fragments of each kind listed in Table \ref{tab_num_of_geo}, as already seen from Eq. (\ref{graph-scaling})  and multiply by the factor ${\cal F}_r^6$ for ${\cal F}_r$ electrons for rank-$r$ fragments. 
For example, we found 559877 water fragments each with 10 electrons in the protonated 21 water cluster trajectory. Then the cost for all water fragments becomes $10^6*559877 = 5.6*10^{11}$. Similarly, there are 34297 $H_3O^+$ geometries found each with 11 electrons. The total cost for all $H_3O^+$ becomes $11^6*34297 = 6.1*10^{10}$. When we repeat the same process for all types of fragments
listed in Table \ref{tab_num_of_geo}, the total fragment cost is the sum of cost for each type of fragments as $5.6*10^{11}+6.1*10^{10}+\dots=2.6\times10^{15}$. 
The total fragment energy cost is three orders of magnitude smaller than the full system cost, indicating a significant computational saving achieved by applying Eq.\ref{eq_graph-ML-main}. This is because the largest major fragments obtained from the protonated 21 water cluster are $H_{9}O_4^+$ which contains only 41 electrons far fewer than the 211 electrons in the full systems.

Similarly, the blue bar in Figure \ref{fig_cost} illustrates the total cost of all fragments used for training our neural network models, which involve both primitive and additional training sets obtained from the two trajectories. For example for water fragments, as can be seen from Table \ref{tab_all}, we use 3891 geometries from the solvated Zundel trajectory for training an initial set of neural network models as described in Section \ref{solZ-results}. Another 16000 water geometries from the protontaed 21 water cluster trajectory is included as the additional training set to adapt the neural networks to the orthogonal regions of the fragment space. Then the total training set cost for water neural network model is computed as $10^6*(3891+16000)=2.0*10^{10}$. We again sum over all fragments that are modeled by independent neural networks and plot as the blue bar in Figure \ref{fig_cost}. These results provide an intuitive demonstration of how Eq. (\ref{eq_proj-Q}) effectively reduces the prohibitive computational cost for direct full system mapping $\cal Q$ to a family of much smaller and tractable mappings $\left\{ {\cal Q}_{\alpha_r,r} \right\}$.


\section{conclusion}\label{sec_conclusion}
The cost of acquiring sufficient training samples to construct accurate neural network models for large system potential energy surfaces at the couple-cluster level of theories has presented a great challenge to computational chemistry. In this paper, we present a transfer learning protocol which extends from a smaller known (learned) system, in our case the solvated Zundel system, to predict the potential energy surface for a much larger system, in our case the protonated 21-water cluster. Thus, we 
construct learning models appropriate for a 51-dimensional space (solvated Zundel) and adapt those models to extrapolate the quantum mechanical energies for a 186-dimensional space (the protonated 21-water cluster). 

Towards this, we begin by using our previously developed graph-theoretic molecular fragmentation procedure for configuring common fragments from both systems. Then, we construct a set of neural network models that learn on the energies from each type of fragment separately from a solvated Zundel library of structures. To better represent the graph-theoretically generated fragment geometries and understand their differences, we designed a permuted inter-atomic distance vector that follows translational, rotational, and permutation invariance as the descriptor. These neural network models provide highly accurate predictions on the solvated Zundel potential energies at an MAE of 0.36 kcal/mol, but lose accuracy when directly used for predicting energies for the protonated 21-water cluster with an MAE of 5.32 kcal/mol due to the change in distribution of fragments structures. Thus, we design a modified, incremental and directed transfer learning process that first distinguishes the extent of fragment structural differences between the two systems. 
Once these differences are understood, an incremental learning process yields results with sub-kcal/mol accuracy of the full potential energy surface of the protonated 21-water cluster. 
We obtain highly accurate potential energy predictions for a protonated 21-water cluster dynamics trajectory, with MAE of 0.36 kcal/mol 
at CCSD level while also maintaining good accuracy for the initial solvated Zundel system using the same family of neural network models.

The graph-base representation provides a robust foundation for treating different fragments independently, making a key deviation from the architecture used in transformer-based large language models. In transformers, all sentence components are processed uniformly through multiple sequential layers of attention computation, resulting in a deep stacking architecture. In contrast, our method creates a 
distributed set 
of independent neural network blocks, corresponding to individual fragment species. We consider this architectural distinction an advantage that allows us to focus on the more 
critical components of the system. Here, importance not only refers to the attention score of a word or the multiplicity of a molecular fragment, as discussed here, 
but also includes the chemical significance of a fragment by incorporating many body interactions. 

\section{Acknowledgment}
This research was supported by the National Science
Foundation grant CHE-2102610 (Creativity Extension) to SSI. The computational facilities at Indiana University are duly acknowledged and have been critical to this effort. Acknowledgment is due in part to the Lilly Endowment, Inc., for their support of the BigRed computing facility at Indiana University widely used in the effort represented in this publication. This work was also supported in part by Shared University Research grants from IBM, Inc., to Indiana University, which supports the Scholarly Data Archives.


\appendix

\section{Illustration of Figures \ref{Sets-illustration} and \ref{fig_graph_complex} for bonded and non-bonded systems}
\label{partitioning}
\begin{figure}
\includegraphics[width=\columnwidth]{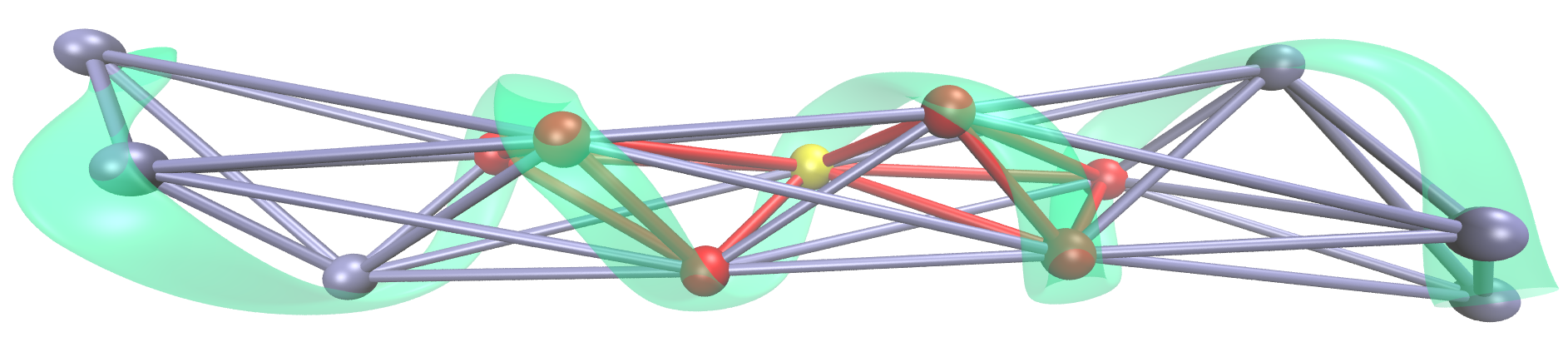}
\caption{\label{Poly-ala} 
In Refs. \onlinecite{CGAIMD,frag-BSSE-AIMD}, a poly-alanine system is divided into simplicial complexes as discussed in Section \ref{graphs}. This is a bonded system and here nodes are alanine fragments. When bonds are cut to create fragments, these terminated using link atoms to satisfy valency as done within the ONIOM protocol of mechanical embedding\cite{oniom}. The figure shows higher order simplexes and, for example, the node in yellow as interactions that are part of several fragments. 
}
\end{figure}
\begin{figure}
\subfigure[]{\includegraphics[width=0.8\columnwidth]{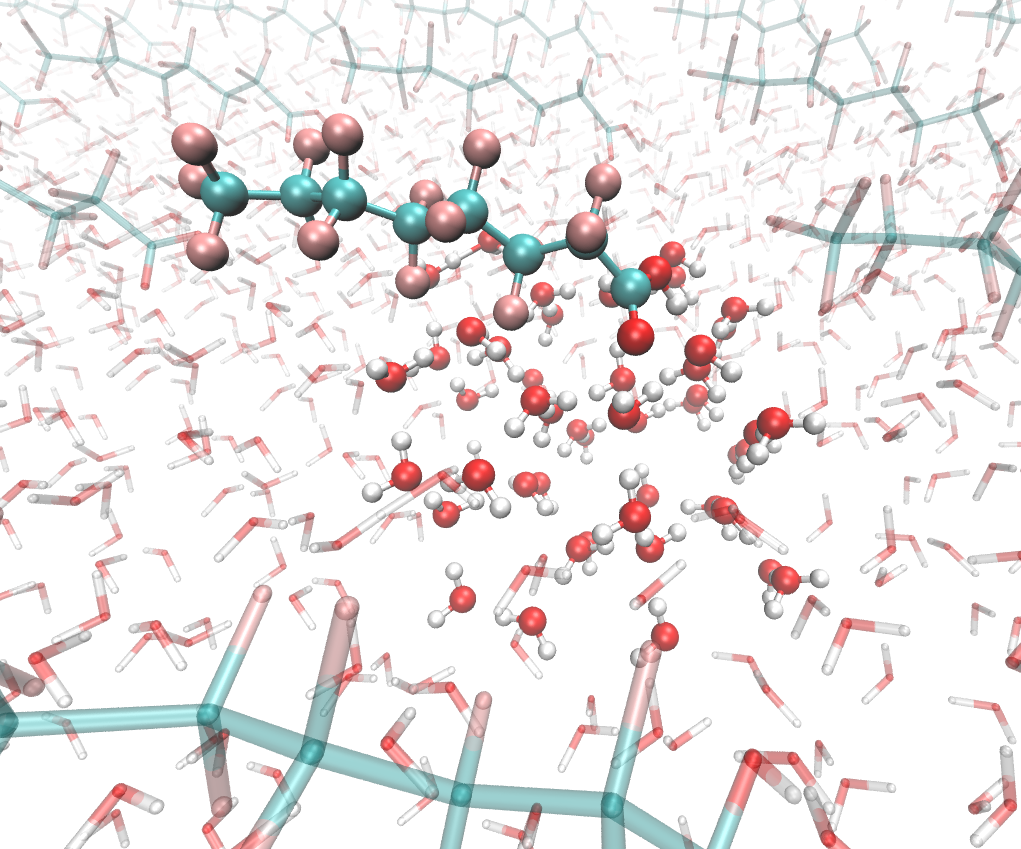}}
\subfigure[]{\includegraphics[width=0.8\columnwidth]{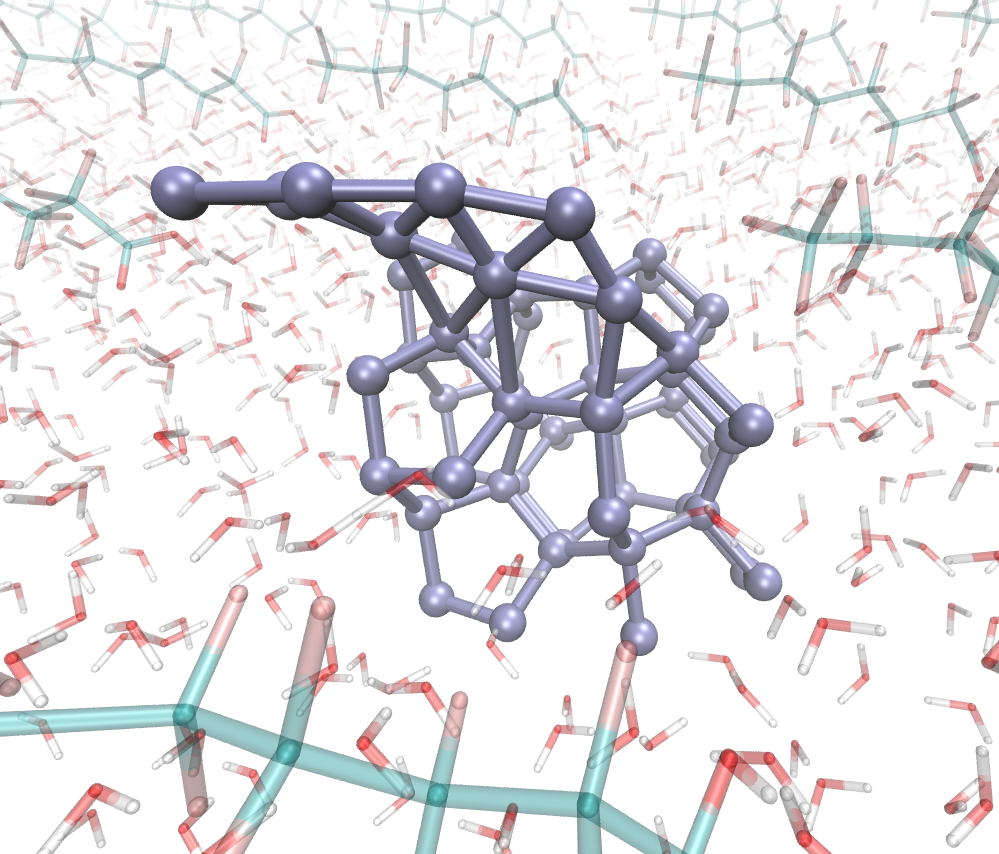}}
\caption{\label{PFOA} 
Similar to Figure \ref{Poly-ala} but for a heterogeneous, solvated system\cite{frag-PFOA}. Interactions between the organic substrate and water are captured at multiple levels through the use of simplicial complexes. Figure (b) shows the graph representation of Figure (a), where higher order simplexes are clearly seen.
}
\end{figure}
In Refs. \onlinecite{CGAIMD,frag-BSSE-AIMD,frag-PFOA} we have used the partitioning scheme in Section \ref{graphs} to study both bonded as well as non-bonded systems. In these studies coupled cluster accuracy is obained at DFT cost from the direct use of the graph-theory approach discussed in Section \ref{graphs}. We  illustrate these here for completeness. In Ref. \onlinecite{CGAIMD} we divide a protein fragment, a poly-alanine chain shown in Figure \ref{Poly-ala}, into nodes, edges, faces and higher rank objects, where each amino acid fragment is chosen as a separate node fragment. As a result edges are dimners, faces are trimers (bonded and non-bonded), and so on. Likewise, in Ref. \onlinecite{frag-PFOA}, we consider asolvation study of per-fluoro-octanoic acid (PFOA), an environmental pollutant, on a periodic slab of water. Here, the partitioning is complicated by additional interactions, and larger rank simplexes involve interactions of the organic system with water. The scheme is presented in Figure \ref{PFOA}. In all cases, when chemical bonds are broken to create fragments, the dangling bonds are passivated through link atoms as is routinely done in the ONIOM protocol\cite{oniom}. This kind of passivation is called mechanical embedding. 

\section{Mini-batch-k-means sampling and initial neural network model setup for the solvated Zundel}\label{sec_kmeans_nn}
To generate the initial neural network model for solvated Zundel fragments, we begin by collecting all fragments from the solvated Zundel trajectory and create a data bank. For each type of fragment, we use mini-batch-k-means for 10\% sampling. The mini-batch-k-means algorithm is a variant of the clustering algorithm k-means clustering, which essentially tessellates or divides the data space into $k$ mutually exclusive regions called clusters $\{C_j\}$. Every cluster is represented by a centroid $\bf{\bar r}_j$ which is computed as the arithmetic mean of all data points $\left\{ \bf{r}_i \right\}$ when these are assigned to the closest centroid. 
The quantities $\left\{ \bf{r}_i \right\}$ refer to interatomic distances within fragments and these are specifically ordered as per the recipe described in Appendix \ref{sec_descriptor}. 
The k-means algorithm aims to find a preset number of centroids or clusters iteratively to minimize the cost function, 
\begin{align}
    \sum_{j=1}^{k} \sum_{\bf{r}_i \in C_j} |{\bf{r}_i-{\bf{\bar r}}_j }|^2.
    \label{kmeans-cost-fn}
\end{align}
The Mini-batch-k-means algorithm provides an efficient approach by using only a random subset of data (known as batch) to update the centroid positions during each iteration. After finding the centroid positions, we construct the primitive training set as the set of closest data points to every centroid
\begin{align}
    \{\bf{s_{j}}\} = \left\{ \bf{r}_i \left\vert \min_{\bf{r}_i} \abs{\bf{r}_i-{\bf{\bar r}}_j} \right. \right\}.
    \label{eq_sj}
\end{align}
The data points $\{\bf{s_{j}}\}$ are now the closest data points to each centroid, $\bf{r}_i$ from the closest centroid $\bf{\bar r}_j$ and represents our training data points represented below as  $x_{n=0}$.

In this paper, we place two neural networks in each model as the array and each neural network is configured with 4 hidden layers. Each hidden layer consists of the number of nodes as 4 times the number of features or the number of inter-atomic distances introduced in Appendix \ref{sec_descriptor}. The activation function on hidden layers has been modified to a Gaussian function 
\begin{align}
    x_{n;a} = exp(-(w_{n-1,n;a} \cdot x_{n-1})^2)
\end{align}
where $x_{n;a}$ is the $a$-th element of the layer vector $x_n$ and $w_{n-1,n;a}$ represents the $a$-th row of the weight matrix $w_{n-1,n}$ connecting layers $n$ and $n-1$. The quantity, $(w_{n-1,n;a} \cdot x_{n-1})$ is the dot product of the $a$-th row of the weight matrix $w_{n-1,n}$ with the vector $x_{n-1}$ corresponding to the $(n-1)$-th layer. Thus, each hidden layer vector is an exponent of an element-wise square of the previous layer vector. During the training process, a bias of ``1'' is appended to each vector in $\{\bf{s_{j}}\}$ from Eq. \ref{eq_sj} as the inputs for neural networks, and hence
\begin{align}
    x_{n=0} &= \begin{bmatrix}{\bf{s_j}}\\1\end{bmatrix} 
\end{align}
The bias with its associated neural network weights is used for optimizing the positions of gaussian functions during training.
The gaussian type function is useful for reducing the long range effect during transferring so that the currently training slice does not interfere much with the previously learned slices of data space. 

The neural network model that we used to fit the fragment energy pattern is a neural network array. This array consists of a series of neural networks each of which learns on the cumulative error on predictions from prior neural networks in the series. So that for $M$ neural networks in the series, the final prediction is
\begin{align}
    \Delta E^{ML} = \sum_{m}^{M}\Delta E^{ML,m}.
\end{align}

\section{A consistent, permutationally invariant, fragment moment of inertia frame of reference for describing fragment geometries and descriptors}
\label{sec_descriptor}

We intend to introduce a consistent framework for description of fragment geometries that arise from different molecular systems. 
Such a descriptor should allow for translational, rotational, and permutational invariance within the fragments, so that chemically identical fragments derived from different systems, or such fragments derived from different physical regions of the same system, have the same consistent mathematical description and thus can be compared with each other.

\begin{figure}
    \centering
    \includegraphics[width=0.99\linewidth]{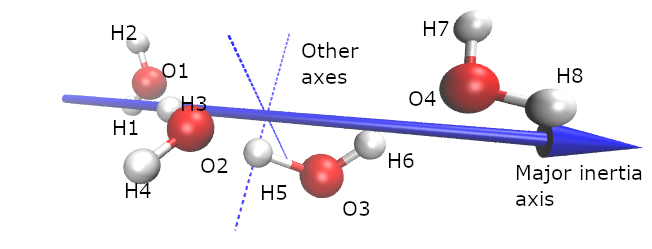}
    \caption{An illustration of the moment of inertia frame of reference and the corresponding atom ordering for a four water wire fragment. The major axis is the mass moment of inertia axis with the smallest absolute mass moment of inertia. The dash lines represent the two other inertia axes.}
    \label{fig_axis}
\end{figure}

We begin with the inter-atomic distance matrix of fragments as descriptors but then recognize that the ordering of atoms between fragments may not be chemically consistent. For example, let us consider a water-wire fragment (${\left(\text{H}_\text{2}\text{O}\right)}_n$) fragment which is found in multiple chemical systems and has a critical role in proton transfer in water clusters and in condensed phase systems. Hence a fragment of this kind can be found in most water cluster systems. An example of such a system can be found in Figure \ref{fig_axis}. This specific system contains eight protons, but all eight protons do not have the same chemical environment and properties. While five protons are on the periphery, three protons are shared between two oxygen atoms. Additionally, the shared protons themselves have different secondary environments. For larger water clusters these differences become further complicated where our chemical description of the cluster must maintain a distinction between the shared protons on the outer layers of the cluster and those on the inner layers. To arrive at a consistent notation that retains the chemical meaning of each atom, we use the following steps
\begin{packed_enum}
\item We first rotate the fragment geometry coordinates into 
center of mass coordinates and mass moment of inertia frame of reference. In Figure \ref{fig_axis}, the large blue axis refers to the axis of the major moment of inertia, that is smallest inertia value. 
\item Following this the atoms are ordered in increasing order of atomic weights, and then 
\begin{packed_enum}
\item increasing order of coordinate projection along the major moment of inertia axis. 
\item For cases where the major axis projected value is identical, the second major axis projection value is used to resolve the order of atoms. 
\end{packed_enum}
\item We then arrive at an ordered list of atoms for any given fragment that consistently represents the chemical environment that the specific atom resides within. 
\item After permuting all atoms based on the relative positions along the fragment moment of inertia axis, the distance matrix is computed and reshaped into a vector to define the fragment descriptor.  These fragment descriptors are referred to as ``$r_j$" in Appendix \ref{sec_kmeans_nn}. 
\end{packed_enum}

\begin{table}[h!]
\centering
\begin{tabular*}{\columnwidth}{@{\extracolsep{\fill}}lcccl}
\hline  \hline
Fragments& $D$ (\AA)   & $D$ (\AA)    \\ 
& without reordering & with reordering \\ \hline
$H_2O$ &	0.06& 0.05\\ 
$H_3O^+$ &	0.21& 0.19\\
$H_4O_2$&	2.88& 2.06\\
$H_5O_2^+$&	4.93& 2.86\\
$H_6O_3$ &	5.12& 3.72\\
$H_7O_3^+$ & 6.97& 4.14\\
$H_8O_4$ &	7.00& 5.55\\
$H_9O_4^+$ & 8.26& 7.07\\
\hline  \hline
\end{tabular*}
\caption{The inertia span in Eq. (\ref{eq_D}) with and without a permutation invariant reordering introduced by Appendix \ref{sec_descriptor}. The $D$ value is computed from Eq. (\ref{eq_D}) for fragments obtained from the protonated 21 water cluster trajectory.}
\label{tab_geo_diff}
\end{table}

This procedure establishes a permutation invariant ordering of atoms within each fragment.
This modification improves the consistency of fragment representations across different systems by better preserving geometrical symmetry. Consequently, the number of training samples required and the training cost are both reduced. The effect can be seen by comparing the geometry span defined by an average inertia given by, 
\begin{align}
    D = \frac{1}{N}\sum_{i}|\bf{r}_i-\bf{\bar{r}}|^2
    \label{eq_D}
\end{align}
where $\bf{r_i}$ is the $i$-th interatomic distance vector, $\bf{\bar{r}}$ is the average of all $\bf{f_i}$ for a total of $N$ geometries. In Table \ref{tab_geo_diff}, we list the $D$ values for all fragments generated from the protonated 21 water cluster trajectory with and without this ordering. As we can see from the table, such an ordering effectively reduces the spanning for most larger fragments.

It must be noted that permutational invariance is part of a broader class of requirements that arise from group theory. Specifically, for systems that obey certain group structures, the neural network must also transform according to the irreducible representations of that group structure. This is essentially the goal of equivariant NNs\cite{Equivariant-NNs-1,Equivariant-NNs-2}. In our work here, we use a  pragmatic approach to enforcing permutational invariance. 
Equivariance is preserved in our case through the mass moment of inertia axis projection of coordinates for each fragment (obtained from the simplicial complex). 

\section{Distinctions between challenges in quantum chemistry-based potential energy surfaces and LLM models}\label{sec_disctinction}
Although languages and molecules can be represented as graphs and may have a similar overarching strategy, the inherent differences between languages and molecules introduce challenges during different stages of the solution process. For molecules, constructing the graph is straightforward, and the weights can be calculated with well-defined inclusion and exclusion relationships. In contrast for language, the relationship between words is more complicated and abstract. Their relationship depends not only on their relative position in sentences but also on their meaning, coherence, and logical reasoning connections. Therefore, the attention is computed and optimized to understand the whole sentence through multiple layers of transformer architectures. 

On the other hand, the primary challenge for molecules lies in obtaining accurate fragment-level information, specifically the fragment energies. Unlike words in language processing, which benefit from clear contextual information available from large and easily available language datasets for training, fragment energy modeling requires extensive and computationally expensive post-Hartree Fock calculations to generate training data. Thus, the neural network models used for fragment energy embedding often have limited accuracy and generalizability. For example, as we demonstrate in this paper, models trained on fragments from solvated Zundel cannot be directly applied to the same sets of fragments from the protonated 21-water clusters. This limitation motivates the development of our systematic approach to incrementally expand the training space for fragment energy modeling.

From another perspective, a key feature of the transformer architecture is its ability to iteratively refine the representation or embedding of each word or token based on its contextual environment (sentences). In each layer, the attention computed from Eq. (\ref{eq_attention}) followed by neural network transformations and residual connections are used to update the {\em query}, {\em key}, and {\em value} vectors for the next iteration of attention computation. Therefore, the same set of words can play different roles in different sentences. This refinement process is philosophically aligned with our approach, following Eq. (\ref{refine}) to update fragment energy representations from $\Delta E_{\alpha_r,r}^{ML, 1}({\bf {\bar x}_{\alpha_r,r}})$ to $\Delta E_{\alpha_r,r}^{ML, 2}({\bf {\bar x}_{\alpha_r,r}})$ based on the full system configuration, allowing local fragment properties to evolve in response to the global molecular information.

\section{Protonated 21 water clusters open new hydrogen transferring paths compared to solvated Zundel}
\label{sec_H_path}

\begin{figure*}
\centering
    \subfigure[H$_5$O$_2^+$]{\includegraphics[width=0.32\linewidth]{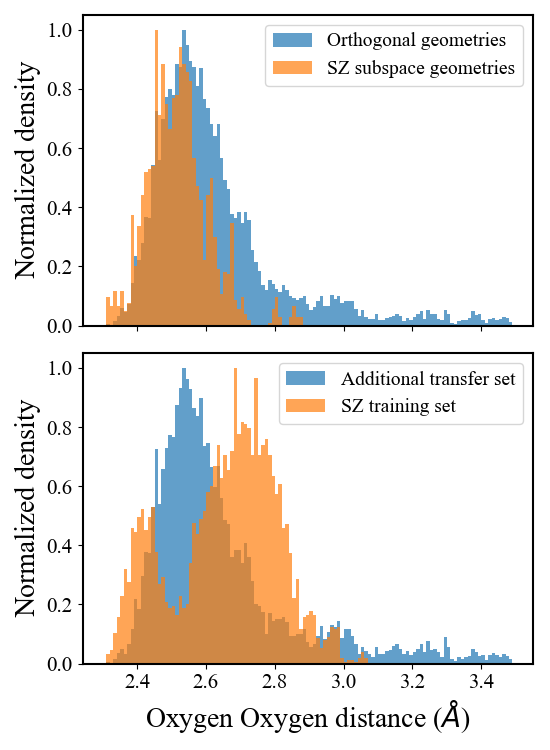}}
    \subfigure[H$_7$O$_3^+$]{\includegraphics[width=0.32\linewidth]{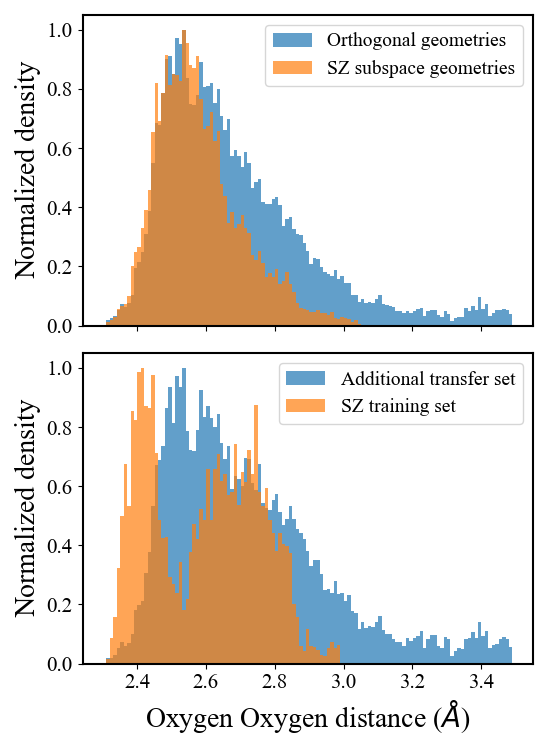}}
    \subfigure[H$_9$O$_4^+$]{\includegraphics[width=0.32\linewidth]{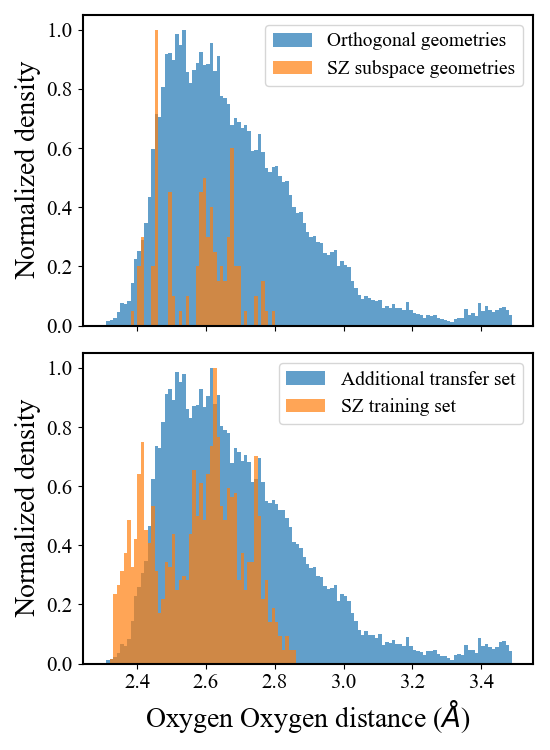}}
\caption{{The top panels show the distribution of OO distances for the H$_5$O$_2^+$ (a),  H$_7$O$_3^+$ (b) and the H$_9$O$_4^+$ (c) fragments from within the protonated 21-mer system. Specifically, the orange trace in the top panel includes those structures from the protonated 21-mer system that are accurately represented using the neural networks designed from solvated Zundel database. The blue traces in the top panels need additional incremental training using the methods discussed in the paper. Clearly as the fragment size increases, the structures deviate from the solvated Zundel database in a big way reflecting the behavior shown in Table \ref{tab_all}. For the bottom panels, the orange traces represent the pre-trained structures from solvated Zundel and the blue traces represent those that need additional training. As the cluster sizes grow the deviation in OO distance sampled is clear. Since the OO distance is one major indicator of proton hops, the protonated 21-mer clearly samples a wider range of possibilities. 
\label{fig1}}}
\end{figure*}

Associated with the  differences in distribution noted in Figure \ref{fig_full_rdf}, the fragment energy surfaces obtained from within the protonated 21-water cluster and solvated Zundel system also show critical differences. The H$_5$O$_2^+$ fragments from the solvated Zundel database span the orange trace in the bottom panel of Figure \ref{fig1}(a)). As discussed in Section \ref{solZ-results}, the initial family of neural network models were constructed for this distribution of structures and faithfully represent the solvated Zundel system. However, when the protonated 21-water cluster system is considered, the H$_5$O$_2^+$ fragments, display the behavior in Figure \ref{fig1}(a)). 
The top panel in in Figure \ref{fig1}(a) shows the distribution of OO RDF for the H$_5$O$_2^+$ fragment geometries obtained from the protonated 21-mer system whose energies can be determined from the solvated Zundel database (orange) and those that cannot with suitable accuracy (blue). Thus the blue distribution is the orthogonal counterpart which we incrementally learn from based on the discussion in Section \ref{transformer-SI}. While the difference may be thought to be marginal for the   H$_5$O$_2^+$, as seen from Table \ref{tab_all}, this still results in a 0.15 kcal/mol error due to the number of H$_5$O$_2^+$ in the larger water cluster. This error is reduced to 0.02 kcal/mol, when the additional distribution (blue) is incrementally added to the neural networks created from the solvated Zundel database. 
\begin{figure}
    \subfigure[H$_7$O$_3^+$]
    {\includegraphics[width=0.45\linewidth]{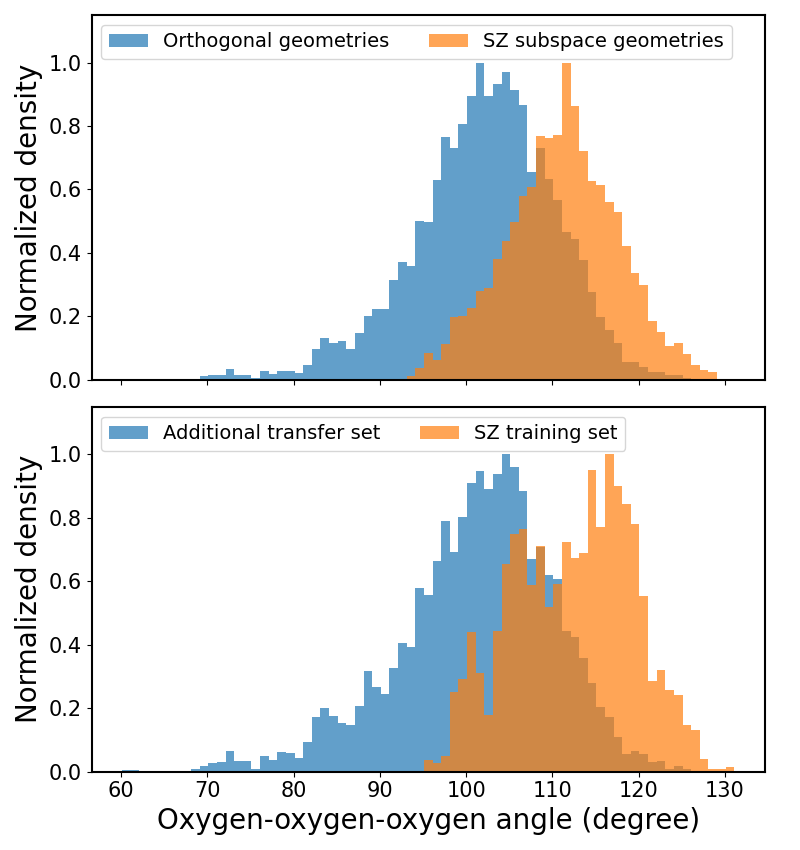}}
    \subfigure[H$_9$O$_4^+$]
    {\includegraphics[width=0.45\linewidth]{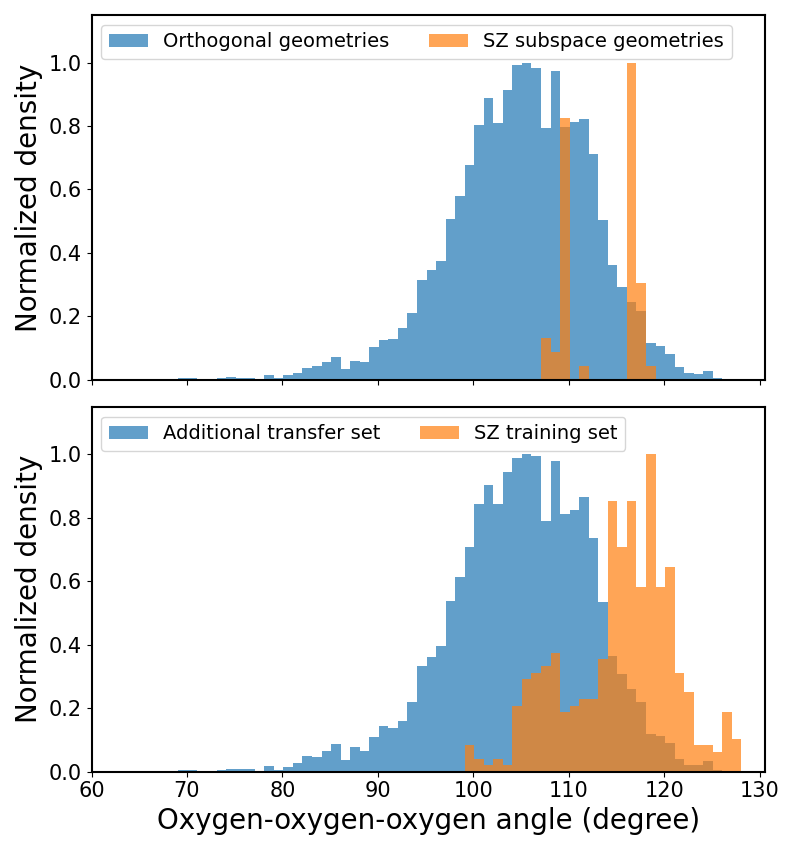}}
\caption{{Similar to Figure \ref{fig1}, but OOO angle distribution for all connected trimer and quartermer subsystems. It is believed\cite{Zundel1969hydration,cptuckerman3, h+oh-solv, schmittvothprotontransport1999, schmitt1998multistate} that the surrounding hydrogen bonded water molecules (around the protonated species in water) facilitate proton transfer and in this figure we probe how this distribution changes with system size. 
\label{fig2}}}
\end{figure}
These differences become even greater for the larger clusters, as seen for H$_7$O$_3^+$ (Figure \ref{fig1}(b)) and 
H$_9$O$_4^+$ (Figure \ref{fig1}(c)). 
Thus, the larger clusters provide additional flexibility to the protonated fragments resulting in (a) more dynamical pathways for proton-hopping, and (b) distinct contributions to the overall energy of the system as seen from Figure \ref{fig_combined}. 

In Figure \ref{fig2} we present the OOO angle distribution which also clearly outlines the differences in geometries of the smaller fragments as the overall system size grows. The orthogonal incremental tesselation algorithm samples these complementary regions of data-space. Figure \ref{fig2}(a) shows the distribution of OOO angles for the H$_7$O$_3^+$ and Figure \ref{fig2}(b) shows the same for H$_9$O$_4^+$ fragments from within the protonated 21-mer system. Clearly, these angle distributions deviate much more than the OO distributions in Figure \ref{fig1}, reflecting the behavior already shown in Table \ref{tab_all} and the need for incremental training as discussed. Importantly, the larger spread of angles seen in Figure \ref{fig2} for the protonated 21-water cluster occur from secondary solvation shell interactions. These secondary solvation shell effects are present within our graph-based formalism.  

\section{The recursive mini-batch-k-means clustering within the slices}
\label{slicing}
The recursive mini-batch-k-means intends to create clusters with sizes approximately the same as those from the 10\% mini-batch-k-means on the primitive data (Slice 1). From Ref. \onlinecite{frag-ML-Xiao} and Table \ref{tab_initial_error}, we conclude that neural network models trained on the 10\% training data, where the training data is obtained from k-means clustering, are highly accurate and represent solvated Zundel potential within sub-kcal/mol accuracy. 
Based on this training data for the primitive, solvated Zundel data set,
we compute the average inertia ($\eta_0$) to measure the average cluster size for the 10\% clustering process and use it as the target for additional slices clustering.
The average inertia is computed as
\begin{align}
    \eta_0 = \frac{1}{M}\sum_a^k\sum_{\bf{r_i}\in C_j}\abs{{\bf{r_i}}-\bf{\bar{r}_j}}^2
    \label{eq_eta}
\end{align}
where $M$ is the number of primitive data points used for clustering the fragments found in the solvated Zundel trajectory, $k$ is the number of training data points and also represents the number of k-means clusters or regions into which the $M$ data points are divided, $C_j$ are the (irregular and multi-dimensional) regions, and also set of points within the $j-th$ cluster with 
$\bf{\bar{r_i}}$ being the centroid of the region $C_j$.

\begin{figure}
    \centering
    \subfigure[Slice 1 partitioning]{\includegraphics[width=0.48\columnwidth]{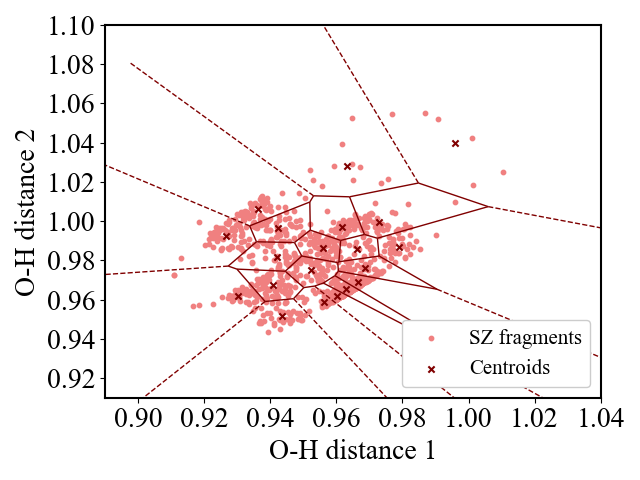}}
    \subfigure[Additional slices in the space]{\includegraphics[width=0.48\columnwidth]{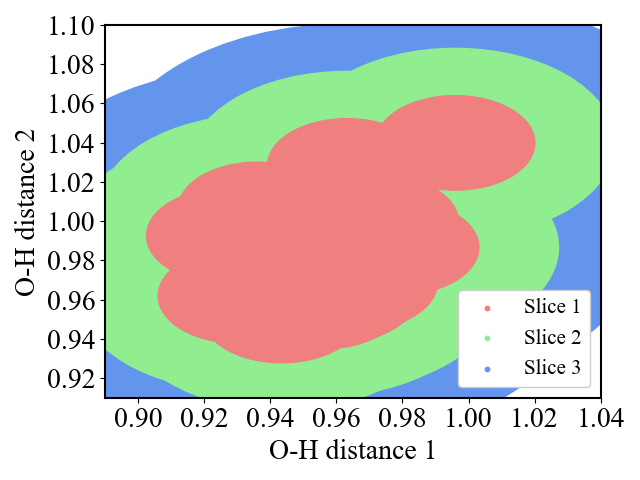}}
    \subfigure[Round 1 coarse partitioning of slice 2]{\includegraphics[width=0.48\columnwidth]{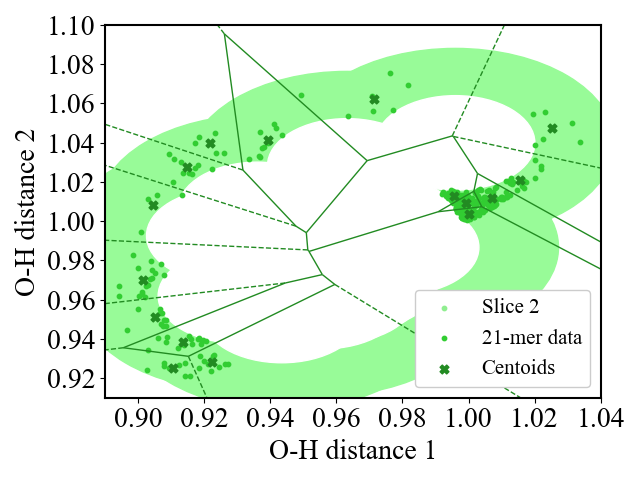}}
    \subfigure[Final partitioning of slice 2]{\includegraphics[width=0.48\columnwidth]{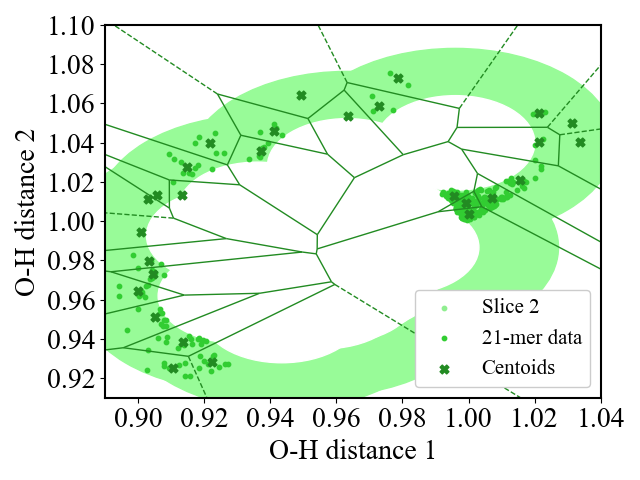}}
    \caption{The process of recursive mini-batch-k-means for additional slices partitioning.}
    \label{fig_skmean}
\end{figure}

For example for the $H_2O$ fragment from within the solvated Zundel data as described in Figure \ref{fig_skmean}(a), $\eta_0$ is computed from the red lines partitioning the space, with the $\bar{m}_a$ geometries represented as ``X". We aim to create similar tessellations for additional slices like those within the green area and blue area in Figure \ref{fig_skmean}(b). We begin with a coarse partitioning of data by using a mini-batch-k-means clustering with the number of clusters chosen as the square root of the number of samples in the slice as shown in Figure \ref{fig_skmean}(c). We label this clustering process as round 1 ($j=1$) and compute the cluster-wise average inertia similar to Eq. \ref{eq_eta} but only for data within the $s$-th slice, $l$-th cluster, iteration round number $j$ as 
\begin{align}
     \eta_{s,j,l} = \frac{1}{M_{s,j,l}}\sum_{\bf{r_i}\in C_{s,j,l}}\abs{{\bf{r_i}}-\bf{\bar{r}_{s,j,l}}}^2.
     \label{eq_eta1}
\end{align}
where 
$M_{s,j,l}$ is the number of sample points in the $s$-th slice, $l-th$ cluster, for iteration round number $j$. As shown in Figure \ref{fig_skmean}(c), some clusters are too large, such as the top and rightmost. The value of $\eta_{s,j,l}$ for every cluster is used to measure the size of the cluster and compare it with the target average inertia of $\eta_0$. If $\eta_{s,j,l} > 2*\eta_0$, we perform a second round ($j=2$) of mini-batch-k-means clustering on the data within the $s$-th slice, $l-th$ cluster requiring that the $l-th$ cluster be further sub-divided into the following number of the cluster:
\begin{align}
    k_{s,j+1,l} = \frac{\eta_{s,j,l}}{\eta_0}.
    \label{eq_inertia_cutoff}
\end{align}
Then we compute cluster-wise average inertia from Eq. \ref{eq_eta1} for the round 2 clusters ($j=2$) and compare their $\eta_{s,j,l}$ with $2*\eta_0$ again. This recursive process continues until all $\eta_{s,j,l}$ for clusters are below the threshold $2*\eta_0$ and ultimately lead to space partitioning shown in Figure \ref{fig_skmean}(d). Here we can clearly see that the large cluster in Figure \ref{fig_skmean}(c) such as the rightmost one is further divided into 4 clusters.

\section{The complementary figures for all other fragments as discussed in Section \ref{sec_h5o2}}
\label{sec_com_fig}

\begin{figure}[H]
    \centering
    \subfigure[1 slices]{\includegraphics[width=0.48\columnwidth]{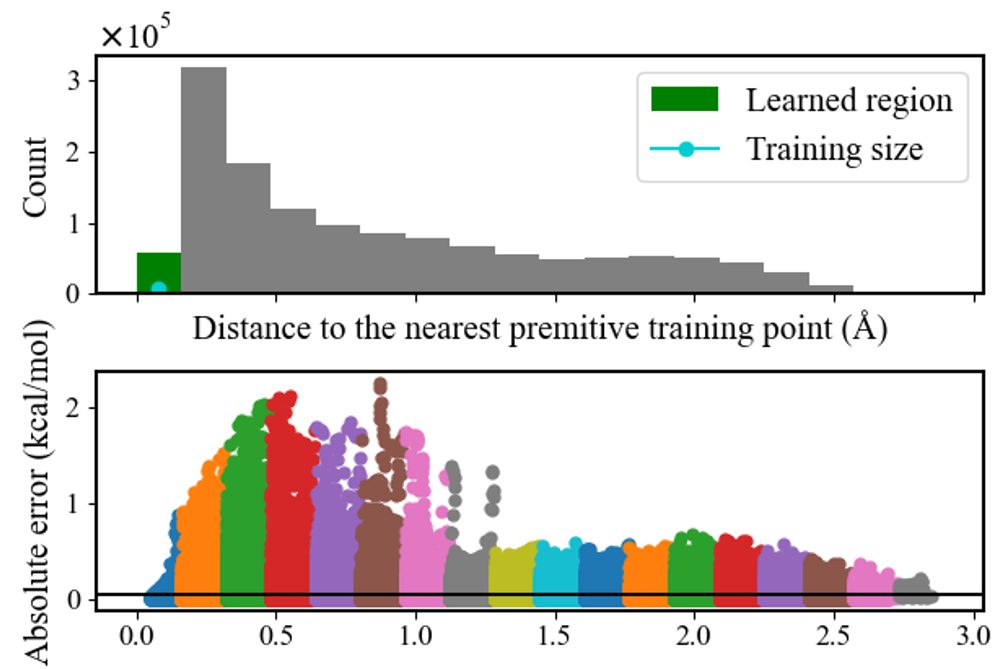}}
    \subfigure[5 slices]{\includegraphics[width=0.48\columnwidth]{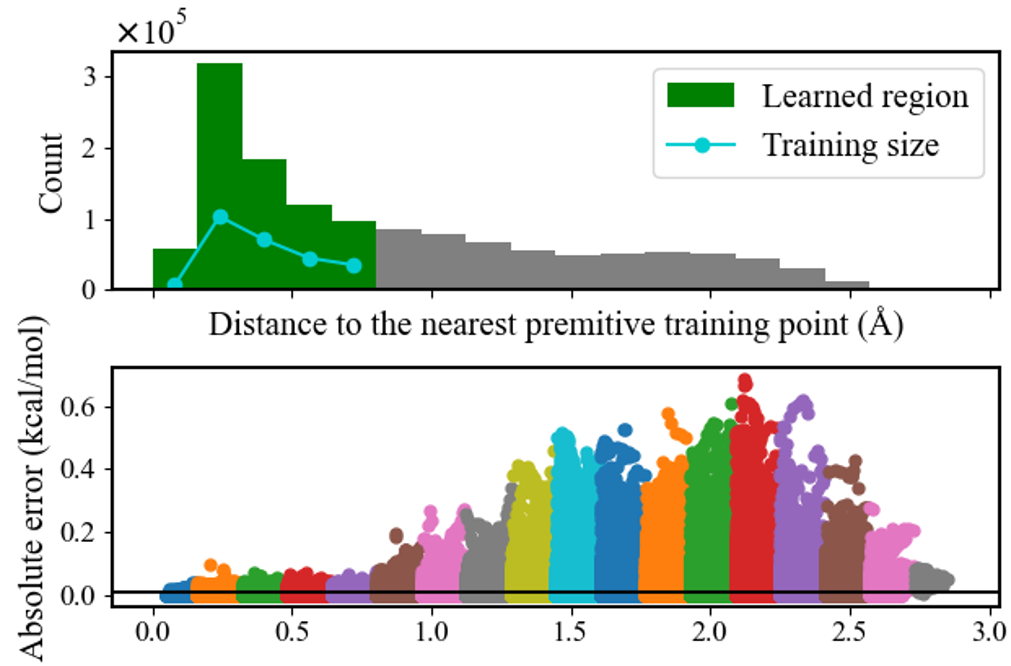}}
    \subfigure[10 slices]{\includegraphics[width=0.48\columnwidth]{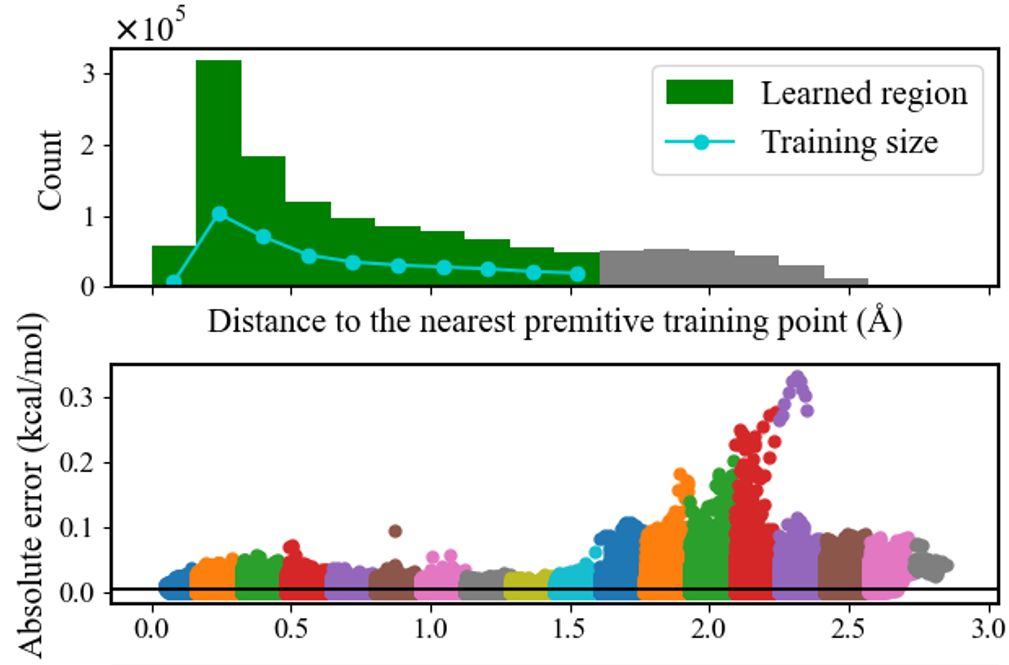}}    
    \subfigure[18 slices]{\includegraphics[width=0.48\columnwidth]{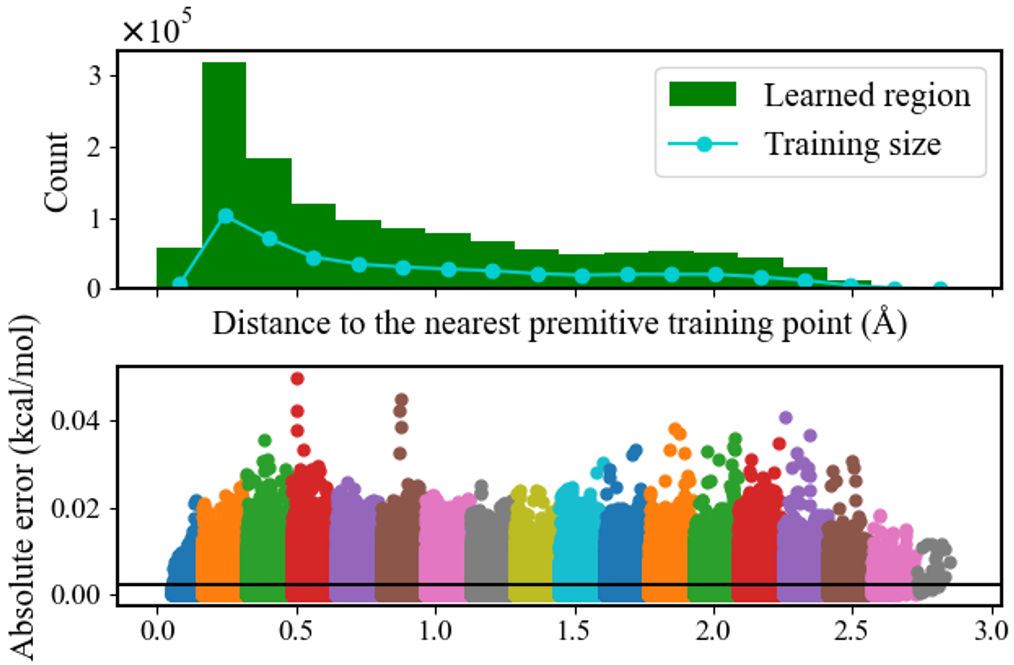}}
    \subfigure[summary]{\includegraphics[width=0.48\columnwidth]{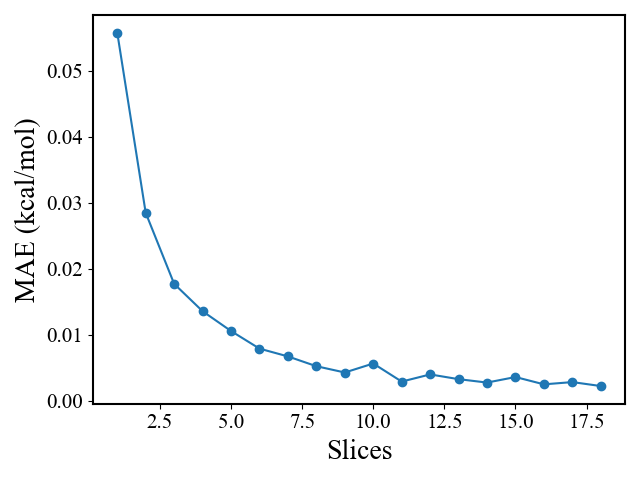}}
    \caption{\label{fig_6}Error distribution change with number of slices trained for $H_4O_2$.}
\end{figure}

\begin{figure}[H]
    \centering
    \subfigure[1 slices]{\includegraphics[width=0.48\columnwidth]{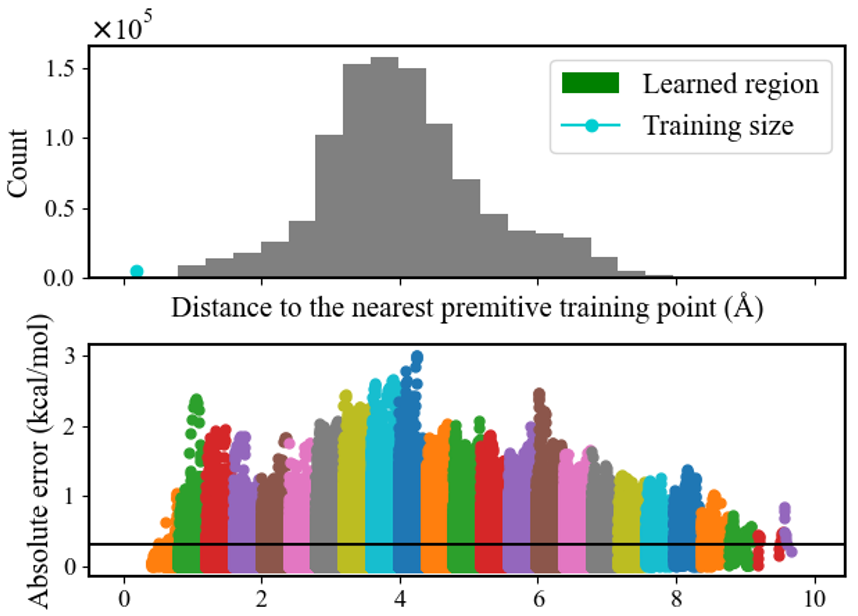}} 
    \subfigure[5 slices]{\includegraphics[width=0.48\columnwidth]{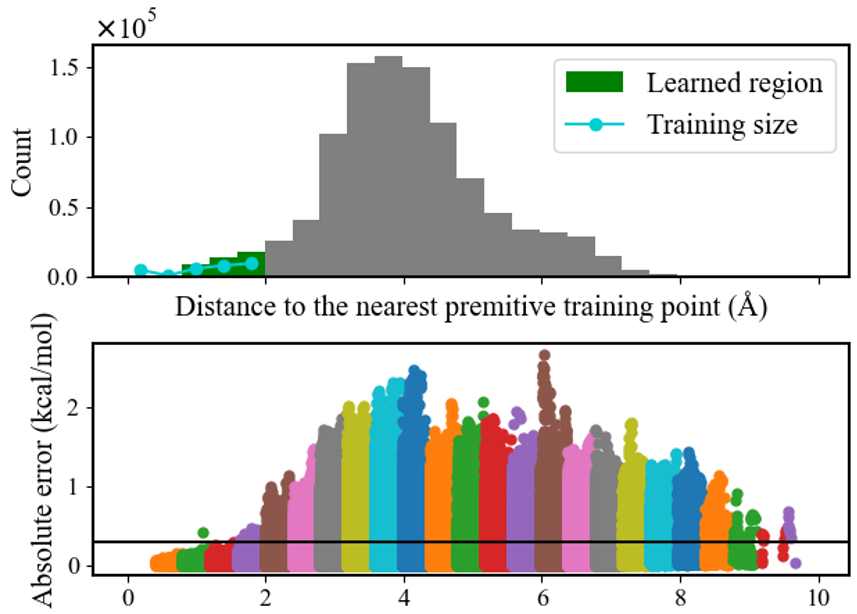}}
    \subfigure[10 slices]{\includegraphics[width=0.48\columnwidth]{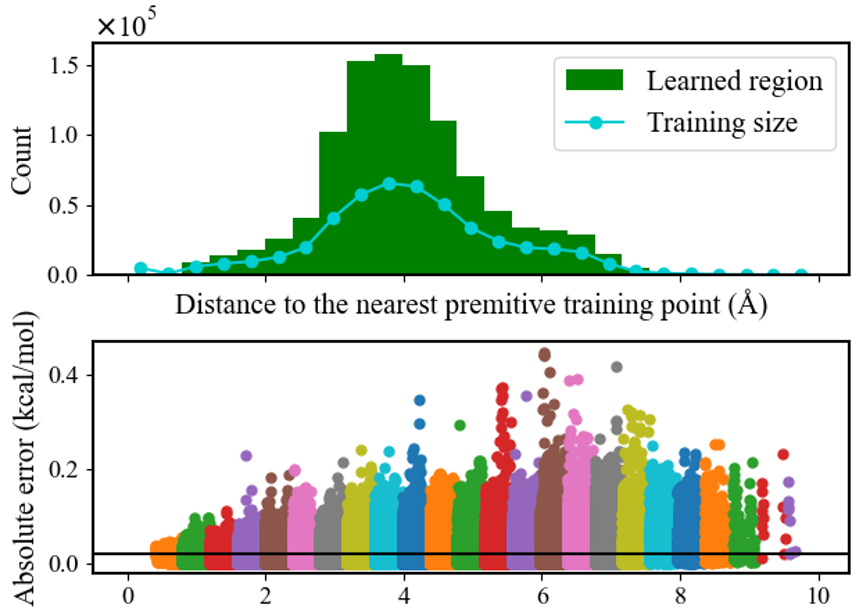}}    
    \subfigure[25 slices]{\includegraphics[width=0.48\columnwidth]{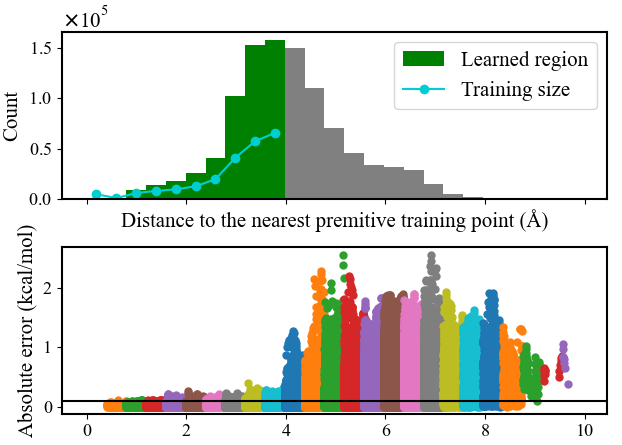}}
    \subfigure[summary]{\includegraphics[width=0.48\columnwidth]{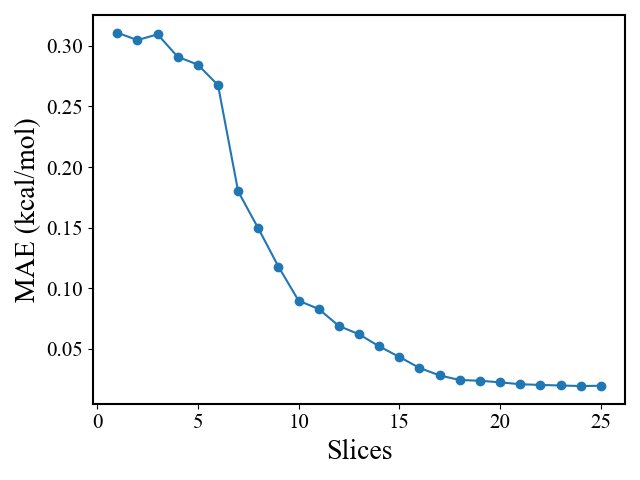}}
    \caption{\label{fig_9}Error distribution change with number of slices trained for $H_6O_3$.}
\end{figure}

\begin{figure}[H]
    \centering
    \subfigure[1 slices]{\includegraphics[width=0.48\columnwidth]{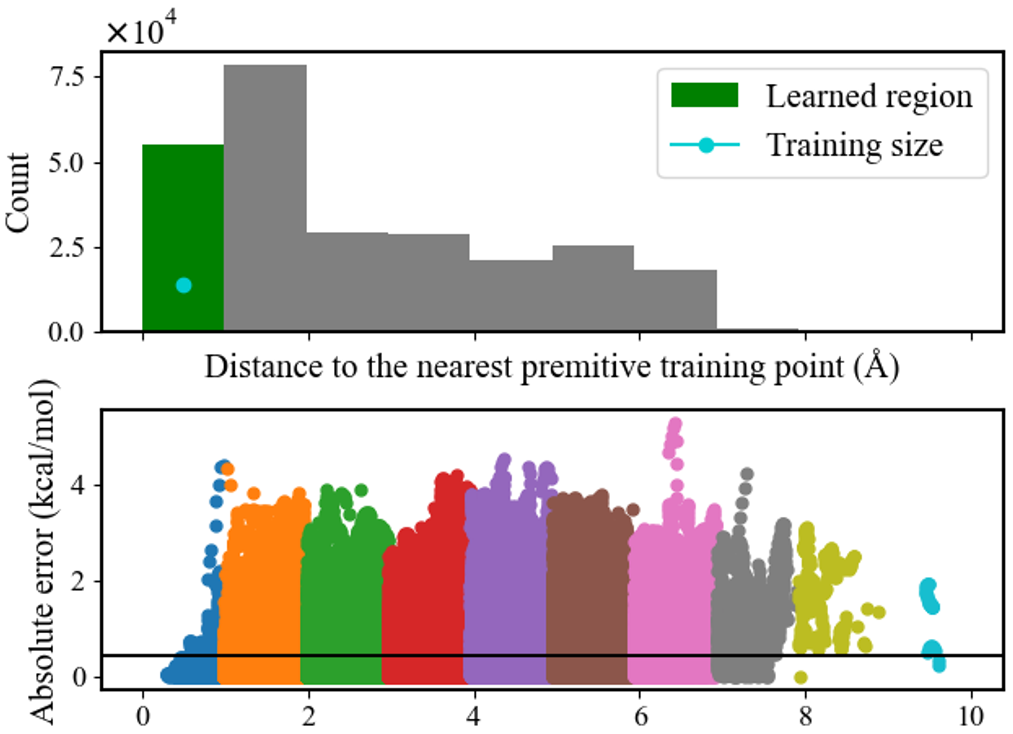}}
    \subfigure[5 slices]{\includegraphics[width=0.48\columnwidth]{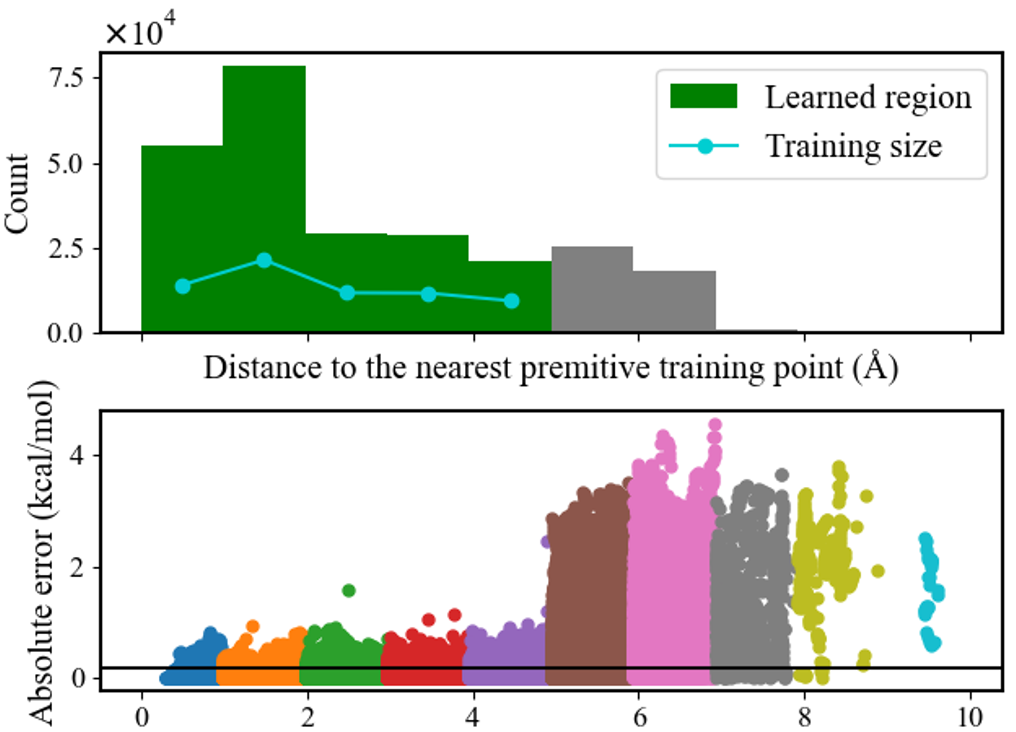}}    
    \subfigure[10 slices]{\includegraphics[width=0.48\columnwidth]{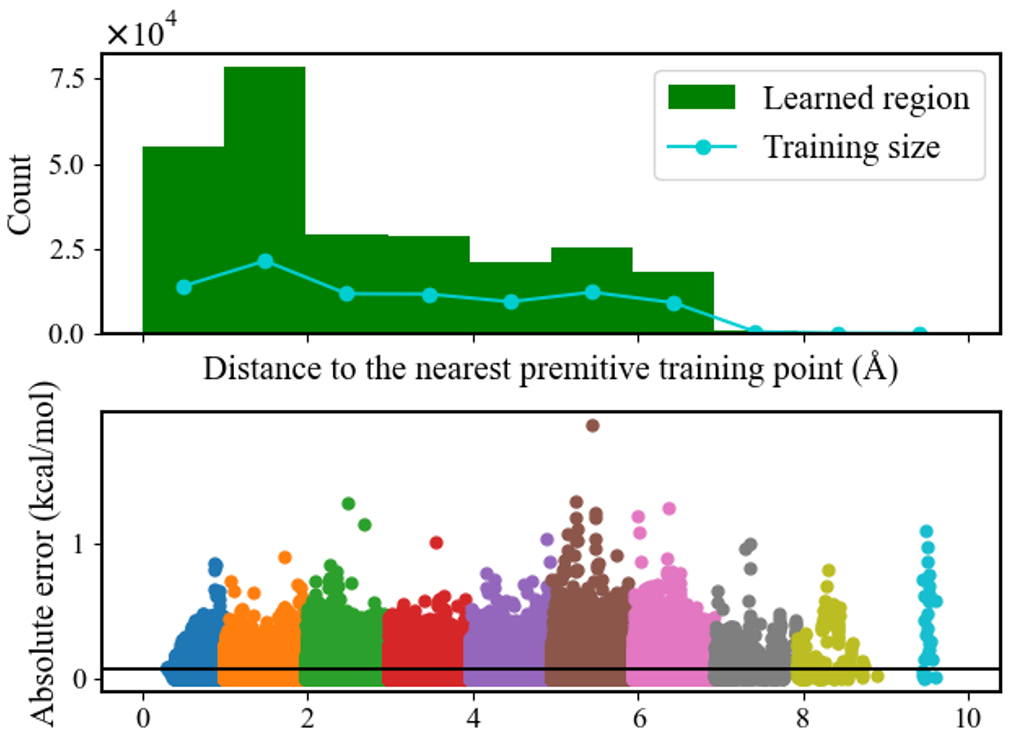}}    
    \subfigure[summary]{\includegraphics[width=0.48\columnwidth]{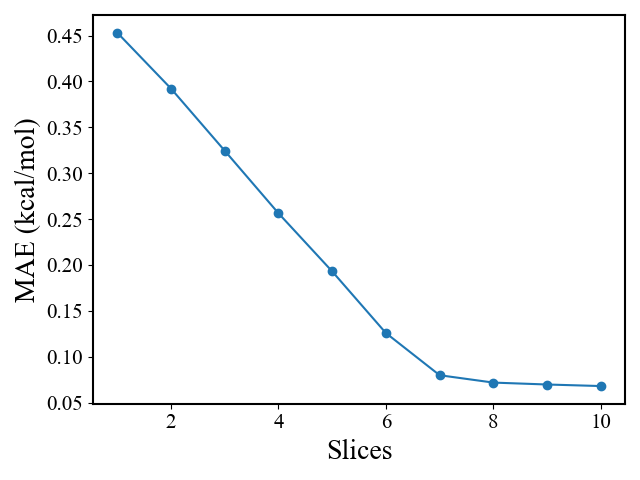}}
    \caption{\label{fig_10}Error distribution change with number of slices trained for $H_7O_3^+$.}
\end{figure}

\begin{figure}[H]
    \centering
    \subfigure[1 slices]{\includegraphics[width=0.48\columnwidth]{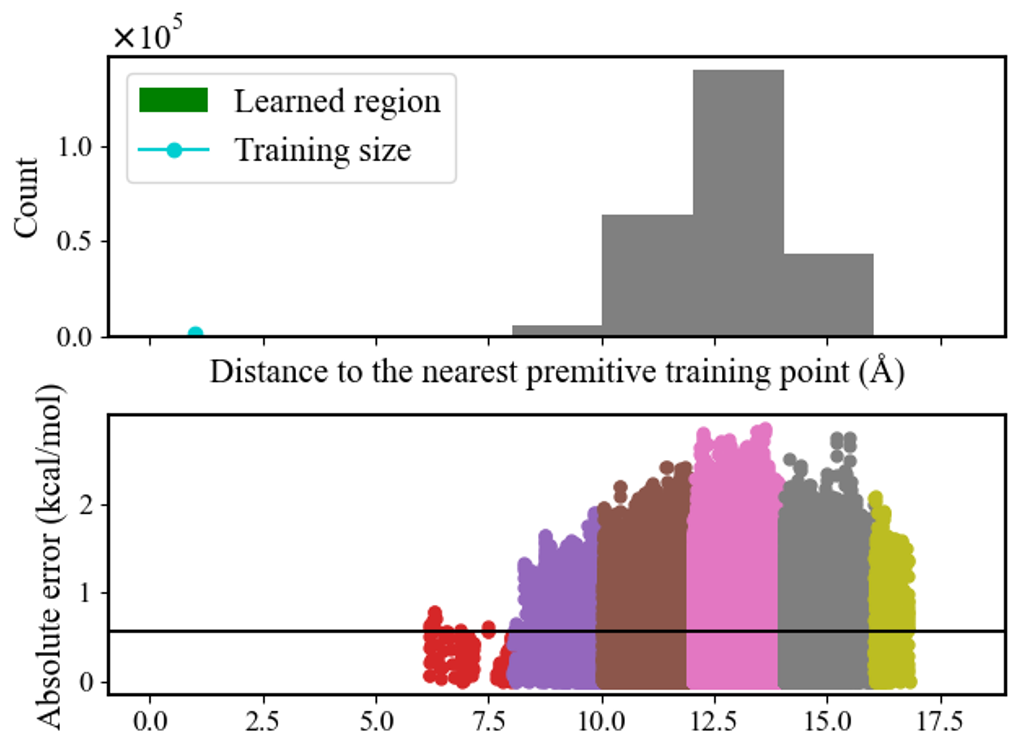}}
    \subfigure[5 slices]{\includegraphics[width=0.48\columnwidth]{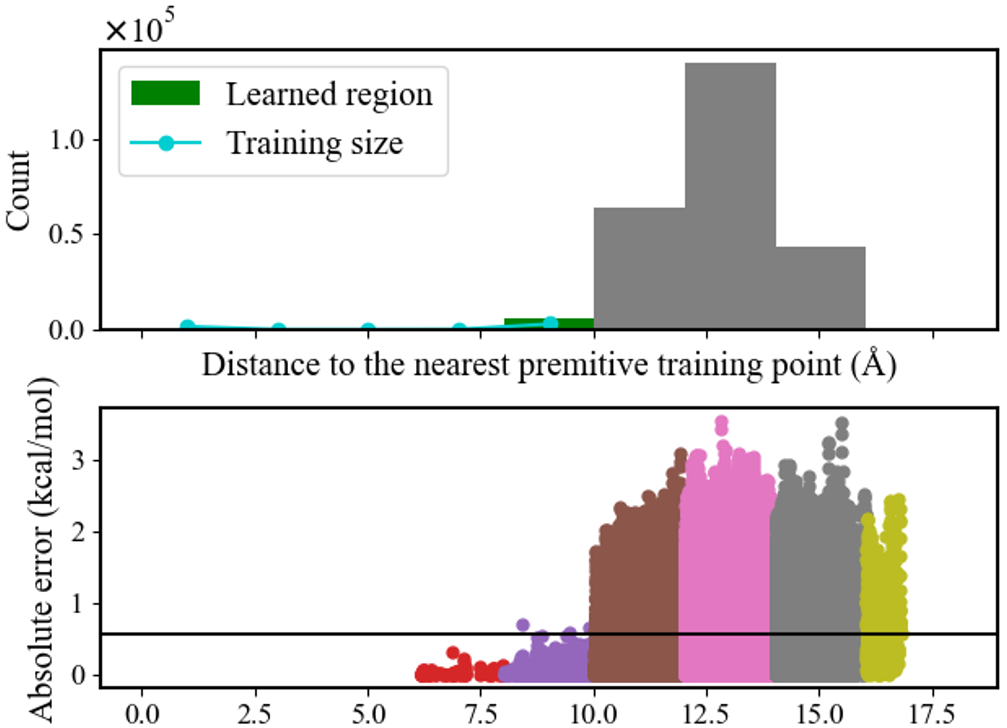}}       
    \subfigure[9 slices]{\includegraphics[width=0.48\columnwidth]{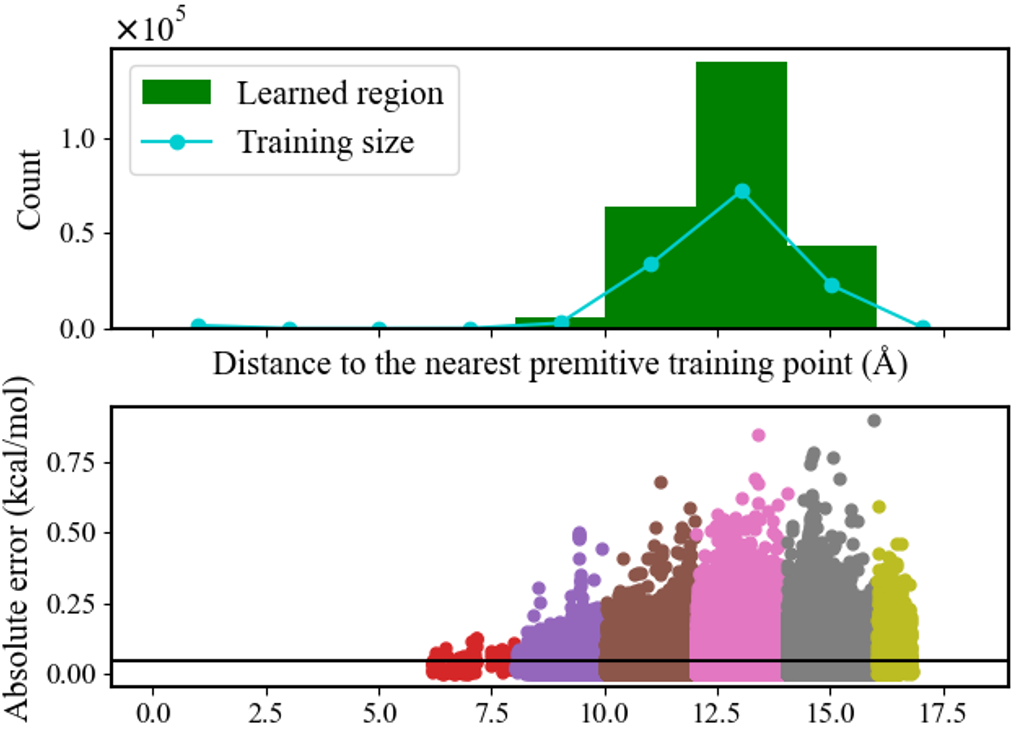}}
    \subfigure[summary]{\includegraphics[width=0.48\columnwidth]{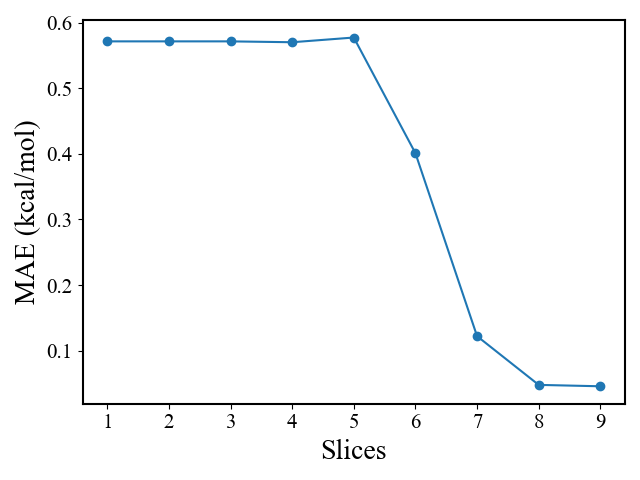}}
    \caption{\label{fig_12}Error distribution change with number of slices trained for $H_8O_4$.}
\end{figure}

\begin{figure}
    \centering
    \subfigure[1 slices]{\includegraphics[width=0.48\columnwidth]{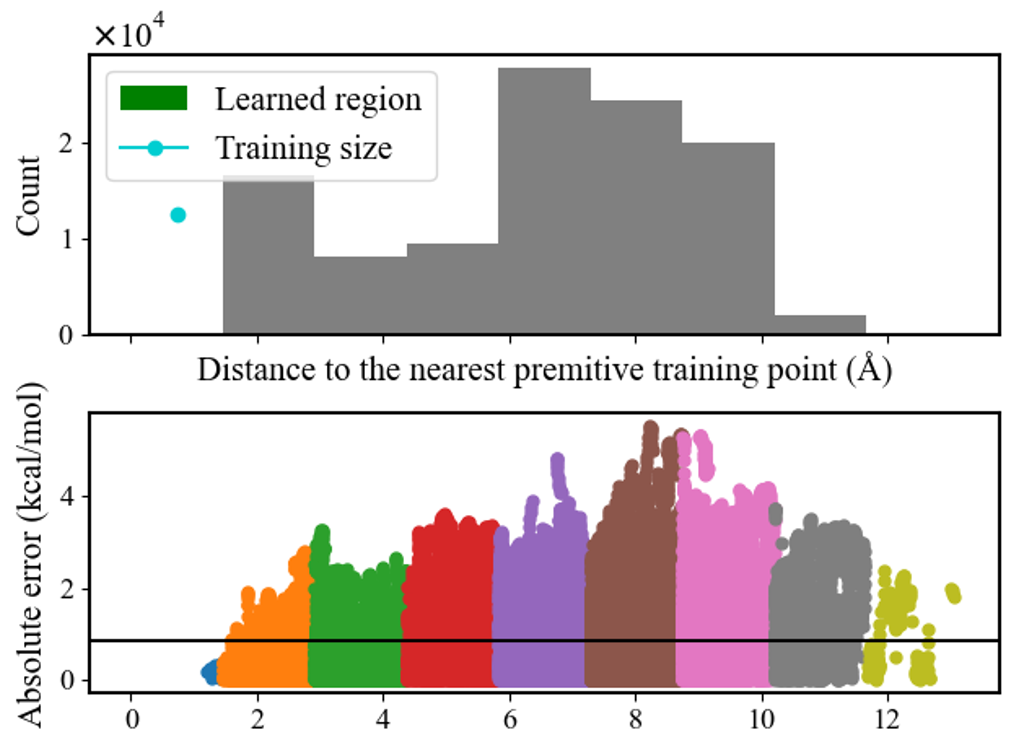}}
    \subfigure[5 slices]{\includegraphics[width=0.48\columnwidth]{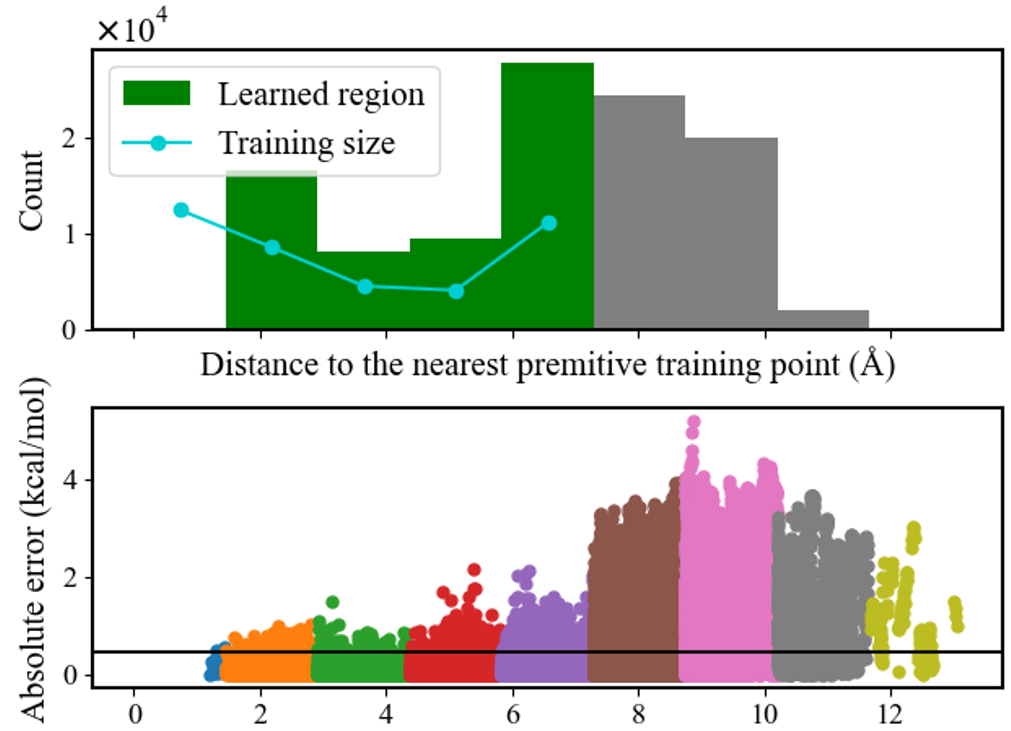}}      
    \subfigure[9 slices]{\includegraphics[width=0.48\columnwidth]{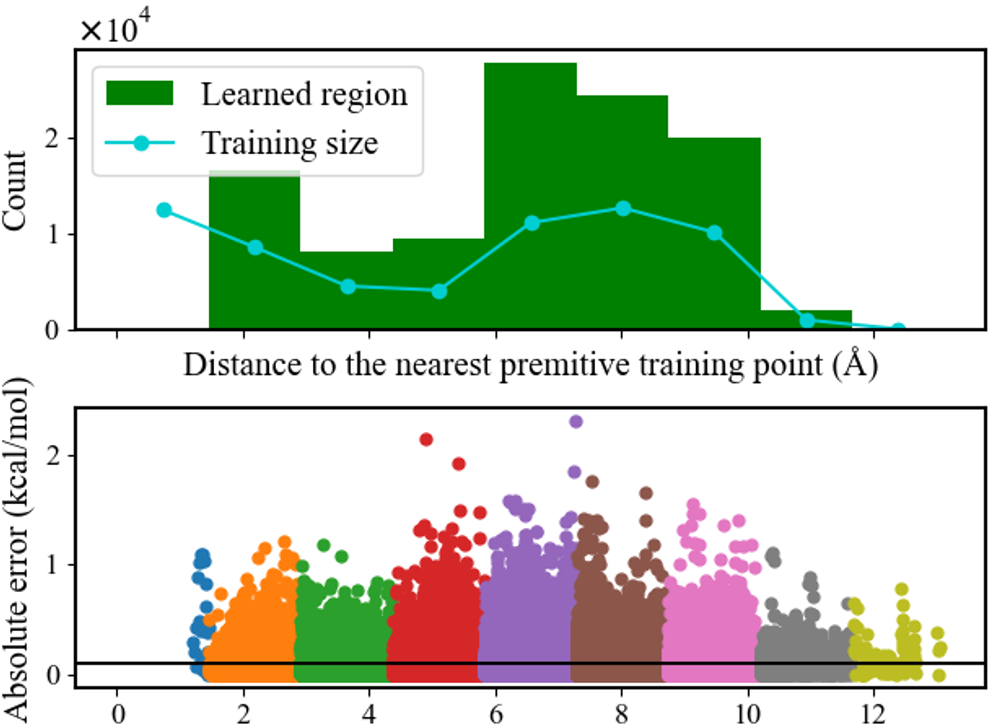}}
    \subfigure[summary]{\includegraphics[width=0.48\columnwidth]{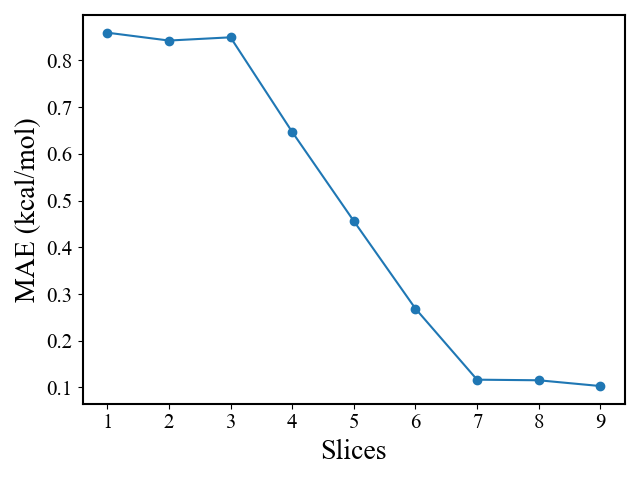}}
    \caption{\label{fig_13}Error distribution change with number of slices trained for $H_9O_4^+$.}
\end{figure}

In this section, we provide complementary figures for fragments species not covered in the main discussion of Section \ref{sec_h5o2}. Specifically, we provide the same set of figures as Figure \ref{fig_slice} for $H_4O_2$, $H_6O_3$, $H_7O_3^+$, $H_8O_4$, $H_9O_4^+$, shown in Figure \ref{fig_6}, \ref{fig_9}, \ref{fig_10}, \ref{fig_12}, \ref{fig_13} respectively. Overall, these fragments behave similar to that of $H_5O_2^+$ as discussed in Section \ref{sec_h5o2}. Throughout the  transferring process, neural networks capture the energy pattern within each additional slice and monotonically reduce prediction errors. As a result, the final predictions are accurate for fragments obtained from both solvated Zundel and protonated 21 water cluster trajectories.

\newpage

\begin{thebibliography}{120}%
\makeatletter
\providecommand \@ifxundefined [1]{%
 \@ifx{#1\undefined}
}%
\providecommand \@ifnum [1]{%
 \ifnum #1\expandafter \@firstoftwo
 \else \expandafter \@secondoftwo
 \fi
}%
\providecommand \@ifx [1]{%
 \ifx #1\expandafter \@firstoftwo
 \else \expandafter \@secondoftwo
 \fi
}%
\providecommand \natexlab [1]{#1}%
\providecommand \enquote  [1]{``#1''}%
\providecommand \bibnamefont  [1]{#1}%
\providecommand \bibfnamefont [1]{#1}%
\providecommand \citenamefont [1]{#1}%
\providecommand \href@noop [0]{\@secondoftwo}%
\providecommand \href [0]{\begingroup \@sanitize@url \@href}%
\providecommand \@href[1]{\@@startlink{#1}\@@href}%
\providecommand \@@href[1]{\endgroup#1\@@endlink}%
\providecommand \@sanitize@url [0]{\catcode `\\12\catcode `\$12\catcode `\&12\catcode `\#12\catcode `\^12\catcode `\_12\catcode `\%12\relax}%
\providecommand \@@startlink[1]{}%
\providecommand \@@endlink[0]{}%
\providecommand \url  [0]{\begingroup\@sanitize@url \@url }%
\providecommand \@url [1]{\endgroup\@href {#1}{\urlprefix }}%
\providecommand \urlprefix  [0]{URL }%
\providecommand \Eprint [0]{\href }%
\providecommand \doibase [0]{https://doi.org/}%
\providecommand \selectlanguage [0]{\@gobble}%
\providecommand \bibinfo  [0]{\@secondoftwo}%
\providecommand \bibfield  [0]{\@secondoftwo}%
\providecommand \translation [1]{[#1]}%
\providecommand \BibitemOpen [0]{}%
\providecommand \bibitemStop [0]{}%
\providecommand \bibitemNoStop [0]{.\EOS\space}%
\providecommand \EOS [0]{\spacefactor3000\relax}%
\providecommand \BibitemShut  [1]{\csname bibitem#1\endcsname}%
\let\auto@bib@innerbib\@empty
\bibitem [{\citenamefont {Murrell}\ \emph {et~al.}(1984)\citenamefont {Murrell}, \citenamefont {Carter}, \citenamefont {Farantos}, \citenamefont {Huxley},\ and\ \citenamefont {Varandas}}]{varandas-PES}%
  \BibitemOpen
  \bibfield  {author} {\bibinfo {author} {\bibfnamefont {J.}~\bibnamefont {Murrell}}, \bibinfo {author} {\bibfnamefont {S.}~\bibnamefont {Carter}}, \bibinfo {author} {\bibfnamefont {S.}~\bibnamefont {Farantos}}, \bibinfo {author} {\bibfnamefont {P.}~\bibnamefont {Huxley}},\ and\ \bibinfo {author} {\bibfnamefont {A.}~\bibnamefont {Varandas}},\ }\href@noop {} {\emph {\bibinfo {title} {Molecular Potential Energy Functions}}}\ (\bibinfo  {publisher} {Wiley, New York},\ \bibinfo {year} {1984})\BibitemShut {NoStop}%
\bibitem [{\citenamefont {Braams}\ and\ \citenamefont {Bowman}(2009)}]{Bowman-IRPC-PES}%
  \BibitemOpen
  \bibfield  {author} {\bibinfo {author} {\bibfnamefont {B.~J.}\ \bibnamefont {Braams}}\ and\ \bibinfo {author} {\bibfnamefont {J.~M.}\ \bibnamefont {Bowman}},\ }\bibfield  {title} {\bibinfo {title} {{Permutationally invariant potential energy surfaces in high dimensionality}},\ }\href@noop {} {\bibfield  {journal} {\bibinfo  {journal} {Int. Revs. Phys. Chem.}\ }\textbf {\bibinfo {volume} {28}},\ \bibinfo {pages} {577} (\bibinfo {year} {2009})}\BibitemShut {NoStop}%
\bibitem [{\citenamefont {Xie}\ and\ \citenamefont {Bowman}(2010)}]{Bowman-polybasis-fitting}%
  \BibitemOpen
  \bibfield  {author} {\bibinfo {author} {\bibfnamefont {Z.}~\bibnamefont {Xie}}\ and\ \bibinfo {author} {\bibfnamefont {J.~M.}\ \bibnamefont {Bowman}},\ }\bibfield  {title} {\bibinfo {title} {{Permutationally Invariant Polynomial Basis for Molecular Energy Surface Fitting Via Monomial Symmetrization}},\ }\href@noop {} {\bibfield  {journal} {\bibinfo  {journal} {J. Chem. Theory and Comput.}\ }\textbf {\bibinfo {volume} {6}},\ \bibinfo {pages} {26} (\bibinfo {year} {2010})}\BibitemShut {NoStop}%
\bibitem [{\citenamefont {Al{\i}{\c{s}}}\ and\ \citenamefont {Rabitz}(2001)}]{alics2001efficient}%
  \BibitemOpen
  \bibfield  {author} {\bibinfo {author} {\bibfnamefont {{\"O}.~F.}\ \bibnamefont {Al{\i}{\c{s}}}}\ and\ \bibinfo {author} {\bibfnamefont {H.}~\bibnamefont {Rabitz}},\ }\bibfield  {title} {\bibinfo {title} {Efficient implementation of high dimensional model representations},\ }\href@noop {} {\bibfield  {journal} {\bibinfo  {journal} {J. Math. Chem.}\ }\textbf {\bibinfo {volume} {29}},\ \bibinfo {pages} {127} (\bibinfo {year} {2001})}\BibitemShut {NoStop}%
\bibitem [{\citenamefont {J{\"a}ckle}\ and\ \citenamefont {Meyer}(1996)}]{jackle1996product}%
  \BibitemOpen
  \bibfield  {author} {\bibinfo {author} {\bibfnamefont {A.}~\bibnamefont {J{\"a}ckle}}\ and\ \bibinfo {author} {\bibfnamefont {H.-D.}\ \bibnamefont {Meyer}},\ }\bibfield  {title} {\bibinfo {title} {Product representation of potential energy surfaces},\ }\href@noop {} {\bibfield  {journal} {\bibinfo  {journal} {J. Chem. Phys.}\ }\textbf {\bibinfo {volume} {104}},\ \bibinfo {pages} {7974} (\bibinfo {year} {1996})}\BibitemShut {NoStop}%
\bibitem [{\citenamefont {Pel\'{a}ez}\ and\ \citenamefont {Meyer}(2013)}]{DieterPOTFIT}%
  \BibitemOpen
  \bibfield  {author} {\bibinfo {author} {\bibfnamefont {D.}~\bibnamefont {Pel\'{a}ez}}\ and\ \bibinfo {author} {\bibfnamefont {H.-D.}\ \bibnamefont {Meyer}},\ }\bibfield  {title} {\bibinfo {title} {The multigrid potfit (mgpf) method: Grid representations of potentials for quantum dynamics of large systems},\ }\href@noop {} {\bibfield  {journal} {\bibinfo  {journal} {J. Chem. Phys.}\ }\textbf {\bibinfo {volume} {138}},\ \bibinfo {pages} {014108} (\bibinfo {year} {2013})}\BibitemShut {NoStop}%
\bibitem [{\citenamefont {Leclerc}\ and\ \citenamefont {Carrington}(2014)}]{TuckerVibSpec2}%
  \BibitemOpen
  \bibfield  {author} {\bibinfo {author} {\bibfnamefont {A.}~\bibnamefont {Leclerc}}\ and\ \bibinfo {author} {\bibfnamefont {T.}~\bibnamefont {Carrington}},\ }\bibfield  {title} {\bibinfo {title} {Calculating vibrational spectra with sum of product basis functions without storing full-dimensional vectors or matrices},\ }\href@noop {} {\bibfield  {journal} {\bibinfo  {journal} {J. Chem. Phys.}\ }\textbf {\bibinfo {volume} {140}},\ \bibinfo {pages} {174111} (\bibinfo {year} {2014})}\BibitemShut {NoStop}%
\bibitem [{\citenamefont {Qu}\ \emph {et~al.}(2021)\citenamefont {Qu}, \citenamefont {Conte}, \citenamefont {Houston},\ and\ \citenamefont {Bowman}}]{full-dimPES-tunneling-Bowman2021}%
  \BibitemOpen
  \bibfield  {author} {\bibinfo {author} {\bibfnamefont {C.}~\bibnamefont {Qu}}, \bibinfo {author} {\bibfnamefont {R.}~\bibnamefont {Conte}}, \bibinfo {author} {\bibfnamefont {P.~L.}\ \bibnamefont {Houston}},\ and\ \bibinfo {author} {\bibfnamefont {J.~M.}\ \bibnamefont {Bowman}},\ }\bibfield  {title} {\bibinfo {title} {Full-dimensional potential energy surface for acetylacetone and tunneling splittings},\ }\href@noop {} {\bibfield  {journal} {\bibinfo  {journal} {Phys. Chem. Chem. Phys.}\ }\textbf {\bibinfo {volume} {23}},\ \bibinfo {pages} {7758} (\bibinfo {year} {2021})}\BibitemShut {NoStop}%
\bibitem [{\citenamefont {Grossmann}\ \emph {et~al.}(2023)\citenamefont {Grossmann}, \citenamefont {Eilermann}, \citenamefont {Rensmeyer}, \citenamefont {Liebert}, \citenamefont {Hohmann}, \citenamefont {Wittke},\ and\ \citenamefont {Niggemann}}]{grossmann2023positionpapermaterialsdesign}%
  \BibitemOpen
  \bibfield  {author} {\bibinfo {author} {\bibfnamefont {W.}~\bibnamefont {Grossmann}}, \bibinfo {author} {\bibfnamefont {S.}~\bibnamefont {Eilermann}}, \bibinfo {author} {\bibfnamefont {T.}~\bibnamefont {Rensmeyer}}, \bibinfo {author} {\bibfnamefont {A.}~\bibnamefont {Liebert}}, \bibinfo {author} {\bibfnamefont {M.}~\bibnamefont {Hohmann}}, \bibinfo {author} {\bibfnamefont {C.}~\bibnamefont {Wittke}},\ and\ \bibinfo {author} {\bibfnamefont {O.}~\bibnamefont {Niggemann}},\ }\href@noop {} {\bibinfo {title} {Position paper on materials design -- a modern approach}} (\bibinfo {year} {2023})\BibitemShut {NoStop}%
\bibitem [{\citenamefont {Vasudevan}\ \emph {et~al.}(2021)\citenamefont {Vasudevan}, \citenamefont {Pilania},\ and\ \citenamefont {Balachandran}}]{Mat-Design-ML-JAP}%
  \BibitemOpen
  \bibfield  {author} {\bibinfo {author} {\bibfnamefont {R.}~\bibnamefont {Vasudevan}}, \bibinfo {author} {\bibfnamefont {G.}~\bibnamefont {Pilania}},\ and\ \bibinfo {author} {\bibfnamefont {P.~V.}\ \bibnamefont {Balachandran}},\ }\bibfield  {title} {\bibinfo {title} {Machine learning for materials design and discovery},\ }\href@noop {} {\bibfield  {journal} {\bibinfo  {journal} {Journal of Applied Physics}\ }\textbf {\bibinfo {volume} {129}},\ \bibinfo {pages} {070401} (\bibinfo {year} {2021})}\BibitemShut {NoStop}%
\bibitem [{\citenamefont {Axelrod}\ \emph {et~al.}(2022)\citenamefont {Axelrod}, \citenamefont {Schwalbe-Koda}, \citenamefont {Mohapatra}, \citenamefont {Damewood}, \citenamefont {Greenman},\ and\ \citenamefont {Gómez-Bombarelli}}]{Mat-Design-ML-AMR}%
  \BibitemOpen
  \bibfield  {author} {\bibinfo {author} {\bibfnamefont {S.}~\bibnamefont {Axelrod}}, \bibinfo {author} {\bibfnamefont {D.}~\bibnamefont {Schwalbe-Koda}}, \bibinfo {author} {\bibfnamefont {S.}~\bibnamefont {Mohapatra}}, \bibinfo {author} {\bibfnamefont {J.}~\bibnamefont {Damewood}}, \bibinfo {author} {\bibfnamefont {K.~P.}\ \bibnamefont {Greenman}},\ and\ \bibinfo {author} {\bibfnamefont {R.}~\bibnamefont {Gómez-Bombarelli}},\ }\bibfield  {title} {\bibinfo {title} {Learning matter: Materials design with machine learning and atomistic simulations},\ }\href@noop {} {\bibfield  {journal} {\bibinfo  {journal} {Accounts of Materials Research}\ }\textbf {\bibinfo {volume} {3}},\ \bibinfo {pages} {343} (\bibinfo {year} {2022})}\BibitemShut {NoStop}%
\bibitem [{\citenamefont {Persson}\ \emph {et~al.}(2010)\citenamefont {Persson}, \citenamefont {Hinuma}, \citenamefont {Meng}, \citenamefont {Van~der Ven},\ and\ \citenamefont {Ceder}}]{Ceder-PhysRevB.82.125416}%
  \BibitemOpen
  \bibfield  {author} {\bibinfo {author} {\bibfnamefont {K.}~\bibnamefont {Persson}}, \bibinfo {author} {\bibfnamefont {Y.}~\bibnamefont {Hinuma}}, \bibinfo {author} {\bibfnamefont {Y.~S.}\ \bibnamefont {Meng}}, \bibinfo {author} {\bibfnamefont {A.}~\bibnamefont {Van~der Ven}},\ and\ \bibinfo {author} {\bibfnamefont {G.}~\bibnamefont {Ceder}},\ }\bibfield  {title} {\bibinfo {title} {Thermodynamic and kinetic properties of the li-graphite system from first-principles calculations},\ }\href@noop {} {\bibfield  {journal} {\bibinfo  {journal} {Phys. Rev. B}\ }\textbf {\bibinfo {volume} {82}},\ \bibinfo {pages} {125416} (\bibinfo {year} {2010})}\BibitemShut {NoStop}%
\bibitem [{\citenamefont {Flam-Shepherd}\ \emph {et~al.}(2022)\citenamefont {Flam-Shepherd}, \citenamefont {Zhigalin},\ and\ \citenamefont {Aspuru-Guzik}}]{Aspuru-design}%
  \BibitemOpen
  \bibfield  {author} {\bibinfo {author} {\bibfnamefont {D.}~\bibnamefont {Flam-Shepherd}}, \bibinfo {author} {\bibfnamefont {A.}~\bibnamefont {Zhigalin}},\ and\ \bibinfo {author} {\bibfnamefont {A.}~\bibnamefont {Aspuru-Guzik}},\ }\href@noop {} {\bibinfo {title} {Scalable fragment-based 3d molecular design with reinforcement learning}} (\bibinfo {year} {2022})\BibitemShut {NoStop}%
\bibitem [{\citenamefont {Feynman}(1982)}]{Feynman1982}%
  \BibitemOpen
  \bibfield  {author} {\bibinfo {author} {\bibfnamefont {R.~P.}\ \bibnamefont {Feynman}},\ }\bibfield  {title} {\bibinfo {title} {Simulating physics with computers},\ }\bibfield  {booktitle} {\emph {\bibinfo {booktitle} {International Journal of Theoretical Physics}},\ }\href@noop {} {\bibfield  {journal} {\bibinfo  {journal} {International Journal of Theoretical Physics}\ }\textbf {\bibinfo {volume} {21}},\ \bibinfo {pages} {467} (\bibinfo {year} {1982})}\BibitemShut {NoStop}%
\bibitem [{\citenamefont {Feynman}\ \emph {et~al.}(1998)\citenamefont {Feynman}, \citenamefont {Hey},\ and\ \citenamefont {Allen}}]{Feynman-Comp}%
  \BibitemOpen
  \bibfield  {author} {\bibinfo {author} {\bibfnamefont {R.~P.}\ \bibnamefont {Feynman}}, \bibinfo {author} {\bibfnamefont {J.}~\bibnamefont {Hey}},\ and\ \bibinfo {author} {\bibfnamefont {R.~W.}\ \bibnamefont {Allen}},\ }\href@noop {} {\emph {\bibinfo {title} {Feynman Lectures on Computation}}}\ (\bibinfo  {publisher} {Addison-Wesley Longman Publishing Co., Inc.},\ \bibinfo {year} {1998})\BibitemShut {NoStop}%
\bibitem [{\citenamefont {Feynman}\ and\ \citenamefont {Hibbs}(1965)}]{hibbs}%
  \BibitemOpen
  \bibfield  {author} {\bibinfo {author} {\bibfnamefont {R.~P.}\ \bibnamefont {Feynman}}\ and\ \bibinfo {author} {\bibfnamefont {A.~R.}\ \bibnamefont {Hibbs}},\ }\href@noop {} {\emph {\bibinfo {title} {Quantum Mechanics and Path Integrals}}}\ (\bibinfo  {publisher} {McGraw-Hill Book Company},\ \bibinfo {address} {New York},\ \bibinfo {year} {1965})\BibitemShut {NoStop}%
\bibitem [{\citenamefont {Meyer}\ \emph {et~al.}(1990)\citenamefont {Meyer}, \citenamefont {Manthe},\ and\ \citenamefont {Cederbaum}}]{MCTDH-Meyer1}%
  \BibitemOpen
  \bibfield  {author} {\bibinfo {author} {\bibfnamefont {H.-D.}\ \bibnamefont {Meyer}}, \bibinfo {author} {\bibfnamefont {U.}~\bibnamefont {Manthe}},\ and\ \bibinfo {author} {\bibfnamefont {L.~S.}\ \bibnamefont {Cederbaum}},\ }\bibfield  {title} {\bibinfo {title} {The multi-configurational time-dependent hartree approach},\ }\href@noop {} {\bibfield  {journal} {\bibinfo  {journal} {Chem. Phys. Lett.}\ }\textbf {\bibinfo {volume} {165}},\ \bibinfo {pages} {73} (\bibinfo {year} {1990})}\BibitemShut {NoStop}%
\bibitem [{\citenamefont {Nielsen}\ and\ \citenamefont {Chuang}(2000)}]{Nielsen-Chuang-QuantComp}%
  \BibitemOpen
  \bibfield  {author} {\bibinfo {author} {\bibfnamefont {M.~A.}\ \bibnamefont {Nielsen}}\ and\ \bibinfo {author} {\bibfnamefont {I.~L.}\ \bibnamefont {Chuang}},\ }\href@noop {} {\emph {\bibinfo {title} {Quantum computation and quantum information}}}\ (\bibinfo  {publisher} {Cambridge University Press, Cambridge},\ \bibinfo {year} {2000})\BibitemShut {NoStop}%
\bibitem [{\citenamefont {Rabitz}\ and\ \citenamefont {Al{\i}{\c{s}}}(1999)}]{Rabitz-HDMR}%
  \BibitemOpen
  \bibfield  {author} {\bibinfo {author} {\bibfnamefont {H.}~\bibnamefont {Rabitz}}\ and\ \bibinfo {author} {\bibfnamefont {{\"O}.~F.}\ \bibnamefont {Al{\i}{\c{s}}}},\ }\bibfield  {title} {\bibinfo {title} {General foundations of high dimensional model representations},\ }\href@noop {} {\bibfield  {journal} {\bibinfo  {journal} {J. Math. Chem.}\ }\textbf {\bibinfo {volume} {25}},\ \bibinfo {pages} {197} (\bibinfo {year} {1999})}\BibitemShut {NoStop}%
\bibitem [{\citenamefont {Manzhos}\ and\ \citenamefont {Carrington~Jr}(2008)}]{manzhos2008using}%
  \BibitemOpen
  \bibfield  {author} {\bibinfo {author} {\bibfnamefont {S.}~\bibnamefont {Manzhos}}\ and\ \bibinfo {author} {\bibfnamefont {T.}~\bibnamefont {Carrington~Jr}},\ }\bibfield  {title} {\bibinfo {title} {Using neural networks, optimized coordinates, and high-dimensional model representations to obtain a vinyl bromide potential surface},\ }\href@noop {} {\bibfield  {journal} {\bibinfo  {journal} {J. Chem. Phys.}\ }\textbf {\bibinfo {volume} {129}},\ \bibinfo {pages} {224104} (\bibinfo {year} {2008})}\BibitemShut {NoStop}%
\bibitem [{\citenamefont {Otto}(2014)}]{Otto_POTFIT}%
  \BibitemOpen
  \bibfield  {author} {\bibinfo {author} {\bibfnamefont {F.}~\bibnamefont {Otto}},\ }\bibfield  {title} {\bibinfo {title} {Multi-layer potfit: An accurate potential representation for efficient high-dimensional quantum dynamics},\ }\href@noop {} {\bibfield  {journal} {\bibinfo  {journal} {J. Chem. Phys.}\ }\textbf {\bibinfo {volume} {140}},\ \bibinfo {pages} {014106} (\bibinfo {year} {2014})}\BibitemShut {NoStop}%
\bibitem [{\citenamefont {Sumner}\ and\ \citenamefont {Iyengar}(2007)}]{qwaimd-wavelet}%
  \BibitemOpen
  \bibfield  {author} {\bibinfo {author} {\bibfnamefont {I.}~\bibnamefont {Sumner}}\ and\ \bibinfo {author} {\bibfnamefont {S.~S.}\ \bibnamefont {Iyengar}},\ }\bibfield  {title} {\bibinfo {title} {Quantum wavepacket {\em ab initio} molecular dynamics: An approach for computing dynamically averaged vibrational spectra including critical nuclear quantum effects},\ }\href@noop {} {\bibfield  {journal} {\bibinfo  {journal} {J. Phys. Chem. A}\ }\textbf {\bibinfo {volume} {111}},\ \bibinfo {pages} {10313} (\bibinfo {year} {2007})}\BibitemShut {NoStop}%
\bibitem [{\citenamefont {DeGregorio}\ and\ \citenamefont {Iyengar}(2018)}]{nicole-Shannon}%
  \BibitemOpen
  \bibfield  {author} {\bibinfo {author} {\bibfnamefont {N.}~\bibnamefont {DeGregorio}}\ and\ \bibinfo {author} {\bibfnamefont {S.~S.}\ \bibnamefont {Iyengar}},\ }\bibfield  {title} {\bibinfo {title} {Efficient and adaptive methods for computing accurate potential surfaces for quantum nuclear effects: Applications to hydrogen-\ transfer reactions},\ }\href@noop {} {\bibfield  {journal} {\bibinfo  {journal} {J. Chem. Theory Comput.}\ }\textbf {\bibinfo {volume} {14}},\ \bibinfo {pages} {30} (\bibinfo {year} {2018})}\BibitemShut {NoStop}%
\bibitem [{\citenamefont {Schlegel}\ and\ \citenamefont {Frisch}(1991{\natexlab{a}})}]{schlegel1991computational}%
  \BibitemOpen
  \bibfield  {author} {\bibinfo {author} {\bibfnamefont {H.~B.}\ \bibnamefont {Schlegel}}\ and\ \bibinfo {author} {\bibfnamefont {M.~J.}\ \bibnamefont {Frisch}},\ }\bibfield  {title} {\bibinfo {title} {{Computational bottlenecks in molecular orbital calculations}},\ }in\ \href@noop {} {\emph {\bibinfo {booktitle} {Theoretical and Computational Models for Organic Chemistry}}}\ (\bibinfo  {publisher} {Springer},\ \bibinfo {year} {1991})\ pp.\ \bibinfo {pages} {5--33}\BibitemShut {NoStop}%
\bibitem [{\citenamefont {Schran}\ \emph {et~al.}(2021)\citenamefont {Schran}, \citenamefont {Brieuc},\ and\ \citenamefont {Marx}}]{waters_transfer_nnp}%
  \BibitemOpen
  \bibfield  {author} {\bibinfo {author} {\bibfnamefont {C.}~\bibnamefont {Schran}}, \bibinfo {author} {\bibfnamefont {F.}~\bibnamefont {Brieuc}},\ and\ \bibinfo {author} {\bibfnamefont {D.}~\bibnamefont {Marx}},\ }\bibfield  {title} {\bibinfo {title} {Transferability of machine learning potentials: Protonated water neural network potential applied to the protonated water hexamer},\ }\href@noop {} {\bibfield  {journal} {\bibinfo  {journal} {J. Chem. Phys.}\ }\textbf {\bibinfo {volume} {154}} (\bibinfo {year} {2021})}\BibitemShut {NoStop}%
\bibitem [{\citenamefont {Nguyen}\ \emph {et~al.}(2018)\citenamefont {Nguyen}, \citenamefont {Szekely}, \citenamefont {Imbalzano}, \citenamefont {Behler}, \citenamefont {Csanyi}, \citenamefont {Ceriotti}, \citenamefont {Goetz},\ and\ \citenamefont {Paesani}}]{ml_compare_paesani}%
  \BibitemOpen
  \bibfield  {author} {\bibinfo {author} {\bibfnamefont {T.~T.}\ \bibnamefont {Nguyen}}, \bibinfo {author} {\bibfnamefont {E.}~\bibnamefont {Szekely}}, \bibinfo {author} {\bibfnamefont {G.}~\bibnamefont {Imbalzano}}, \bibinfo {author} {\bibfnamefont {J.}~\bibnamefont {Behler}}, \bibinfo {author} {\bibfnamefont {G.}~\bibnamefont {Csanyi}}, \bibinfo {author} {\bibfnamefont {M.}~\bibnamefont {Ceriotti}}, \bibinfo {author} {\bibfnamefont {A.~W.}\ \bibnamefont {Goetz}},\ and\ \bibinfo {author} {\bibfnamefont {F.}~\bibnamefont {Paesani}},\ }\bibfield  {title} {\bibinfo {title} {Comparison of permutationally invariant polynomials, neural networks, and gaussian approximation potentials in representing water interactions through many-body expansions},\ }\href@noop {} {\bibfield  {journal} {\bibinfo  {journal} {J. Chem. Phys.}\ }\textbf {\bibinfo {volume} {148}} (\bibinfo {year} {2018})}\BibitemShut {NoStop}%
\bibitem [{\citenamefont {Schran}\ \emph {et~al.}(2020)\citenamefont {Schran}, \citenamefont {Behler},\ and\ \citenamefont {Marx}}]{waters_nnp}%
  \BibitemOpen
  \bibfield  {author} {\bibinfo {author} {\bibfnamefont {C.}~\bibnamefont {Schran}}, \bibinfo {author} {\bibfnamefont {J.}~\bibnamefont {Behler}},\ and\ \bibinfo {author} {\bibfnamefont {D.}~\bibnamefont {Marx}},\ }\bibfield  {title} {\bibinfo {title} {Automated fitting of neural network potentials at coupled cluster accuracy: Protonated water clusters as testing ground},\ }\href@noop {} {\bibfield  {journal} {\bibinfo  {journal} {J. Chem. Theory Comput.}\ }\textbf {\bibinfo {volume} {16}},\ \bibinfo {pages} {88} (\bibinfo {year} {2020})}\BibitemShut {NoStop}%
\bibitem [{\citenamefont {Lu}\ \emph {et~al.}(2020)\citenamefont {Lu}, \citenamefont {Behler},\ and\ \citenamefont {Li}}]{H+CH3OH_NNP}%
  \BibitemOpen
  \bibfield  {author} {\bibinfo {author} {\bibfnamefont {D.}~\bibnamefont {Lu}}, \bibinfo {author} {\bibfnamefont {J.}~\bibnamefont {Behler}},\ and\ \bibinfo {author} {\bibfnamefont {J.}~\bibnamefont {Li}},\ }\bibfield  {title} {\bibinfo {title} {Accurate global potential energy surfaces for the {H + CH$_3$OH} reaction by neural network fitting with permutation invariance},\ }\href@noop {} {\bibfield  {journal} {\bibinfo  {journal} {J. Phys. Chem. A}\ }\textbf {\bibinfo {volume} {124}},\ \bibinfo {pages} {5737} (\bibinfo {year} {2020})}\BibitemShut {NoStop}%
\bibitem [{\citenamefont {Lu}\ \emph {et~al.}(2016)\citenamefont {Lu}, \citenamefont {Qi}, \citenamefont {Yang}, \citenamefont {Behler}, \citenamefont {Song},\ and\ \citenamefont {Li}}]{H2+SH_NNP}%
  \BibitemOpen
  \bibfield  {author} {\bibinfo {author} {\bibfnamefont {D.}~\bibnamefont {Lu}}, \bibinfo {author} {\bibfnamefont {J.}~\bibnamefont {Qi}}, \bibinfo {author} {\bibfnamefont {M.}~\bibnamefont {Yang}}, \bibinfo {author} {\bibfnamefont {J.}~\bibnamefont {Behler}}, \bibinfo {author} {\bibfnamefont {H.}~\bibnamefont {Song}},\ and\ \bibinfo {author} {\bibfnamefont {J.}~\bibnamefont {Li}},\ }\bibfield  {title} {\bibinfo {title} {Mode specific dynamics in the {H$_2$ + SH $\xrightarrow{}$ H + H$_2$S} reaction},\ }\href@noop {} {\bibfield  {journal} {\bibinfo  {journal} {Phys. Chem. Chem. Phys.}\ }\textbf {\bibinfo {volume} {18}},\ \bibinfo {pages} {29113} (\bibinfo {year} {2016})}\BibitemShut {NoStop}%
\bibitem [{\citenamefont {Li}\ \emph {et~al.}(2019)\citenamefont {Li}, \citenamefont {Song},\ and\ \citenamefont {Behler}}]{critical_compare_nnp}%
  \BibitemOpen
  \bibfield  {author} {\bibinfo {author} {\bibfnamefont {J.}~\bibnamefont {Li}}, \bibinfo {author} {\bibfnamefont {K.}~\bibnamefont {Song}},\ and\ \bibinfo {author} {\bibfnamefont {J.}~\bibnamefont {Behler}},\ }\bibfield  {title} {\bibinfo {title} {A critical comparison of neural network potentials for molecular reaction dynamics with exact permutation symmetry},\ }\href@noop {} {\bibfield  {journal} {\bibinfo  {journal} {Phys. Chem. Chem. Phys.}\ }\textbf {\bibinfo {volume} {21}},\ \bibinfo {pages} {9672} (\bibinfo {year} {2019})}\BibitemShut {NoStop}%
\bibitem [{\citenamefont {Smith}\ \emph {et~al.}(2019)\citenamefont {Smith}, \citenamefont {Nebgen}, \citenamefont {Zubatyuk}, \citenamefont {Lubbers}, \citenamefont {Devereux}, \citenamefont {Barros}, \citenamefont {Tretiak}, \citenamefont {Isayev},\ and\ \citenamefont {Roitberg}}]{dft-ccsd-transfer_nnp}%
  \BibitemOpen
  \bibfield  {author} {\bibinfo {author} {\bibfnamefont {J.~S.}\ \bibnamefont {Smith}}, \bibinfo {author} {\bibfnamefont {B.~T.}\ \bibnamefont {Nebgen}}, \bibinfo {author} {\bibfnamefont {R.}~\bibnamefont {Zubatyuk}}, \bibinfo {author} {\bibfnamefont {N.}~\bibnamefont {Lubbers}}, \bibinfo {author} {\bibfnamefont {C.}~\bibnamefont {Devereux}}, \bibinfo {author} {\bibfnamefont {K.}~\bibnamefont {Barros}}, \bibinfo {author} {\bibfnamefont {S.}~\bibnamefont {Tretiak}}, \bibinfo {author} {\bibfnamefont {O.}~\bibnamefont {Isayev}},\ and\ \bibinfo {author} {\bibfnamefont {A.~E.}\ \bibnamefont {Roitberg}},\ }\bibfield  {title} {\bibinfo {title} {Approaching coupled cluster accuracy with a general-purpose neural network potential through transfer learning},\ }\href@noop {} {\bibfield  {journal} {\bibinfo  {journal} {Nat Commun}\ }\textbf {\bibinfo {volume} {10}},\ \bibinfo {pages} {2903} (\bibinfo {year} {2019})}\BibitemShut {NoStop}%
\bibitem [{\citenamefont {Li}\ and\ \citenamefont {Guo}(2014)}]{H2O++H2_NNP}%
  \BibitemOpen
  \bibfield  {author} {\bibinfo {author} {\bibfnamefont {A.}~\bibnamefont {Li}}\ and\ \bibinfo {author} {\bibfnamefont {H.}~\bibnamefont {Guo}},\ }\bibfield  {title} {\bibinfo {title} {A nine-dimensional ab initio global potential energy surface for the {H$_2$O$^+$ + H$_2$ $\xrightarrow{}$ H$_3$O$^+$ + H} reaction},\ }\href@noop {} {\bibfield  {journal} {\bibinfo  {journal} {J. Chem. Phys.}\ }\textbf {\bibinfo {volume} {140}},\ \bibinfo {pages} {224313} (\bibinfo {year} {2014})}\BibitemShut {NoStop}%
\bibitem [{\citenamefont {Lu}\ \emph {et~al.}(2022)\citenamefont {Lu}, \citenamefont {Cheng}, \citenamefont {DiRisio}, \citenamefont {Finney}, \citenamefont {Boyer}, \citenamefont {Moonkaen}, \citenamefont {Sun}, \citenamefont {Lee}, \citenamefont {Deustua}, \citenamefont {Miller},\ and\ \citenamefont {McCoy}}]{C2H5nnp_miller}%
  \BibitemOpen
  \bibfield  {author} {\bibinfo {author} {\bibfnamefont {F.}~\bibnamefont {Lu}}, \bibinfo {author} {\bibfnamefont {L.}~\bibnamefont {Cheng}}, \bibinfo {author} {\bibfnamefont {R.~J.}\ \bibnamefont {DiRisio}}, \bibinfo {author} {\bibfnamefont {J.~M.}\ \bibnamefont {Finney}}, \bibinfo {author} {\bibfnamefont {M.~A.}\ \bibnamefont {Boyer}}, \bibinfo {author} {\bibfnamefont {P.}~\bibnamefont {Moonkaen}}, \bibinfo {author} {\bibfnamefont {J.}~\bibnamefont {Sun}}, \bibinfo {author} {\bibfnamefont {S.~J.~R.}\ \bibnamefont {Lee}}, \bibinfo {author} {\bibfnamefont {J.~E.}\ \bibnamefont {Deustua}}, \bibinfo {author} {\bibfnamefont {T.~F.}\ \bibnamefont {Miller}, \bibfnamefont {III}},\ and\ \bibinfo {author} {\bibfnamefont {A.~B.}\ \bibnamefont {McCoy}},\ }\bibfield  {title} {\bibinfo {title} {Fast near ab initio potential energy surfaces using machine learning},\ }\href@noop {} {\bibfield  {journal} {\bibinfo  {journal} {J. Phys. Chem. A}\ }\textbf {\bibinfo {volume} {126}},\ \bibinfo {pages} {4013} (\bibinfo {year}
  {2022})}\BibitemShut {NoStop}%
\bibitem [{\citenamefont {Aldossary}\ \emph {et~al.}(2024)\citenamefont {Aldossary}, \citenamefont {Campos-Gonzalez-Angulo}, \citenamefont {Pablo-García}, \citenamefont {Leong}, \citenamefont {Rajaonson}, \citenamefont {Thiede}, \citenamefont {Tom}, \citenamefont {Wang}, \citenamefont {Avagliano},\ and\ \citenamefont {Aspuru-Guzik}}]{Aspuru-Guzik-ML-QC}%
  \BibitemOpen
  \bibfield  {author} {\bibinfo {author} {\bibfnamefont {A.}~\bibnamefont {Aldossary}}, \bibinfo {author} {\bibfnamefont {J.~A.}\ \bibnamefont {Campos-Gonzalez-Angulo}}, \bibinfo {author} {\bibfnamefont {S.}~\bibnamefont {Pablo-García}}, \bibinfo {author} {\bibfnamefont {S.~X.}\ \bibnamefont {Leong}}, \bibinfo {author} {\bibfnamefont {E.~M.}\ \bibnamefont {Rajaonson}}, \bibinfo {author} {\bibfnamefont {L.}~\bibnamefont {Thiede}}, \bibinfo {author} {\bibfnamefont {G.}~\bibnamefont {Tom}}, \bibinfo {author} {\bibfnamefont {A.}~\bibnamefont {Wang}}, \bibinfo {author} {\bibfnamefont {D.}~\bibnamefont {Avagliano}},\ and\ \bibinfo {author} {\bibfnamefont {A.}~\bibnamefont {Aspuru-Guzik}},\ }\bibfield  {title} {\bibinfo {title} {In silico chemical experiments in the age of ai: From quantum chemistry to machine learning and back},\ }\href@noop {} {\bibfield  {journal} {\bibinfo  {journal} {Advanced Materials}\ }\textbf {\bibinfo {volume} {36}},\ \bibinfo {pages} {2402369} (\bibinfo {year} {2024})}\BibitemShut
  {NoStop}%
\bibitem [{\citenamefont {Behler}\ and\ \citenamefont {Parrinello}(2007)}]{BPNN}%
  \BibitemOpen
  \bibfield  {author} {\bibinfo {author} {\bibfnamefont {J.}~\bibnamefont {Behler}}\ and\ \bibinfo {author} {\bibfnamefont {M.}~\bibnamefont {Parrinello}},\ }\bibfield  {title} {\bibinfo {title} {Generalized neural-network representation of high-dimensional potential-energy surfaces},\ }\href@noop {} {\bibfield  {journal} {\bibinfo  {journal} {Phys. Rev. Lett.}\ }\textbf {\bibinfo {volume} {98}} (\bibinfo {year} {2007})}\BibitemShut {NoStop}%
\bibitem [{\citenamefont {Behler}(2021)}]{nnp_review1}%
  \BibitemOpen
  \bibfield  {author} {\bibinfo {author} {\bibfnamefont {J.}~\bibnamefont {Behler}},\ }\bibfield  {title} {\bibinfo {title} {Four generations of high-dimensional neural network potentials},\ }\href@noop {} {\bibfield  {journal} {\bibinfo  {journal} {Chem. Rev.}\ }\textbf {\bibinfo {volume} {121}},\ \bibinfo {pages} {10037} (\bibinfo {year} {2021})}\BibitemShut {NoStop}%
\bibitem [{\citenamefont {Vaswani}\ \emph {et~al.}(2017)\citenamefont {Vaswani}, \citenamefont {Shazeer}, \citenamefont {Parmar}, \citenamefont {Uszkoreit}, \citenamefont {Jones}, \citenamefont {Gomez}, \citenamefont {Kaiser},\ and\ \citenamefont {Polosukhin}}]{trans-attention}%
  \BibitemOpen
  \bibfield  {author} {\bibinfo {author} {\bibfnamefont {A.}~\bibnamefont {Vaswani}}, \bibinfo {author} {\bibfnamefont {N.}~\bibnamefont {Shazeer}}, \bibinfo {author} {\bibfnamefont {N.}~\bibnamefont {Parmar}}, \bibinfo {author} {\bibfnamefont {J.}~\bibnamefont {Uszkoreit}}, \bibinfo {author} {\bibfnamefont {L.}~\bibnamefont {Jones}}, \bibinfo {author} {\bibfnamefont {A.~N.}\ \bibnamefont {Gomez}}, \bibinfo {author} {\bibfnamefont {L.~u.}\ \bibnamefont {Kaiser}},\ and\ \bibinfo {author} {\bibfnamefont {I.}~\bibnamefont {Polosukhin}},\ }\bibfield  {title} {\bibinfo {title} {Attention is all you need},\ }in\ \href@noop {} {\emph {\bibinfo {booktitle} {Advances in Neural Information Processing Systems}}},\ Vol.~\bibinfo {volume} {30},\ \bibinfo {editor} {edited by\ \bibinfo {editor} {\bibfnamefont {I.}~\bibnamefont {Guyon}}, \bibinfo {editor} {\bibfnamefont {U.~V.}\ \bibnamefont {Luxburg}}, \bibinfo {editor} {\bibfnamefont {S.}~\bibnamefont {Bengio}}, \bibinfo {editor} {\bibfnamefont {H.}~\bibnamefont
  {Wallach}}, \bibinfo {editor} {\bibfnamefont {R.}~\bibnamefont {Fergus}}, \bibinfo {editor} {\bibfnamefont {S.}~\bibnamefont {Vishwanathan}},\ and\ \bibinfo {editor} {\bibfnamefont {R.}~\bibnamefont {Garnett}}}\ (\bibinfo  {publisher} {Curran Associates, Inc.},\ \bibinfo {year} {2017})\BibitemShut {NoStop}%
\bibitem [{\citenamefont {Li}\ and\ \citenamefont {Iyengar}(2015)}]{fragAIMD}%
  \BibitemOpen
  \bibfield  {author} {\bibinfo {author} {\bibfnamefont {J.}~\bibnamefont {Li}}\ and\ \bibinfo {author} {\bibfnamefont {S.~S.}\ \bibnamefont {Iyengar}},\ }\bibfield  {title} {\bibinfo {title} {Ab initio molecular dynamics using recursive, spatially separated, overlapping model subsystems mixed within an oniom based fragmentation energy extrapolation technique},\ }\href@noop {} {\bibfield  {journal} {\bibinfo  {journal} {J. Chem. Theory Comput.}\ }\textbf {\bibinfo {volume} {11}},\ \bibinfo {pages} {3978} (\bibinfo {year} {2015})}\BibitemShut {NoStop}%
\bibitem [{\citenamefont {Li}\ \emph {et~al.}(2016)\citenamefont {Li}, \citenamefont {Haycraft},\ and\ \citenamefont {Iyengar}}]{fragAIMD-elbo}%
  \BibitemOpen
  \bibfield  {author} {\bibinfo {author} {\bibfnamefont {J.}~\bibnamefont {Li}}, \bibinfo {author} {\bibfnamefont {C.}~\bibnamefont {Haycraft}},\ and\ \bibinfo {author} {\bibfnamefont {S.~S.}\ \bibnamefont {Iyengar}},\ }\bibfield  {title} {\bibinfo {title} {Hybrid, extended lagrangian $\textendash$ born-oppenheimer {\em ab initio} molecular dynamics using fragment-based electronic structure},\ }\href@noop {} {\bibfield  {journal} {\bibinfo  {journal} {J. Chem. Theory Comput.}\ }\textbf {\bibinfo {volume} {12}},\ \bibinfo {pages} {2493} (\bibinfo {year} {2016})}\BibitemShut {NoStop}%
\bibitem [{\citenamefont {Haycraft}\ \emph {et~al.}(2017)\citenamefont {Haycraft}, \citenamefont {Li},\ and\ \citenamefont {Iyengar}}]{fragAIMD-CC}%
  \BibitemOpen
  \bibfield  {author} {\bibinfo {author} {\bibfnamefont {C.}~\bibnamefont {Haycraft}}, \bibinfo {author} {\bibfnamefont {J.}~\bibnamefont {Li}},\ and\ \bibinfo {author} {\bibfnamefont {S.~S.}\ \bibnamefont {Iyengar}},\ }\bibfield  {title} {\bibinfo {title} {Efficient, ``on-the-fly'' born$\textendash$oppenheimer and car$\textendash$parrinello$\textendash$type dynamics with coupled cluster accuracy through fragment based electronic structure},\ }\href@noop {} {\bibfield  {journal} {\bibinfo  {journal} {J. Chem. Theory Comput.}\ }\textbf {\bibinfo {volume} {13}},\ \bibinfo {pages} {21887} (\bibinfo {year} {2017})}\BibitemShut {NoStop}%
\bibitem [{\citenamefont {Ricard}\ \emph {et~al.}(2018)\citenamefont {Ricard}, \citenamefont {Haycraft},\ and\ \citenamefont {Iyengar}}]{CGAIMD}%
  \BibitemOpen
  \bibfield  {author} {\bibinfo {author} {\bibfnamefont {T.~C.}\ \bibnamefont {Ricard}}, \bibinfo {author} {\bibfnamefont {C.}~\bibnamefont {Haycraft}},\ and\ \bibinfo {author} {\bibfnamefont {S.~S.}\ \bibnamefont {Iyengar}},\ }\bibfield  {title} {\bibinfo {title} {Adaptive, geometric networks for efficient coarse-grained {\em ab initio} molecular dynamics with post-hartree-fock accuracy},\ }\href@noop {} {\bibfield  {journal} {\bibinfo  {journal} {J. Chem. Theory Comput.}\ }\textbf {\bibinfo {volume} {14}},\ \bibinfo {pages} {2852} (\bibinfo {year} {2018})}\BibitemShut {NoStop}%
\bibitem [{\citenamefont {Ricard}\ and\ \citenamefont {Iyengar}(2018)}]{frag-BSSE-AIMD}%
  \BibitemOpen
  \bibfield  {author} {\bibinfo {author} {\bibfnamefont {T.~C.}\ \bibnamefont {Ricard}}\ and\ \bibinfo {author} {\bibfnamefont {S.~S.}\ \bibnamefont {Iyengar}},\ }\bibfield  {title} {\bibinfo {title} {Efficiently capturing weak interactions in {\em ab initio} molecular dynamics through ``on-the-fly'' basis set extrapolation},\ }\href@noop {} {\bibfield  {journal} {\bibinfo  {journal} {J. Chem. Theory Comput.}\ }\textbf {\bibinfo {volume} {14}},\ \bibinfo {pages} {5535} (\bibinfo {year} {2018})}\BibitemShut {NoStop}%
\bibitem [{\citenamefont {Kumar}\ and\ \citenamefont {Iyengar}(2019)}]{frag-AIMD-multitop}%
  \BibitemOpen
  \bibfield  {author} {\bibinfo {author} {\bibfnamefont {A.}~\bibnamefont {Kumar}}\ and\ \bibinfo {author} {\bibfnamefont {S.~S.}\ \bibnamefont {Iyengar}},\ }\bibfield  {title} {\bibinfo {title} {Fragment-based electronic structure for potential energy surfaces using a superposition of fragmentation topologies},\ }\href@noop {} {\bibfield  {journal} {\bibinfo  {journal} {J. Chem. Theory Comput.}\ }\textbf {\bibinfo {volume} {15}},\ \bibinfo {pages} {5769} (\bibinfo {year} {2019})}\BibitemShut {NoStop}%
\bibitem [{\citenamefont {Ricard}\ and\ \citenamefont {Iyengar}(2020)}]{fragPBC}%
  \BibitemOpen
  \bibfield  {author} {\bibinfo {author} {\bibfnamefont {T.~C.}\ \bibnamefont {Ricard}}\ and\ \bibinfo {author} {\bibfnamefont {S.~S.}\ \bibnamefont {Iyengar}},\ }\bibfield  {title} {\bibinfo {title} {An efficient and accurate approach to estimate hybrid functional and large basis set contributions to condensed phase systems and molecule-surface interactions},\ }\href@noop {} {\bibfield  {journal} {\bibinfo  {journal} {J. Chem. Theory Comput.}\ }\textbf {\bibinfo {volume} {16}},\ \bibinfo {pages} {4790} (\bibinfo {year} {2020})}\BibitemShut {NoStop}%
\bibitem [{\citenamefont {Ricard}\ \emph {et~al.}(2020)\citenamefont {Ricard}, \citenamefont {Kumar},\ and\ \citenamefont {Iyengar}}]{fragIJQC-review}%
  \BibitemOpen
  \bibfield  {author} {\bibinfo {author} {\bibfnamefont {T.~C.}\ \bibnamefont {Ricard}}, \bibinfo {author} {\bibfnamefont {A.}~\bibnamefont {Kumar}},\ and\ \bibinfo {author} {\bibfnamefont {S.~S.}\ \bibnamefont {Iyengar}},\ }\bibfield  {title} {\bibinfo {title} {Embedded, graph‐theoretically defined many‐body approximations for wavefunction‐in‐dft and dft‐in‐dft: Applications to gas‐ and condensed‐phase ab initio molecular dynamics, and potential surfaces for quantum nuclear effects},\ }\href@noop {} {\bibfield  {journal} {\bibinfo  {journal} {Int. J. Quantum Chem.}\ }\textbf {\bibinfo {volume} {120}},\ \bibinfo {pages} {e26244} (\bibinfo {year} {2020})}\BibitemShut {NoStop}%
\bibitem [{\citenamefont {Ricard}\ \emph {et~al.}(2023)\citenamefont {Ricard}, \citenamefont {Zhu},\ and\ \citenamefont {Iyengar}}]{frag-PFOA}%
  \BibitemOpen
  \bibfield  {author} {\bibinfo {author} {\bibfnamefont {T.~C.}\ \bibnamefont {Ricard}}, \bibinfo {author} {\bibfnamefont {X.}~\bibnamefont {Zhu}},\ and\ \bibinfo {author} {\bibfnamefont {S.~S.}\ \bibnamefont {Iyengar}},\ }\bibfield  {title} {\bibinfo {title} {Capturing weak interactions in surface adsorbate systems at coupled cluster accuracy: a graph-theoretic molecular fragmentation approach improved through machine learning},\ }\href@noop {} {\bibfield  {journal} {\bibinfo  {journal} {J. Chem. Theory Comput.}\ }\textbf {\bibinfo {volume} {19}},\ \bibinfo {pages} {8541} (\bibinfo {year} {2023})}\BibitemShut {NoStop}%
\bibitem [{\citenamefont {Kumar}\ \emph {et~al.}(2021)\citenamefont {Kumar}, \citenamefont {DeGregorio},\ and\ \citenamefont {Iyengar}}]{frag-AIMD-multitop-2}%
  \BibitemOpen
  \bibfield  {author} {\bibinfo {author} {\bibfnamefont {A.}~\bibnamefont {Kumar}}, \bibinfo {author} {\bibfnamefont {N.}~\bibnamefont {DeGregorio}},\ and\ \bibinfo {author} {\bibfnamefont {S.~S.}\ \bibnamefont {Iyengar}},\ }\bibfield  {title} {\bibinfo {title} {Graph-theory-based molecular fragmentation for efficient and accurate potential surface calculations in multiple dimensions},\ }\href@noop {} {\bibfield  {journal} {\bibinfo  {journal} {J. Chem. Theory Comput.}\ }\textbf {\bibinfo {volume} {17}},\ \bibinfo {pages} {6671} (\bibinfo {year} {2021})}\BibitemShut {NoStop}%
\bibitem [{\citenamefont {Zhang}\ \emph {et~al.}(2021)\citenamefont {Zhang}, \citenamefont {Ricard}, \citenamefont {Haycraft},\ and\ \citenamefont {Iyengar}}]{Harry-weighted-graphs}%
  \BibitemOpen
  \bibfield  {author} {\bibinfo {author} {\bibfnamefont {J.~H.}\ \bibnamefont {Zhang}}, \bibinfo {author} {\bibfnamefont {T.~C.}\ \bibnamefont {Ricard}}, \bibinfo {author} {\bibfnamefont {C.}~\bibnamefont {Haycraft}},\ and\ \bibinfo {author} {\bibfnamefont {S.~S.}\ \bibnamefont {Iyengar}},\ }\bibfield  {title} {\bibinfo {title} {Weighted-graph-theoretic methods for many-body corrections within oniom: Smooth aimd and the role of high-order many-body terms},\ }\href@noop {} {\bibfield  {journal} {\bibinfo  {journal} {J. Chem. Theory Comput.}\ }\textbf {\bibinfo {volume} {17}},\ \bibinfo {pages} {2672} (\bibinfo {year} {2021})}\BibitemShut {NoStop}%
\bibitem [{\citenamefont {Zhu}\ and\ \citenamefont {Iyengar}(2022)}]{frag-ML-Xiao}%
  \BibitemOpen
  \bibfield  {author} {\bibinfo {author} {\bibfnamefont {X.}~\bibnamefont {Zhu}}\ and\ \bibinfo {author} {\bibfnamefont {S.~S.}\ \bibnamefont {Iyengar}},\ }\bibfield  {title} {\bibinfo {title} {Graph theoretic molecular fragmentation for multidimensional potential energy surfaces yield an adaptive and general transfer machine learning protocol},\ }\href@noop {} {\bibfield  {journal} {\bibinfo  {journal} {J. Chem. Theory Comput.}\ }\textbf {\bibinfo {volume} {18}},\ \bibinfo {pages} {5125} (\bibinfo {year} {2022})}\BibitemShut {NoStop}%
\bibitem [{\citenamefont {Zhang}\ and\ \citenamefont {Iyengar}(2022)}]{frag-QC-Harry}%
  \BibitemOpen
  \bibfield  {author} {\bibinfo {author} {\bibfnamefont {J.~H.}\ \bibnamefont {Zhang}}\ and\ \bibinfo {author} {\bibfnamefont {S.~S.}\ \bibnamefont {Iyengar}},\ }\bibfield  {title} {\bibinfo {title} {Graph-$\ket{Q}\bra{C}$: A graph-based quantum-classical algorithm for efficient electronic structure on hybrid quantum/classical hardware systems: Improved quantum circuit depth performance},\ }\href@noop {} {\bibfield  {journal} {\bibinfo  {journal} {J. Chem. Theory Comput.}\ }\textbf {\bibinfo {volume} {18}},\ \bibinfo {pages} {2885} (\bibinfo {year} {2022})}\BibitemShut {NoStop}%
\bibitem [{\citenamefont {Kumar}\ \emph {et~al.}(2022)\citenamefont {Kumar}, \citenamefont {DeGregorio}, \citenamefont {Ricard},\ and\ \citenamefont {Iyengar}}]{frag-TN-Anup}%
  \BibitemOpen
  \bibfield  {author} {\bibinfo {author} {\bibfnamefont {A.}~\bibnamefont {Kumar}}, \bibinfo {author} {\bibfnamefont {N.}~\bibnamefont {DeGregorio}}, \bibinfo {author} {\bibfnamefont {T.}~\bibnamefont {Ricard}},\ and\ \bibinfo {author} {\bibfnamefont {S.~S.}\ \bibnamefont {Iyengar}},\ }\bibfield  {title} {\bibinfo {title} {Graph-theoretic molecular fragmentation for potential surfaces leads naturally to a tensor network form and allows accurate and efficient quantum nuclear dynamics},\ }\href@noop {} {\bibfield  {journal} {\bibinfo  {journal} {J. Chem. Theory Comput.}\ }\textbf {\bibinfo {volume} {18}},\ \bibinfo {pages} {7243} (\bibinfo {year} {2022})}\BibitemShut {NoStop}%
\bibitem [{\citenamefont {Iyengar}\ \emph {et~al.}(2023{\natexlab{a}})\citenamefont {Iyengar}, \citenamefont {Saha}, \citenamefont {Dwivedi}, \citenamefont {Lopez-Ruiz}, \citenamefont {Kumar}, \citenamefont {Zhang}, \citenamefont {Ricard}, \citenamefont {Richerme},\ and\ \citenamefont {Sabry}}]{SSI-Review1-QC-ES-QN}%
  \BibitemOpen
  \bibfield  {author} {\bibinfo {author} {\bibfnamefont {S.~S.}\ \bibnamefont {Iyengar}}, \bibinfo {author} {\bibfnamefont {D.}~\bibnamefont {Saha}}, \bibinfo {author} {\bibfnamefont {A.}~\bibnamefont {Dwivedi}}, \bibinfo {author} {\bibfnamefont {M.~A.}\ \bibnamefont {Lopez-Ruiz}}, \bibinfo {author} {\bibfnamefont {A.}~\bibnamefont {Kumar}}, \bibinfo {author} {\bibfnamefont {J.~H.}\ \bibnamefont {Zhang}}, \bibinfo {author} {\bibfnamefont {T.~C.}\ \bibnamefont {Ricard}}, \bibinfo {author} {\bibfnamefont {P.}~\bibnamefont {Richerme}},\ and\ \bibinfo {author} {\bibfnamefont {A.}~\bibnamefont {Sabry}},\ }\bibfield  {title} {\bibinfo {title} {Quantum algorithms for the study of electronic structure and molecular dynamics: Novel computational protocols},\ }in\ \href@noop {} {\emph {\bibinfo {booktitle} {Comprehensive Computational Chemistry}}}\ (\bibinfo  {publisher} {Elsevier},\ \bibinfo {year} {2023})\BibitemShut {NoStop}%
\bibitem [{\citenamefont {Iyengar}\ \emph {et~al.}(2023{\natexlab{b}})\citenamefont {Iyengar}, \citenamefont {Zhang}, \citenamefont {Saha},\ and\ \citenamefont {Ricard}}]{frag-QC-2}%
  \BibitemOpen
  \bibfield  {author} {\bibinfo {author} {\bibfnamefont {S.~S.}\ \bibnamefont {Iyengar}}, \bibinfo {author} {\bibfnamefont {J.~H.}\ \bibnamefont {Zhang}}, \bibinfo {author} {\bibfnamefont {D.}~\bibnamefont {Saha}},\ and\ \bibinfo {author} {\bibfnamefont {T.~C.}\ \bibnamefont {Ricard}},\ }\bibfield  {title} {\bibinfo {title} {Graph-$\ket{Q}\bra{C}$: A quantum algorithm with reduced quantum circuit depth for electronic structure},\ }\href@noop {} {\bibfield  {journal} {\bibinfo  {journal} {J. Phys. Chem. A}\ }\textbf {\bibinfo {volume} {127}},\ \bibinfo {pages} {9334} (\bibinfo {year} {2023}{\natexlab{b}})}\BibitemShut {NoStop}%
\bibitem [{\citenamefont {Iyengar}\ \emph {et~al.}(2024)\citenamefont {Iyengar}, \citenamefont {Ricard},\ and\ \citenamefont {Zhu}}]{frag-ONIOM-frag-reformulation}%
  \BibitemOpen
  \bibfield  {author} {\bibinfo {author} {\bibfnamefont {S.~S.}\ \bibnamefont {Iyengar}}, \bibinfo {author} {\bibfnamefont {T.~C.}\ \bibnamefont {Ricard}},\ and\ \bibinfo {author} {\bibfnamefont {X.}~\bibnamefont {Zhu}},\ }\bibfield  {title} {\bibinfo {title} {A reformulation of all oniom-type molecular fragmentation approaches using graph theory-based projection operators: Applications to dynamics, molecular potential surfaces, and machine learning and quantum computing},\ }\href@noop {} {\bibfield  {journal} {\bibinfo  {journal} {J. Phys. Chem. A}\ }\textbf {\bibinfo {volume} {128}},\ \bibinfo {pages} {466} (\bibinfo {year} {2024})}\BibitemShut {NoStop}%
\bibitem [{\citenamefont {Zhang}\ \emph {et~al.}(1990)\citenamefont {Zhang}, \citenamefont {Itoh}, \citenamefont {Tanida},\ and\ \citenamefont {Ichioka}}]{Zhang_cnn}%
  \BibitemOpen
  \bibfield  {author} {\bibinfo {author} {\bibfnamefont {W.}~\bibnamefont {Zhang}}, \bibinfo {author} {\bibfnamefont {K.}~\bibnamefont {Itoh}}, \bibinfo {author} {\bibfnamefont {J.}~\bibnamefont {Tanida}},\ and\ \bibinfo {author} {\bibfnamefont {Y.}~\bibnamefont {Ichioka}},\ }\bibfield  {title} {\bibinfo {title} {Parallel distributed processing model with local space-invariant interconnections and its optical architecture},\ }\href@noop {} {\bibfield  {journal} {\bibinfo  {journal} {Appl. Opt.}\ }\textbf {\bibinfo {volume} {29}},\ \bibinfo {pages} {4790} (\bibinfo {year} {1990})}\BibitemShut {NoStop}%
\bibitem [{\citenamefont {Duvenaud}\ \emph {et~al.}(2015)\citenamefont {Duvenaud}, \citenamefont {Maclaurin}, \citenamefont {Iparraguirre}, \citenamefont {Bombarell}, \citenamefont {Hirzel}, \citenamefont {Aspuru-Guzik},\ and\ \citenamefont {Adams}}]{duvenaud2015convolutional}%
  \BibitemOpen
  \bibfield  {author} {\bibinfo {author} {\bibfnamefont {D.~K.}\ \bibnamefont {Duvenaud}}, \bibinfo {author} {\bibfnamefont {D.}~\bibnamefont {Maclaurin}}, \bibinfo {author} {\bibfnamefont {J.}~\bibnamefont {Iparraguirre}}, \bibinfo {author} {\bibfnamefont {R.}~\bibnamefont {Bombarell}}, \bibinfo {author} {\bibfnamefont {T.}~\bibnamefont {Hirzel}}, \bibinfo {author} {\bibfnamefont {A.}~\bibnamefont {Aspuru-Guzik}},\ and\ \bibinfo {author} {\bibfnamefont {R.~P.}\ \bibnamefont {Adams}},\ }\bibfield  {title} {\bibinfo {title} {Convolutional networks on graphs for learning molecular fingerprints},\ }in\ \href@noop {} {\emph {\bibinfo {booktitle} {Advances in neural information processing systems}}}\ (\bibinfo {year} {2015})\ pp.\ \bibinfo {pages} {2224--2232}\BibitemShut {NoStop}%
\bibitem [{\citenamefont {Veličković}\ \emph {et~al.}(2018)\citenamefont {Veličković}, \citenamefont {Cucurull}, \citenamefont {Casanova}, \citenamefont {Romero}, \citenamefont {Liò},\ and\ \citenamefont {Bengio}}]{veličković2018graphattentionnetworks}%
  \BibitemOpen
  \bibfield  {author} {\bibinfo {author} {\bibfnamefont {P.}~\bibnamefont {Veličković}}, \bibinfo {author} {\bibfnamefont {G.}~\bibnamefont {Cucurull}}, \bibinfo {author} {\bibfnamefont {A.}~\bibnamefont {Casanova}}, \bibinfo {author} {\bibfnamefont {A.}~\bibnamefont {Romero}}, \bibinfo {author} {\bibfnamefont {P.}~\bibnamefont {Liò}},\ and\ \bibinfo {author} {\bibfnamefont {Y.}~\bibnamefont {Bengio}},\ }\href@noop {} {\bibinfo {title} {Graph attention networks}} (\bibinfo {year} {2018})\BibitemShut {NoStop}%
\bibitem [{\citenamefont {Fournier}\ \emph {et~al.}(2015)\citenamefont {Fournier}, \citenamefont {Wolke}, \citenamefont {Johnson}, \citenamefont {Odbadrakh}, \citenamefont {Jordan}, \citenamefont {Kathmann},\ and\ \citenamefont {Xantheas}}]{Johnson-Jordan-H+-water-cluster-sizes}%
  \BibitemOpen
  \bibfield  {author} {\bibinfo {author} {\bibfnamefont {J.~A.}\ \bibnamefont {Fournier}}, \bibinfo {author} {\bibfnamefont {C.~T.}\ \bibnamefont {Wolke}}, \bibinfo {author} {\bibfnamefont {M.~A.}\ \bibnamefont {Johnson}}, \bibinfo {author} {\bibfnamefont {T.~T.}\ \bibnamefont {Odbadrakh}}, \bibinfo {author} {\bibfnamefont {K.~D.}\ \bibnamefont {Jordan}}, \bibinfo {author} {\bibfnamefont {S.~M.}\ \bibnamefont {Kathmann}},\ and\ \bibinfo {author} {\bibfnamefont {S.~S.}\ \bibnamefont {Xantheas}},\ }\bibfield  {title} {\bibinfo {title} {Snapshots of proton accommodation at a microscopic water surface: Understanding the vibrational spectral signatures of the charge defect in cryogenically cooled {H${\left( \text{H}_2\text{O}\right)}_n^+$}=2–28 clusters},\ }\href@noop {} {\bibfield  {journal} {\bibinfo  {journal} {J. Phys. Chem. A}\ }\textbf {\bibinfo {volume} {119}},\ \bibinfo {pages} {37} (\bibinfo {year} {2015})}\BibitemShut {NoStop}%
\bibitem [{\citenamefont {Shin}\ \emph {et~al.}(2004)\citenamefont {Shin}, \citenamefont {Hammer}, \citenamefont {Diken}, \citenamefont {Johnson}, \citenamefont {Walters}, \citenamefont {Jaeger}, \citenamefont {Duncan}, \citenamefont {Christie},\ and\ \citenamefont {Jordan}}]{johnson-jordan-21mer}%
  \BibitemOpen
  \bibfield  {author} {\bibinfo {author} {\bibfnamefont {J.-W.}\ \bibnamefont {Shin}}, \bibinfo {author} {\bibfnamefont {N.~I.}\ \bibnamefont {Hammer}}, \bibinfo {author} {\bibfnamefont {E.~G.}\ \bibnamefont {Diken}}, \bibinfo {author} {\bibfnamefont {M.~A.}\ \bibnamefont {Johnson}}, \bibinfo {author} {\bibfnamefont {R.~S.}\ \bibnamefont {Walters}}, \bibinfo {author} {\bibfnamefont {T.~D.}\ \bibnamefont {Jaeger}}, \bibinfo {author} {\bibfnamefont {M.~A.}\ \bibnamefont {Duncan}}, \bibinfo {author} {\bibfnamefont {R.~A.}\ \bibnamefont {Christie}},\ and\ \bibinfo {author} {\bibfnamefont {K.~D.}\ \bibnamefont {Jordan}},\ }\bibfield  {title} {\bibinfo {title} {Infrared signature of structures associated with the {H$^{+}$(H$_{2}$O)$_{n}$} (n = 6 to 27) clusters},\ }\href@noop {} {\bibfield  {journal} {\bibinfo  {journal} {Science}\ }\textbf {\bibinfo {volume} {304}},\ \bibinfo {pages} {1137} (\bibinfo {year} {2004})}\BibitemShut {NoStop}%
\bibitem [{\citenamefont {Iyengar}\ \emph {et~al.}(2005)\citenamefont {Iyengar}, \citenamefont {Petersen}, \citenamefont {Day}, \citenamefont {Burnham}, \citenamefont {Teige},\ and\ \citenamefont {Voth}}]{admp-21mer}%
  \BibitemOpen
  \bibfield  {author} {\bibinfo {author} {\bibfnamefont {S.~S.}\ \bibnamefont {Iyengar}}, \bibinfo {author} {\bibfnamefont {M.~K.}\ \bibnamefont {Petersen}}, \bibinfo {author} {\bibfnamefont {T.~J.~F.}\ \bibnamefont {Day}}, \bibinfo {author} {\bibfnamefont {C.~J.}\ \bibnamefont {Burnham}}, \bibinfo {author} {\bibfnamefont {V.~E.}\ \bibnamefont {Teige}},\ and\ \bibinfo {author} {\bibfnamefont {G.~A.}\ \bibnamefont {Voth}},\ }\bibfield  {title} {\bibinfo {title} {The properties of ion-water clusters. i. the protonated 21-water cluster},\ }\href@noop {} {\bibfield  {journal} {\bibinfo  {journal} {J. Chem. Phys.}\ }\textbf {\bibinfo {volume} {123}},\ \bibinfo {pages} {084309} (\bibinfo {year} {2005})}\BibitemShut {NoStop}%
\bibitem [{\citenamefont {Iyengar}(2007)}]{admp-21mer-2}%
  \BibitemOpen
  \bibfield  {author} {\bibinfo {author} {\bibfnamefont {S.~S.}\ \bibnamefont {Iyengar}},\ }\bibfield  {title} {\bibinfo {title} {Further analysis of the dynamically averaged vibrational spectrum for the ``magic'' protonated 21-water cluster},\ }\href@noop {} {\bibfield  {journal} {\bibinfo  {journal} {J. Chem. Phys.}\ }\textbf {\bibinfo {volume} {126}},\ \bibinfo {pages} {216101} (\bibinfo {year} {2007})}\BibitemShut {NoStop}%
\bibitem [{\citenamefont {Hammer}\ \emph {et~al.}(2005)\citenamefont {Hammer}, \citenamefont {Diken}, \citenamefont {Roscioli}, \citenamefont {Johnson}, \citenamefont {Myshakin}, \citenamefont {Jordan}, \citenamefont {McCoy}, \citenamefont {Huang}, \citenamefont {Bowman},\ and\ \citenamefont {Carter}}]{Johnson-Jordan-Zundel-JCP}%
  \BibitemOpen
  \bibfield  {author} {\bibinfo {author} {\bibfnamefont {N.~I.}\ \bibnamefont {Hammer}}, \bibinfo {author} {\bibfnamefont {E.~G.}\ \bibnamefont {Diken}}, \bibinfo {author} {\bibfnamefont {J.~R.}\ \bibnamefont {Roscioli}}, \bibinfo {author} {\bibfnamefont {M.~A.}\ \bibnamefont {Johnson}}, \bibinfo {author} {\bibfnamefont {E.~M.}\ \bibnamefont {Myshakin}}, \bibinfo {author} {\bibfnamefont {K.~D.}\ \bibnamefont {Jordan}}, \bibinfo {author} {\bibfnamefont {A.~B.}\ \bibnamefont {McCoy}}, \bibinfo {author} {\bibfnamefont {X.}~\bibnamefont {Huang}}, \bibinfo {author} {\bibfnamefont {J.~M.}\ \bibnamefont {Bowman}},\ and\ \bibinfo {author} {\bibfnamefont {S.}~\bibnamefont {Carter}},\ }\bibfield  {title} {\bibinfo {title} {The vibrational predissociation spectra of the {H}$_5${O}$_2^+ \cdot${RG}$_n$({RG = Ar,Ne}) clusters: Correlation of the solvent perturbations in the free oh and shared proton transitions of the zundel ion},\ }\href@noop {} {\bibfield  {journal} {\bibinfo  {journal} {J. Chem. Phys.}\ }\textbf
  {\bibinfo {volume} {122}},\ \bibinfo {eid} {244301} (\bibinfo {year} {2005})}\BibitemShut {NoStop}%
\bibitem [{\citenamefont {Headrick}\ \emph {et~al.}(2005)\citenamefont {Headrick}, \citenamefont {Diken}, \citenamefont {Walters}, \citenamefont {Hammer}, \citenamefont {Christie}, \citenamefont {Cui}, \citenamefont {Myshakin}, \citenamefont {Duncan}, \citenamefont {Johnson},\ and\ \citenamefont {Jordan}}]{Johnson-Jordan-Zundel-Science}%
  \BibitemOpen
  \bibfield  {author} {\bibinfo {author} {\bibfnamefont {J.~M.}\ \bibnamefont {Headrick}}, \bibinfo {author} {\bibfnamefont {E.~G.}\ \bibnamefont {Diken}}, \bibinfo {author} {\bibfnamefont {R.~S.}\ \bibnamefont {Walters}}, \bibinfo {author} {\bibfnamefont {N.~I.}\ \bibnamefont {Hammer}}, \bibinfo {author} {\bibfnamefont {R.~A.}\ \bibnamefont {Christie}}, \bibinfo {author} {\bibfnamefont {J.}~\bibnamefont {Cui}}, \bibinfo {author} {\bibfnamefont {E.~M.}\ \bibnamefont {Myshakin}}, \bibinfo {author} {\bibfnamefont {M.~A.}\ \bibnamefont {Duncan}}, \bibinfo {author} {\bibfnamefont {M.~A.}\ \bibnamefont {Johnson}},\ and\ \bibinfo {author} {\bibfnamefont {K.}~\bibnamefont {Jordan}},\ }\bibfield  {title} {\bibinfo {title} {Spectral signatures of hydrated proton vibrations in water clusters},\ }\href@noop {} {\bibfield  {journal} {\bibinfo  {journal} {Science}\ }\textbf {\bibinfo {volume} {308}},\ \bibinfo {pages} {1765} (\bibinfo {year} {2005})}\BibitemShut {NoStop}%
\bibitem [{\citenamefont {Dietrick}\ and\ \citenamefont {Iyengar}(2012{\natexlab{a}})}]{Zundel-scott}%
  \BibitemOpen
  \bibfield  {author} {\bibinfo {author} {\bibfnamefont {S.~M.}\ \bibnamefont {Dietrick}}\ and\ \bibinfo {author} {\bibfnamefont {S.~S.}\ \bibnamefont {Iyengar}},\ }\bibfield  {title} {\bibinfo {title} {Constructing periodic phase space orbits from ab initio molecular dynamics trajectories to analyze vibrational spectra: Case study of the zundel {(H$_5$O$_2^+$)} cation},\ }\href@noop {} {\bibfield  {journal} {\bibinfo  {journal} {J. Chem. Theory and Comp.}\ }\textbf {\bibinfo {volume} {8}},\ \bibinfo {pages} {4876} (\bibinfo {year} {2012}{\natexlab{a}})}\BibitemShut {NoStop}%
\bibitem [{\citenamefont {Asmis}\ \emph {et~al.}(2003)\citenamefont {Asmis}, \citenamefont {Pivonka}, \citenamefont {Santambrogio}, \citenamefont {Brümmer}, \citenamefont {Kaposta}, \citenamefont {Neumark},\ and\ \citenamefont {W\"{o}ste}}]{Asmis-Zundel}%
  \BibitemOpen
  \bibfield  {author} {\bibinfo {author} {\bibfnamefont {K.~R.}\ \bibnamefont {Asmis}}, \bibinfo {author} {\bibfnamefont {N.~L.}\ \bibnamefont {Pivonka}}, \bibinfo {author} {\bibfnamefont {G.}~\bibnamefont {Santambrogio}}, \bibinfo {author} {\bibfnamefont {M.}~\bibnamefont {Brümmer}}, \bibinfo {author} {\bibfnamefont {C.}~\bibnamefont {Kaposta}}, \bibinfo {author} {\bibfnamefont {D.~M.}\ \bibnamefont {Neumark}},\ and\ \bibinfo {author} {\bibfnamefont {L.}~\bibnamefont {W\"{o}ste}},\ }\bibfield  {title} {\bibinfo {title} {Gas-phase infrared spectrum of the protonated water dimer},\ }\href@noop {} {\bibfield  {journal} {\bibinfo  {journal} {Science}\ }\textbf {\bibinfo {volume} {299}},\ \bibinfo {pages} {1375} (\bibinfo {year} {2003})}\BibitemShut {NoStop}%
\bibitem [{\citenamefont {Vendrell}\ \emph {et~al.}(2007)\citenamefont {Vendrell}, \citenamefont {Gatti},\ and\ \citenamefont {Meyer}}]{HDMeyer-Zundel-1}%
  \BibitemOpen
  \bibfield  {author} {\bibinfo {author} {\bibfnamefont {O.}~\bibnamefont {Vendrell}}, \bibinfo {author} {\bibfnamefont {F.}~\bibnamefont {Gatti}},\ and\ \bibinfo {author} {\bibfnamefont {H.-D.}\ \bibnamefont {Meyer}},\ }\bibfield  {title} {\bibinfo {title} {Dynamics and infrared spectroscopy of the protonated water dimer},\ }\href@noop {} {\bibfield  {journal} {\bibinfo  {journal} {Ang. Chem. Intl. Ed.}\ }\textbf {\bibinfo {volume} {46}},\ \bibinfo {pages} {6918} (\bibinfo {year} {2007})}\BibitemShut {NoStop}%
\bibitem [{\citenamefont {Diken}\ \emph {et~al.}(2005)\citenamefont {Diken}, \citenamefont {Headrick}, \citenamefont {Roscioli}, \citenamefont {Bopp}, \citenamefont {Johnson},\ and\ \citenamefont {McCoy}}]{Johnson-Zundel-OH-H2O-quantum}%
  \BibitemOpen
  \bibfield  {author} {\bibinfo {author} {\bibfnamefont {E.~G.}\ \bibnamefont {Diken}}, \bibinfo {author} {\bibfnamefont {J.~M.}\ \bibnamefont {Headrick}}, \bibinfo {author} {\bibfnamefont {J.~R.}\ \bibnamefont {Roscioli}}, \bibinfo {author} {\bibfnamefont {J.~C.}\ \bibnamefont {Bopp}}, \bibinfo {author} {\bibfnamefont {M.~A.}\ \bibnamefont {Johnson}},\ and\ \bibinfo {author} {\bibfnamefont {A.~B.}\ \bibnamefont {McCoy}},\ }\bibfield  {title} {\bibinfo {title} {Fundamental excitations of the shared proton in the {H$_3$O$_2^-$} and {H$_5$O$_2^+$} complexes},\ }\href@noop {} {\bibfield  {journal} {\bibinfo  {journal} {J. Phys. Chem. A}\ }\textbf {\bibinfo {volume} {109}},\ \bibinfo {pages} {1487} (\bibinfo {year} {2005})}\BibitemShut {NoStop}%
\bibitem [{\citenamefont {Robertson}\ \emph {et~al.}(2003)\citenamefont {Robertson}, \citenamefont {Diken}, \citenamefont {Price}, \citenamefont {Shin},\ and\ \citenamefont {Johnson}}]{JohnsonOH-Science}%
  \BibitemOpen
  \bibfield  {author} {\bibinfo {author} {\bibfnamefont {W.~H.}\ \bibnamefont {Robertson}}, \bibinfo {author} {\bibfnamefont {E.~G.}\ \bibnamefont {Diken}}, \bibinfo {author} {\bibfnamefont {E.~A.}\ \bibnamefont {Price}}, \bibinfo {author} {\bibfnamefont {J.-W.}\ \bibnamefont {Shin}},\ and\ \bibinfo {author} {\bibfnamefont {M.~A.}\ \bibnamefont {Johnson}},\ }\bibfield  {title} {\bibinfo {title} {Spectroscopic determination of the oh- solvation shell in the {OH$^-$(H$_2$O)$_n$} clusters},\ }\href@noop {} {\bibfield  {journal} {\bibinfo  {journal} {Science}\ }\textbf {\bibinfo {volume} {299}},\ \bibinfo {pages} {1367} (\bibinfo {year} {2003})}\BibitemShut {NoStop}%
\bibitem [{\citenamefont {Lorentz}\ \emph {et~al.}(1996)\citenamefont {Lorentz}, \citenamefont {Golitschek},\ and\ \citenamefont {Makovoz}}]{Kolmogorov-HDMR}%
  \BibitemOpen
  \bibfield  {author} {\bibinfo {author} {\bibfnamefont {G.}~\bibnamefont {Lorentz}}, \bibinfo {author} {\bibfnamefont {M.}~\bibnamefont {Golitschek}},\ and\ \bibinfo {author} {\bibfnamefont {Y.}~\bibnamefont {Makovoz}},\ }\href@noop {} {\emph {\bibinfo {title} {Constructive Approximation}}}\ (\bibinfo  {publisher} {Springer, New York},\ \bibinfo {year} {1996})\BibitemShut {NoStop}%
\bibitem [{\citenamefont {Girosi}\ and\ \citenamefont {Poggio}(1989)}]{Kolmogorov-HDMR-2}%
  \BibitemOpen
  \bibfield  {author} {\bibinfo {author} {\bibfnamefont {F.}~\bibnamefont {Girosi}}\ and\ \bibinfo {author} {\bibfnamefont {T.}~\bibnamefont {Poggio}},\ }\bibfield  {title} {\bibinfo {title} {Representation properties of networks: Kolmogorov’s theorem is irrelevant},\ }\href@noop {} {\bibfield  {journal} {\bibinfo  {journal} {Neurocomputing}\ }\textbf {\bibinfo {volume} {1}},\ \bibinfo {pages} {465–469} (\bibinfo {year} {1989})}\BibitemShut {NoStop}%
\bibitem [{\citenamefont {Sobol}(1967)}]{SOBOL196786}%
  \BibitemOpen
  \bibfield  {author} {\bibinfo {author} {\bibfnamefont {I.~M.}\ \bibnamefont {Sobol}},\ }\bibfield  {title} {\bibinfo {title} {On the distribution of points in a cube and the approximate evaluation of integrals},\ }\href@noop {} {\bibfield  {journal} {\bibinfo  {journal} {USSR Comput. Math. Math. Phys.}\ }\textbf {\bibinfo {volume} {7}},\ \bibinfo {pages} {86} (\bibinfo {year} {1967})}\BibitemShut {NoStop}%
\bibitem [{\citenamefont {Roux}(2002)}]{roux}%
  \BibitemOpen
  \bibfield  {author} {\bibinfo {author} {\bibfnamefont {B.}~\bibnamefont {Roux}},\ }\bibfield  {title} {\bibinfo {title} {{Computational studies of the gramicidin channel}},\ }\href@noop {} {\bibfield  {journal} {\bibinfo  {journal} {Acc. Chem. Res.}\ }\textbf {\bibinfo {volume} {35}},\ \bibinfo {pages} {366} (\bibinfo {year} {2002})}\BibitemShut {NoStop}%
\bibitem [{\citenamefont {Allen}\ \emph {et~al.}(2003)\citenamefont {Allen}, \citenamefont {Ba{\c{s}}tu{\u{g}}}, \citenamefont {Kuyucak},\ and\ \citenamefont {Chung}}]{allen2003gramicidin}%
  \BibitemOpen
  \bibfield  {author} {\bibinfo {author} {\bibfnamefont {T.~W.}\ \bibnamefont {Allen}}, \bibinfo {author} {\bibfnamefont {T.}~\bibnamefont {Ba{\c{s}}tu{\u{g}}}}, \bibinfo {author} {\bibfnamefont {S.}~\bibnamefont {Kuyucak}},\ and\ \bibinfo {author} {\bibfnamefont {S.-H.}\ \bibnamefont {Chung}},\ }\bibfield  {title} {\bibinfo {title} {{Gramicidin A channel as a test ground for molecular dynamics force fields}},\ }\href@noop {} {\bibfield  {journal} {\bibinfo  {journal} {Biophys. J.}\ }\textbf {\bibinfo {volume} {84}},\ \bibinfo {pages} {2159} (\bibinfo {year} {2003})}\BibitemShut {NoStop}%
\bibitem [{\citenamefont {Pullman}\ and\ \citenamefont {Etchebest}(1983)}]{pullman1983gramicidin}%
  \BibitemOpen
  \bibfield  {author} {\bibinfo {author} {\bibfnamefont {A.}~\bibnamefont {Pullman}}\ and\ \bibinfo {author} {\bibfnamefont {C.}~\bibnamefont {Etchebest}},\ }\bibfield  {title} {\bibinfo {title} {{The gramicidin A channel: the energy profile for single and double occupancy in a head-to-head $\beta$6. 33, 3-helical dimer backbone}},\ }\href@noop {} {\bibfield  {journal} {\bibinfo  {journal} {FEBS Lett.}\ }\textbf {\bibinfo {volume} {163}},\ \bibinfo {pages} {199} (\bibinfo {year} {1983})}\BibitemShut {NoStop}%
\bibitem [{\citenamefont {Bj\"{o}rklund}\ \emph {et~al.}(2009)\citenamefont {Bj\"{o}rklund}, \citenamefont {Husfeldt},\ and\ \citenamefont {Koivisto}}]{PIE}%
  \BibitemOpen
  \bibfield  {author} {\bibinfo {author} {\bibfnamefont {A.}~\bibnamefont {Bj\"{o}rklund}}, \bibinfo {author} {\bibfnamefont {T.}~\bibnamefont {Husfeldt}},\ and\ \bibinfo {author} {\bibfnamefont {M.}~\bibnamefont {Koivisto}},\ }\bibfield  {title} {\bibinfo {title} {Set partitioning via inclusion exclusion},\ }\href@noop {} {\bibfield  {journal} {\bibinfo  {journal} {SIAM J. Comput.}\ }\textbf {\bibinfo {volume} {39}},\ \bibinfo {pages} {546} (\bibinfo {year} {2009})}\BibitemShut {NoStop}%
\bibitem [{\citenamefont {Dey}\ and\ \citenamefont {Shah}(1997)}]{dey1997267}%
  \BibitemOpen
  \bibfield  {author} {\bibinfo {author} {\bibfnamefont {T.~K.}\ \bibnamefont {Dey}}\ and\ \bibinfo {author} {\bibfnamefont {N.~R.}\ \bibnamefont {Shah}},\ }\bibfield  {title} {\bibinfo {title} {On the number of simplicial complexes in rd},\ }\href@noop {} {\bibfield  {journal} {\bibinfo  {journal} {Comput. Geom.}\ }\textbf {\bibinfo {volume} {8}},\ \bibinfo {pages} {267} (\bibinfo {year} {1997})}\BibitemShut {NoStop}%
\bibitem [{\citenamefont {Adams}\ and\ \citenamefont {Franzosa}(2008)}]{adams2008introduction}%
  \BibitemOpen
  \bibfield  {author} {\bibinfo {author} {\bibfnamefont {C.~C.}\ \bibnamefont {Adams}}\ and\ \bibinfo {author} {\bibfnamefont {R.~D.}\ \bibnamefont {Franzosa}},\ }\href@noop {} {\emph {\bibinfo {title} {Introduction to topology: pure and applied}}},\ \bibinfo {number} {Sirsi) i9780131848696}\ (\bibinfo {year} {2008})\BibitemShut {NoStop}%
\bibitem [{\citenamefont {Berger}\ \emph {et~al.}(1984)\citenamefont {Berger}, \citenamefont {Pansu}, \citenamefont {Berry},\ and\ \citenamefont {Saint-Raymond}}]{berger1984affine}%
  \BibitemOpen
  \bibfield  {author} {\bibinfo {author} {\bibfnamefont {M.}~\bibnamefont {Berger}}, \bibinfo {author} {\bibfnamefont {P.}~\bibnamefont {Pansu}}, \bibinfo {author} {\bibfnamefont {J.-P.}\ \bibnamefont {Berry}},\ and\ \bibinfo {author} {\bibfnamefont {X.}~\bibnamefont {Saint-Raymond}},\ }\bibfield  {title} {\bibinfo {title} {Affine spaces},\ }in\ \href@noop {} {\emph {\bibinfo {booktitle} {Problems in Geometry}}}\ (\bibinfo  {publisher} {Springer},\ \bibinfo {year} {1984})\ p.~\bibinfo {pages} {11}\BibitemShut {NoStop}%
\bibitem [{\citenamefont {Bowyer}(1981)}]{Bowyer_Dirichlet}%
  \BibitemOpen
  \bibfield  {author} {\bibinfo {author} {\bibfnamefont {A.}~\bibnamefont {Bowyer}},\ }\bibfield  {title} {\bibinfo {title} {Computing dirichlet tessellations},\ }\href@noop {} {\bibfield  {journal} {\bibinfo  {journal} {Comput. J.}\ }\textbf {\bibinfo {volume} {24}},\ \bibinfo {pages} {162–166} (\bibinfo {year} {1981})}\BibitemShut {NoStop}%
\bibitem [{\citenamefont {Watson}(1981)}]{Watson_Voronoi}%
  \BibitemOpen
  \bibfield  {author} {\bibinfo {author} {\bibfnamefont {D.}~\bibnamefont {Watson}},\ }\bibfield  {title} {\bibinfo {title} {Computing the n-dimensional delaunay tessellation with applications to voronoi polytopes},\ }\href@noop {} {\bibfield  {journal} {\bibinfo  {journal} {Comput. J.}\ }\textbf {\bibinfo {volume} {24}},\ \bibinfo {pages} {167–172} (\bibinfo {year} {1981})}\BibitemShut {NoStop}%
\bibitem [{\citenamefont {Aurenhammer}(1991)}]{Voronoi-1}%
  \BibitemOpen
  \bibfield  {author} {\bibinfo {author} {\bibfnamefont {F.}~\bibnamefont {Aurenhammer}},\ }\bibfield  {title} {\bibinfo {title} {Voronoi diagrams --- a survey of a fundamental geometric data structure},\ }\href@noop {} {\bibfield  {journal} {\bibinfo  {journal} {ACM Comput. Survey}\ }\textbf {\bibinfo {volume} {23}},\ \bibinfo {pages} {345} (\bibinfo {year} {1991})}\BibitemShut {NoStop}%
\bibitem [{\citenamefont {Okabe}\ \emph {et~al.}(2000)\citenamefont {Okabe}, \citenamefont {Boots}, \citenamefont {Sugihara},\ and\ \citenamefont {Chiu}}]{Voronoi-2}%
  \BibitemOpen
  \bibfield  {author} {\bibinfo {author} {\bibfnamefont {A.}~\bibnamefont {Okabe}}, \bibinfo {author} {\bibfnamefont {B.}~\bibnamefont {Boots}}, \bibinfo {author} {\bibfnamefont {K.}~\bibnamefont {Sugihara}},\ and\ \bibinfo {author} {\bibfnamefont {S.~N.}\ \bibnamefont {Chiu}},\ }\href@noop {} {\emph {\bibinfo {title} {Spatial Tessellations --- Concepts and applications of Voronoi diagrams}}}\ (\bibinfo  {publisher} {John Wiley and Sons},\ \bibinfo {year} {2000})\BibitemShut {NoStop}%
\bibitem [{\citenamefont {Hert}\ and\ \citenamefont {Seel}(2018)}]{cgal}%
  \BibitemOpen
  \bibfield  {author} {\bibinfo {author} {\bibfnamefont {S.}~\bibnamefont {Hert}}\ and\ \bibinfo {author} {\bibfnamefont {M.}~\bibnamefont {Seel}},\ }\bibinfo {title} {{CGAL} user and reference manual}\ (\bibinfo  {publisher} {CGAL Editorial Board},\ \bibinfo {year} {2018})\ Chap.\ \bibinfo {chapter} {{dD} Convex Hulls and Delaunay Triangulations},\ \bibinfo {edition} {4th}\ ed.\BibitemShut {Stop}%
\bibitem [{\citenamefont {Farin}(1990)}]{Dirichlet-tessellations-Farin}%
  \BibitemOpen
  \bibfield  {author} {\bibinfo {author} {\bibfnamefont {G.}~\bibnamefont {Farin}},\ }\bibfield  {title} {\bibinfo {title} {Surfaces over dirichlet tessellations},\ }\href@noop {} {\bibfield  {journal} {\bibinfo  {journal} {Comput Aided Geom Des}\ }\textbf {\bibinfo {volume} {7}},\ \bibinfo {pages} {281} (\bibinfo {year} {1990})}\BibitemShut {NoStop}%
\bibitem [{\citenamefont {Maseras}\ and\ \citenamefont {Morokuma}(1995)}]{oniom}%
  \BibitemOpen
  \bibfield  {author} {\bibinfo {author} {\bibfnamefont {F.}~\bibnamefont {Maseras}}\ and\ \bibinfo {author} {\bibfnamefont {K.}~\bibnamefont {Morokuma}},\ }\bibfield  {title} {\bibinfo {title} {Imomm: A new integrated ab initio + molecular mechanics geometry optimization scheme of equilibrium structures and transition states},\ }\href@noop {} {\bibfield  {journal} {\bibinfo  {journal} {J. Comput. Chem.}\ }\textbf {\bibinfo {volume} {16}},\ \bibinfo {pages} {1170} (\bibinfo {year} {1995})}\BibitemShut {NoStop}%
\bibitem [{\citenamefont {Nandi}\ \emph {et~al.}(2021)\citenamefont {Nandi}, \citenamefont {Qu}, \citenamefont {Houston}, \citenamefont {Conte},\ and\ \citenamefont {Bowman}}]{bowman-2021-deltaML}%
  \BibitemOpen
  \bibfield  {author} {\bibinfo {author} {\bibfnamefont {A.}~\bibnamefont {Nandi}}, \bibinfo {author} {\bibfnamefont {C.}~\bibnamefont {Qu}}, \bibinfo {author} {\bibfnamefont {P.~L.}\ \bibnamefont {Houston}}, \bibinfo {author} {\bibfnamefont {R.}~\bibnamefont {Conte}},\ and\ \bibinfo {author} {\bibfnamefont {J.~M.}\ \bibnamefont {Bowman}},\ }\bibfield  {title} {\bibinfo {title} {${\Delta}$-machine learning for potential energy surfaces: A pip approach to bring a dft-based pes to ccsd(t) level of theory},\ }\href@noop {} {\bibfield  {journal} {\bibinfo  {journal} {J. Chem. Phys.}\ }\textbf {\bibinfo {volume} {154}},\ \bibinfo {pages} {051102} (\bibinfo {year} {2021})}\BibitemShut {NoStop}%
\bibitem [{\citenamefont {Ramakrishnan}\ \emph {et~al.}(2015)\citenamefont {Ramakrishnan}, \citenamefont {Dral}, \citenamefont {Rupp},\ and\ \citenamefont {von Lilienfeld}}]{delta_ml_intr}%
  \BibitemOpen
  \bibfield  {author} {\bibinfo {author} {\bibfnamefont {R.}~\bibnamefont {Ramakrishnan}}, \bibinfo {author} {\bibfnamefont {P.~O.}\ \bibnamefont {Dral}}, \bibinfo {author} {\bibfnamefont {M.}~\bibnamefont {Rupp}},\ and\ \bibinfo {author} {\bibfnamefont {O.~A.}\ \bibnamefont {von Lilienfeld}},\ }\bibfield  {title} {\bibinfo {title} {Big data meets quantum chemistry approximations: The ${\Delta}$-machine learning approach},\ }\href@noop {} {\bibfield  {journal} {\bibinfo  {journal} {J. Chem. Theory Comput.}\ }\textbf {\bibinfo {volume} {11}},\ \bibinfo {pages} {2087} (\bibinfo {year} {2015})}\BibitemShut {NoStop}%
\bibitem [{\citenamefont {Chen}\ \emph {et~al.}(2022)\citenamefont {Chen}, \citenamefont {Xu},\ and\ \citenamefont {Zhang}}]{delta_ml_2}%
  \BibitemOpen
  \bibfield  {author} {\bibinfo {author} {\bibfnamefont {J.}~\bibnamefont {Chen}}, \bibinfo {author} {\bibfnamefont {W.}~\bibnamefont {Xu}},\ and\ \bibinfo {author} {\bibfnamefont {R.}~\bibnamefont {Zhang}},\ }\bibfield  {title} {\bibinfo {title} {${\Delta}$-machine learning-driven discovery of double hybrid organic–inorganic perovskites},\ }\href@noop {} {\bibfield  {journal} {\bibinfo  {journal} {J. Mater. Chem. A}\ }\textbf {\bibinfo {volume} {10}},\ \bibinfo {pages} {1402} (\bibinfo {year} {2022})}\BibitemShut {NoStop}%
\bibitem [{\citenamefont {Unzueta}\ \emph {et~al.}(2021)\citenamefont {Unzueta}, \citenamefont {Greenwell},\ and\ \citenamefont {Beran}}]{delta_ml_3}%
  \BibitemOpen
  \bibfield  {author} {\bibinfo {author} {\bibfnamefont {P.~A.}\ \bibnamefont {Unzueta}}, \bibinfo {author} {\bibfnamefont {C.~S.}\ \bibnamefont {Greenwell}},\ and\ \bibinfo {author} {\bibfnamefont {G.~J.~O.}\ \bibnamefont {Beran}},\ }\bibfield  {title} {\bibinfo {title} {Predicting density functional theory-quality nuclear magnetic resonance chemical shifts via ${\Delta}$-machine learning},\ }\href@noop {} {\bibfield  {journal} {\bibinfo  {journal} {J. Chem. Theory Comput.}\ }\textbf {\bibinfo {volume} {17}},\ \bibinfo {pages} {826} (\bibinfo {year} {2021})}\BibitemShut {NoStop}%
\bibitem [{\citenamefont {Nandi}\ \emph {et~al.}(2024)\citenamefont {Nandi}, \citenamefont {Pandey}, \citenamefont {Houston}, \citenamefont {Qu}, \citenamefont {Yu}, \citenamefont {Conte}, \citenamefont {Tkatchenko},\ and\ \citenamefont {Bowman}}]{deltaml_2024}%
  \BibitemOpen
  \bibfield  {author} {\bibinfo {author} {\bibfnamefont {A.}~\bibnamefont {Nandi}}, \bibinfo {author} {\bibfnamefont {P.}~\bibnamefont {Pandey}}, \bibinfo {author} {\bibfnamefont {P.~L.}\ \bibnamefont {Houston}}, \bibinfo {author} {\bibfnamefont {C.}~\bibnamefont {Qu}}, \bibinfo {author} {\bibfnamefont {Q.}~\bibnamefont {Yu}}, \bibinfo {author} {\bibfnamefont {R.}~\bibnamefont {Conte}}, \bibinfo {author} {\bibfnamefont {A.}~\bibnamefont {Tkatchenko}},\ and\ \bibinfo {author} {\bibfnamefont {J.~M.}\ \bibnamefont {Bowman}},\ }\bibfield  {title} {\bibinfo {title} {${\Delta}$-machine learning to elevate dft-based potentials and a force field to the ccsd(t) level illustrated for ethanol},\ }\href@noop {} {\bibfield  {journal} {\bibinfo  {journal} {J. Chem. Theory Comput.}\ }\textbf {\bibinfo {volume} {20}},\ \bibinfo {pages} {8807} (\bibinfo {year} {2024})}\BibitemShut {NoStop}%
\bibitem [{\citenamefont {Riesz}\ and\ \citenamefont {Sz.-Nagy}(1990)}]{Riesznagy}%
  \BibitemOpen
  \bibfield  {author} {\bibinfo {author} {\bibfnamefont {F.}~\bibnamefont {Riesz}}\ and\ \bibinfo {author} {\bibfnamefont {B.}~\bibnamefont {Sz.-Nagy}},\ }\href@noop {} {\emph {\bibinfo {title} {Functional Analysis}}}\ (\bibinfo  {publisher} {Dover Publications, Inc., Mineola, New York},\ \bibinfo {year} {1990})\BibitemShut {NoStop}%
\bibitem [{\citenamefont {Bahdanau}\ \emph {et~al.}(2016)\citenamefont {Bahdanau}, \citenamefont {Cho},\ and\ \citenamefont {Bengio}}]{softmax_att}%
  \BibitemOpen
  \bibfield  {author} {\bibinfo {author} {\bibfnamefont {D.}~\bibnamefont {Bahdanau}}, \bibinfo {author} {\bibfnamefont {K.}~\bibnamefont {Cho}},\ and\ \bibinfo {author} {\bibfnamefont {Y.}~\bibnamefont {Bengio}},\ }\href@noop {} {\bibinfo {title} {Neural machine translation by jointly learning to align and translate}} (\bibinfo {year} {2016})\BibitemShut {NoStop}%
\bibitem [{\citenamefont {Microsoft}(2025)}]{microsoft2025copilot}%
  \BibitemOpen
  \bibfield  {author} {\bibinfo {author} {\bibnamefont {Microsoft}},\ }\href@noop {} {\bibinfo {title} {Copilot (july 10 version)}} (\bibinfo {year} {2025}),\ \bibinfo {note} {large language model}\BibitemShut {NoStop}%
\bibitem [{\citenamefont {Zundel}(2012)}]{zundel2012hydration}%
  \BibitemOpen
  \bibfield  {author} {\bibinfo {author} {\bibfnamefont {G.}~\bibnamefont {Zundel}},\ }\href@noop {} {\emph {\bibinfo {title} {Hydration and Intermolecular Interaction: Infrared Investigations with Polyelectrolyte Membranes}}}\ (\bibinfo  {publisher} {Academic Press},\ \bibinfo {year} {2012})\BibitemShut {NoStop}%
\bibitem [{\citenamefont {Tuckerman}\ \emph {et~al.}(1996)\citenamefont {Tuckerman}, \citenamefont {Ungar}, \citenamefont {Vonrosenvinge},\ and\ \citenamefont {Klein}}]{cptuckerman3}%
  \BibitemOpen
  \bibfield  {author} {\bibinfo {author} {\bibfnamefont {M.~E.}\ \bibnamefont {Tuckerman}}, \bibinfo {author} {\bibfnamefont {P.~J.}\ \bibnamefont {Ungar}}, \bibinfo {author} {\bibfnamefont {T.}~\bibnamefont {Vonrosenvinge}},\ and\ \bibinfo {author} {\bibfnamefont {M.~L.}\ \bibnamefont {Klein}},\ }\bibfield  {title} {\bibinfo {title} {Ab initio molecular dynamics simulations},\ }\href@noop {} {\bibfield  {journal} {\bibinfo  {journal} {J. Phys. Chem.}\ }\textbf {\bibinfo {volume} {100}},\ \bibinfo {pages} {12878} (\bibinfo {year} {1996})}\BibitemShut {NoStop}%
\bibitem [{\citenamefont {Tuckerman}\ \emph {et~al.}(1995)\citenamefont {Tuckerman}, \citenamefont {Laasonen}, \citenamefont {Sprik},\ and\ \citenamefont {Parrinello}}]{h+oh-solv}%
  \BibitemOpen
  \bibfield  {author} {\bibinfo {author} {\bibfnamefont {M.}~\bibnamefont {Tuckerman}}, \bibinfo {author} {\bibfnamefont {K.}~\bibnamefont {Laasonen}}, \bibinfo {author} {\bibfnamefont {M.}~\bibnamefont {Sprik}},\ and\ \bibinfo {author} {\bibfnamefont {M.}~\bibnamefont {Parrinello}},\ }\href@noop {} {\bibfield  {journal} {\bibinfo  {journal} {J. Phys. Chem.}\ }\textbf {\bibinfo {volume} {99}},\ \bibinfo {pages} {5749} (\bibinfo {year} {1995})}\BibitemShut {NoStop}%
\bibitem [{\citenamefont {Schmitt}\ and\ \citenamefont {Voth}(1999)}]{schmittvothprotontransport1999}%
  \BibitemOpen
  \bibfield  {author} {\bibinfo {author} {\bibfnamefont {U.~W.}\ \bibnamefont {Schmitt}}\ and\ \bibinfo {author} {\bibfnamefont {G.~A.}\ \bibnamefont {Voth}},\ }\bibfield  {title} {\bibinfo {title} {{The computer simulation of proton transport in water}},\ }\href@noop {} {\bibfield  {journal} {\bibinfo  {journal} {J. Chem. Phys.}\ }\textbf {\bibinfo {volume} {111}},\ \bibinfo {pages} {9361} (\bibinfo {year} {1999})}\BibitemShut {NoStop}%
\bibitem [{\citenamefont {Schmitt}\ and\ \citenamefont {Voth}(1998)}]{schmitt1998multistate}%
  \BibitemOpen
  \bibfield  {author} {\bibinfo {author} {\bibfnamefont {U.~W.}\ \bibnamefont {Schmitt}}\ and\ \bibinfo {author} {\bibfnamefont {G.~A.}\ \bibnamefont {Voth}},\ }\bibfield  {title} {\bibinfo {title} {{Multistate empirical valence bond model for proton transport in water}},\ }\href@noop {} {\bibfield  {journal} {\bibinfo  {journal} {J. Phys. Chem. B}\ }\textbf {\bibinfo {volume} {102}},\ \bibinfo {pages} {5547} (\bibinfo {year} {1998})}\BibitemShut {NoStop}%
\bibitem [{\citenamefont {Pomes}\ and\ \citenamefont {Roux}(1996{\natexlab{a}})}]{protonwire1}%
  \BibitemOpen
  \bibfield  {author} {\bibinfo {author} {\bibfnamefont {R.}~\bibnamefont {Pomes}}\ and\ \bibinfo {author} {\bibfnamefont {B.}~\bibnamefont {Roux}},\ }\bibfield  {title} {\bibinfo {title} {Structure and dynamics of a proton wire: A theoretical study of {$H^+$} translocation along the single-file water chain in the gramicidin a channel},\ }\href@noop {} {\bibfield  {journal} {\bibinfo  {journal} {Biophys J.}\ }\textbf {\bibinfo {volume} {71}},\ \bibinfo {pages} {19} (\bibinfo {year} {1996}{\natexlab{a}})}\BibitemShut {NoStop}%
\bibitem [{\citenamefont {McEwan}\ and\ \citenamefont {Phillips}(1975)}]{atmosph-clusters1}%
  \BibitemOpen
  \bibfield  {author} {\bibinfo {author} {\bibfnamefont {M.~J.}\ \bibnamefont {McEwan}}\ and\ \bibinfo {author} {\bibfnamefont {L.~F.}\ \bibnamefont {Phillips}},\ }\href@noop {} {\emph {\bibinfo {title} {Chemistry of the Atmosphere}}}\ (\bibinfo  {publisher} {Eward Arnold: London},\ \bibinfo {year} {1975})\BibitemShut {NoStop}%
\bibitem [{\citenamefont {Wayne}(1994)}]{atmosph-clusters2}%
  \BibitemOpen
  \bibfield  {author} {\bibinfo {author} {\bibfnamefont {R.~P.}\ \bibnamefont {Wayne}},\ }\href@noop {} {\emph {\bibinfo {title} {Chemistry of the Atmosphere}}}\ (\bibinfo  {publisher} {Clarendon Press: Oxford},\ \bibinfo {year} {1994})\BibitemShut {NoStop}%
\bibitem [{\citenamefont {Pomes}\ and\ \citenamefont {Roux}(1996{\natexlab{b}})}]{pomesroux2}%
  \BibitemOpen
  \bibfield  {author} {\bibinfo {author} {\bibfnamefont {R.}~\bibnamefont {Pomes}}\ and\ \bibinfo {author} {\bibfnamefont {B.}~\bibnamefont {Roux}},\ }\bibfield  {title} {\bibinfo {title} {{Theoretical study of H+ translocation along a model proton wire}},\ }\href@noop {} {\bibfield  {journal} {\bibinfo  {journal} {J. Phys. Chem.}\ }\textbf {\bibinfo {volume} {100}},\ \bibinfo {pages} {2519} (\bibinfo {year} {1996}{\natexlab{b}})}\BibitemShut {NoStop}%
\bibitem [{\citenamefont {Decornez}\ \emph {et~al.}(1999)\citenamefont {Decornez}, \citenamefont {Drukker},\ and\ \citenamefont {Hammes-Schiffer}}]{protonwire3}%
  \BibitemOpen
  \bibfield  {author} {\bibinfo {author} {\bibfnamefont {H.}~\bibnamefont {Decornez}}, \bibinfo {author} {\bibfnamefont {K.}~\bibnamefont {Drukker}},\ and\ \bibinfo {author} {\bibfnamefont {S.}~\bibnamefont {Hammes-Schiffer}},\ }\bibfield  {title} {\bibinfo {title} {Solvation and hydrogen-bonding effects on proton wires},\ }\href@noop {} {\bibfield  {journal} {\bibinfo  {journal} {J. Phys. Chem. A}\ }\textbf {\bibinfo {volume} {103}},\ \bibinfo {pages} {2891} (\bibinfo {year} {1999})}\BibitemShut {NoStop}%
\bibitem [{\citenamefont {Brewer}\ \emph {et~al.}(2001)\citenamefont {Brewer}, \citenamefont {Schmitt},\ and\ \citenamefont {Voth}}]{protonwire4}%
  \BibitemOpen
  \bibfield  {author} {\bibinfo {author} {\bibfnamefont {M.~L.}\ \bibnamefont {Brewer}}, \bibinfo {author} {\bibfnamefont {U.~W.}\ \bibnamefont {Schmitt}},\ and\ \bibinfo {author} {\bibfnamefont {G.~A.}\ \bibnamefont {Voth}},\ }\bibfield  {title} {\bibinfo {title} {The formation and dynamics of proton wires in channel environments},\ }\href@noop {} {\bibfield  {journal} {\bibinfo  {journal} {Biophys J.}\ }\textbf {\bibinfo {volume} {80}},\ \bibinfo {pages} {1691} (\bibinfo {year} {2001})}\BibitemShut {NoStop}%
\bibitem [{\citenamefont {Teeter}(1984)}]{teeter}%
  \BibitemOpen
  \bibfield  {author} {\bibinfo {author} {\bibfnamefont {M.}~\bibnamefont {Teeter}},\ }\bibfield  {title} {\bibinfo {title} {{Water structure of a hydrophobic protein at atomic resolution: Pentagon rings of water molecules in crystals of crambin}},\ }\href@noop {} {\bibfield  {journal} {\bibinfo  {journal} {Proceedings of the National Academy of Sciences}\ }\textbf {\bibinfo {volume} {81}},\ \bibinfo {pages} {6014} (\bibinfo {year} {1984})}\BibitemShut {NoStop}%
\bibitem [{\citenamefont {Neidle}\ \emph {et~al.}(1980)\citenamefont {Neidle}, \citenamefont {Berman},\ and\ \citenamefont {Shieh}}]{bio-clusters2}%
  \BibitemOpen
  \bibfield  {author} {\bibinfo {author} {\bibfnamefont {S.}~\bibnamefont {Neidle}}, \bibinfo {author} {\bibfnamefont {H.~M.}\ \bibnamefont {Berman}},\ and\ \bibinfo {author} {\bibfnamefont {H.~S.}\ \bibnamefont {Shieh}},\ }\bibfield  {title} {\bibinfo {title} {Highly structured water network in crystals of a deoxydinucleoside-drug complex},\ }\href@noop {} {\bibfield  {journal} {\bibinfo  {journal} {Nature}\ }\textbf {\bibinfo {volume} {288}},\ \bibinfo {pages} {129} (\bibinfo {year} {1980})}\BibitemShut {NoStop}%
\bibitem [{\citenamefont {Lipscomb}\ \emph {et~al.}(1994)\citenamefont {Lipscomb}, \citenamefont {Peek}, \citenamefont {Zhou}, \citenamefont {Bertrand}, \citenamefont {VanDerveer},\ and\ \citenamefont {Williams}}]{lipscombpeek}%
  \BibitemOpen
  \bibfield  {author} {\bibinfo {author} {\bibfnamefont {L.~A.}\ \bibnamefont {Lipscomb}}, \bibinfo {author} {\bibfnamefont {M.~E.}\ \bibnamefont {Peek}}, \bibinfo {author} {\bibfnamefont {F.~X.}\ \bibnamefont {Zhou}}, \bibinfo {author} {\bibfnamefont {J.~A.}\ \bibnamefont {Bertrand}}, \bibinfo {author} {\bibfnamefont {D.}~\bibnamefont {VanDerveer}},\ and\ \bibinfo {author} {\bibfnamefont {L.~D.}\ \bibnamefont {Williams}},\ }\bibfield  {title} {\bibinfo {title} {{Water ring structure at DNA interfaces: hydration and dynamics of DNA-anthracycline complexes}},\ }\href@noop {} {\bibfield  {journal} {\bibinfo  {journal} {Biochemistry}\ }\textbf {\bibinfo {volume} {33}},\ \bibinfo {pages} {3649} (\bibinfo {year} {1994})}\BibitemShut {NoStop}%
\bibitem [{\citenamefont {Tu}\ \emph {et~al.}(2002)\citenamefont {Tu}, \citenamefont {Rowlett}, \citenamefont {Tripp}, \citenamefont {Ferry},\ and\ \citenamefont {Silverman}}]{turowlett}%
  \BibitemOpen
  \bibfield  {author} {\bibinfo {author} {\bibfnamefont {C.}~\bibnamefont {Tu}}, \bibinfo {author} {\bibfnamefont {R.~S.}\ \bibnamefont {Rowlett}}, \bibinfo {author} {\bibfnamefont {B.~C.}\ \bibnamefont {Tripp}}, \bibinfo {author} {\bibfnamefont {J.~G.}\ \bibnamefont {Ferry}},\ and\ \bibinfo {author} {\bibfnamefont {D.~N.}\ \bibnamefont {Silverman}},\ }\bibfield  {title} {\bibinfo {title} {{Chemical rescue of proton transfer in catalysis by carbonic anhydrases in the $\beta$-and $\gamma$-class}},\ }\href@noop {} {\bibfield  {journal} {\bibinfo  {journal} {Biochemistry}\ }\textbf {\bibinfo {volume} {41}},\ \bibinfo {pages} {15429} (\bibinfo {year} {2002})}\BibitemShut {NoStop}%
\bibitem [{\citenamefont {Dietrick}\ and\ \citenamefont {Iyengar}(2012{\natexlab{b}})}]{Scott-proj}%
  \BibitemOpen
  \bibfield  {author} {\bibinfo {author} {\bibfnamefont {S.~M.}\ \bibnamefont {Dietrick}}\ and\ \bibinfo {author} {\bibfnamefont {S.~S.}\ \bibnamefont {Iyengar}},\ }\bibfield  {title} {\bibinfo {title} {Constructing periodic phase space orbits from ab initio molecular dynamics trajectories to analyze vibrational spectra: Case study of the zundel {(H$_5$O$_2^+$)} cation},\ }\href@noop {} {\bibfield  {journal} {\bibinfo  {journal} {J. Chem. Theory and Comput.}\ }\textbf {\bibinfo {volume} {8}},\ \bibinfo {pages} {4876} (\bibinfo {year} {2012}{\natexlab{b}})}\BibitemShut {NoStop}%
\bibitem [{\citenamefont {Sadhukhan}\ \emph {et~al.}(1999)\citenamefont {Sadhukhan}, \citenamefont {Munoz}, \citenamefont {Adamo},\ and\ \citenamefont {Scuseria}}]{H5O2+XCcomp}%
  \BibitemOpen
  \bibfield  {author} {\bibinfo {author} {\bibfnamefont {S.}~\bibnamefont {Sadhukhan}}, \bibinfo {author} {\bibfnamefont {D.}~\bibnamefont {Munoz}}, \bibinfo {author} {\bibfnamefont {C.}~\bibnamefont {Adamo}},\ and\ \bibinfo {author} {\bibfnamefont {G.~E.}\ \bibnamefont {Scuseria}},\ }\bibfield  {title} {\bibinfo {title} {Predicting proton transfer barriers with density functional methods},\ }\href@noop {} {\bibfield  {journal} {\bibinfo  {journal} {Chem. Phys. Lett.}\ }\textbf {\bibinfo {volume} {306}},\ \bibinfo {pages} {83} (\bibinfo {year} {1999})}\BibitemShut {NoStop}%
\bibitem [{\citenamefont {Xie}\ \emph {et~al.}(1994)\citenamefont {Xie}, \citenamefont {Remington},\ and\ \citenamefont {Schaefer}}]{Schaefer:94}%
  \BibitemOpen
  \bibfield  {author} {\bibinfo {author} {\bibfnamefont {Y.}~\bibnamefont {Xie}}, \bibinfo {author} {\bibfnamefont {R.~B.}\ \bibnamefont {Remington}},\ and\ \bibinfo {author} {\bibfnamefont {H.~F.}\ \bibnamefont {Schaefer}, \bibfnamefont {III}},\ }\bibfield  {title} {\bibinfo {title} {The protonated water dimer: Extensive theoretical studies of h$_5$o$_2^+$},\ }\href@noop {} {\bibfield  {journal} {\bibinfo  {journal} {J. Chem. Phys.}\ }\textbf {\bibinfo {volume} {101}},\ \bibinfo {pages} {4878} (\bibinfo {year} {1994})}\BibitemShut {NoStop}%
\bibitem [{\citenamefont {Klimes}\ and\ \citenamefont {Michaelides}(2012)}]{DFT-vDW-Michaelides}%
  \BibitemOpen
  \bibfield  {author} {\bibinfo {author} {\bibfnamefont {J.}~\bibnamefont {Klimes}}\ and\ \bibinfo {author} {\bibfnamefont {A.}~\bibnamefont {Michaelides}},\ }\bibfield  {title} {\bibinfo {title} {Perspective: Advances and challenges in treating van der waals dispersion forces in density functional theory},\ }\href@noop {} {\bibfield  {journal} {\bibinfo  {journal} {J. Chem. Phys.}\ }\textbf {\bibinfo {volume} {137}},\ \bibinfo {pages} {120901} (\bibinfo {year} {2012})}\BibitemShut {NoStop}%
\bibitem [{\citenamefont {Peverati}\ and\ \citenamefont {Truhlar}(2014)}]{Truhlar-functional-review}%
  \BibitemOpen
  \bibfield  {author} {\bibinfo {author} {\bibfnamefont {R.}~\bibnamefont {Peverati}}\ and\ \bibinfo {author} {\bibfnamefont {D.}~\bibnamefont {Truhlar}},\ }\bibfield  {title} {\bibinfo {title} {The quest for a universal density functional: The accuracy of density functionals across a broad spectrum of databases in chemistry and physics},\ }\href@noop {} {\bibfield  {journal} {\bibinfo  {journal} {Philos. Trans. Royal Soc. A}\ }\textbf {\bibinfo {volume} {372}},\ \bibinfo {pages} {10120476} (\bibinfo {year} {2014})}\BibitemShut {NoStop}%
\bibitem [{\citenamefont {Cohen}\ \emph {et~al.}(2012)\citenamefont {Cohen}, \citenamefont {Mori-Sánchez},\ and\ \citenamefont {Yang}}]{yang-dft-review}%
  \BibitemOpen
  \bibfield  {author} {\bibinfo {author} {\bibfnamefont {A.~J.}\ \bibnamefont {Cohen}}, \bibinfo {author} {\bibfnamefont {P.}~\bibnamefont {Mori-Sánchez}},\ and\ \bibinfo {author} {\bibfnamefont {W.}~\bibnamefont {Yang}},\ }\bibfield  {title} {\bibinfo {title} {Challenges for density functional theory},\ }\href@noop {} {\bibfield  {journal} {\bibinfo  {journal} {Chem. Revs.}\ }\textbf {\bibinfo {volume} {112}},\ \bibinfo {pages} {289} (\bibinfo {year} {2012})}\BibitemShut {NoStop}%
\bibitem [{\citenamefont {Hartigan}\ and\ \citenamefont {Wong}(1979)}]{kmeans}%
  \BibitemOpen
  \bibfield  {author} {\bibinfo {author} {\bibfnamefont {J.~A.}\ \bibnamefont {Hartigan}}\ and\ \bibinfo {author} {\bibfnamefont {M.~A.}\ \bibnamefont {Wong}},\ }\bibfield  {title} {\bibinfo {title} {Algorithm as 136: A k-means clustering algorithm},\ }\href@noop {} {\bibfield  {journal} {\bibinfo  {journal} {J. R. Stat. Soc. series c (applied statistics)}\ }\textbf {\bibinfo {volume} {28}},\ \bibinfo {pages} {100} (\bibinfo {year} {1979})}\BibitemShut {NoStop}%
\bibitem [{\citenamefont {{Google Gemini}}(2025)}]{GoogleGemini_YYYYMMDD}%
  \BibitemOpen
  \bibfield  {author} {\bibinfo {author} {\bibnamefont {{Google Gemini}}},\ }\href@noop {} {\bibinfo {title} {Gemini ai chatbot}} (\bibinfo {year} {2025}),\ \bibinfo {note} {accessed: 2025-07-10. Version: [Gemini Advanced].}\BibitemShut {Stop}%
\bibitem [{\citenamefont {Schlegel}\ and\ \citenamefont {Frisch}(1991{\natexlab{b}})}]{Schlegel-Bottlenecks}%
  \BibitemOpen
  \bibfield  {author} {\bibinfo {author} {\bibfnamefont {H.~B.}\ \bibnamefont {Schlegel}}\ and\ \bibinfo {author} {\bibfnamefont {M.~J.}\ \bibnamefont {Frisch}},\ }\bibinfo {title} {Computational bottlenecks in molecular orbital calculations},\ in\ \href@noop {} {\emph {\bibinfo {booktitle} {Theoretical and Computational Models for Organic Chemistry}}},\ \bibinfo {editor} {edited by\ \bibinfo {editor} {\bibfnamefont {S.~J.}\ \bibnamefont {Formosinho}}, \bibinfo {editor} {\bibfnamefont {I.~G.}\ \bibnamefont {Csizmadia}},\ and\ \bibinfo {editor} {\bibfnamefont {L.~G.}\ \bibnamefont {Arnaut}}}\ (\bibinfo  {publisher} {Springer Netherlands},\ \bibinfo {address} {Dordrecht},\ \bibinfo {year} {1991})\ pp.\ \bibinfo {pages} {5--33}\BibitemShut {NoStop}%
\bibitem [{\citenamefont {Lim}\ and\ \citenamefont {Nelson}(2022)}]{Equivariant-NNs-1}%
  \BibitemOpen
  \bibfield  {author} {\bibinfo {author} {\bibfnamefont {L.-H.}\ \bibnamefont {Lim}}\ and\ \bibinfo {author} {\bibfnamefont {B.~J.}\ \bibnamefont {Nelson}},\ }\href@noop {} {\bibinfo {title} {What is an equivariant neural network?}} (\bibinfo {year} {2022})\BibitemShut {NoStop}%
\bibitem [{\citenamefont {Kondor}(2025)}]{Equivariant-NNs-2}%
  \BibitemOpen
  \bibfield  {author} {\bibinfo {author} {\bibfnamefont {R.}~\bibnamefont {Kondor}},\ }\bibfield  {title} {\bibinfo {title} {The principles behind equivariant neural networks for physics and chemistry},\ }\href@noop {} {\bibfield  {journal} {\bibinfo  {journal} {Proceedings of the National Academy of Sciences}\ }\textbf {\bibinfo {volume} {122}},\ \bibinfo {pages} {e2415656122} (\bibinfo {year} {2025})}\BibitemShut {NoStop}%
\bibitem [{\citenamefont {Zundel}(1969)}]{Zundel1969hydration}%
  \BibitemOpen
  \bibfield  {author} {\bibinfo {author} {\bibfnamefont {G.}~\bibnamefont {Zundel}},\ }\href@noop {} {\emph {\bibinfo {title} {{Hydration and intermolecular interaction}}}}\ (\bibinfo  {publisher} {Academic Press},\ \bibinfo {year} {1969})\BibitemShut {NoStop}%
\end{thebibliography}

%

\end{document}